\documentclass{article}

\usepackage{iclr2027_conference_2,times}

\usepackage[utf8]{inputenc} % allow utf-8 input
\usepackage[T1]{fontenc}   % use 8-bit T1 fonts
\usepackage[english]{babel} 
\usepackage{alphabeta}
\usepackage{hyperref}       % hyperlinks
\usepackage{url}            % simple URL typesetting
\usepackage{booktabs}       % professional-quality tables
\usepackage{amsfonts}       % blackboard math symbols
\usepackage{nicefrac}       % compact symbols for 1/2, etc.
\usepackage{microtype}      % microtypography
\usepackage{xcolor}         % colors
\usepackage{amsmath}
\usepackage{graphicx}
\usepackage{comment}
\usepackage{cleveref}
\usepackage{rotating}
\usepackage{amssymb,amsthm}
\usepackage{mathtools}
\usepackage{tabularx,adjustbox,makecell}
\usepackage{algorithm}
\usepackage{array}
\usepackage{tabularray}
\usepackage{float}
\usepackage{codehigh}
\usepackage[normalem]{ulem}
\usepackage{caption}
\usepackage{subcaption}
\usepackage{listings}
\usepackage{enumitem}
\usepackage{wrapfig}
\usepackage{threeparttable}
\usepackage[most]{tcolorbox}
\usepackage{xurl} 
\usepackage{lineno}
\usepackage{tcolorbox}
\tcbset{
    examplebox/.style={
        colback=gray!10,
        colframe=gray!50,
        boxrule=0.4pt,
        arc=3pt,
        left=6pt,
        right=6pt,
        top=4pt,
        bottom=4pt,
        fontupper=\small\itshape
    }
}

\newcommand{\authsym}[1]{%
  \textsuperscript{\ensuremath{#1}}%
}

\usepackage{amsmath,amsfonts,bm}

\def\eqref#1{equation~\ref{#1}}
\def\1{\bm{1}}

\DeclareMathAlphabet{\mathsfit}{\encodingdefault}{\sfdefault}{m}{sl}
\SetMathAlphabet{\mathsfit}{bold}{\encodingdefault}{\sfdefault}{bx}{n}

\newcommand{\tableeqnum}{%
  \refstepcounter{equation}% Step the equation counter
  (\theequation)% Print the equation number enclosed in parentheses
}

\theoremstyle{plain}
\newtheorem{theorem}{Theorem}[section]
\newtheorem{proposition}[theorem]{Proposition}
\newtheorem{lemma}[theorem]{Lemma}
\newtheorem{corollary}[theorem]{Corollary}
\theoremstyle{definition}

\theoremstyle{remark}
\newtheorem{remark}[theorem]{Remark}

\title{Agent Evaluation Reliability: More Tasks Won't (Always) Fix An Agent Leaderboard}

\author{%
  \begin{tabular}{@{}l@{}}
    \textbf{Michael Hardy}\authsym{\diamondsuit,*}
    \quad
    \textbf{Ruhana Azam}\authsym{\clubsuit}
    \\
    \textbf{Anka Reuel}\authsym{\diamondsuit,\dagger}
    \quad
    \textbf{Mykel Kochenderfer}\authsym{\diamondsuit,\dagger}
    \quad
    \textbf{Sanmi Koyejo}\authsym{\diamondsuit,\dagger}
    \\[3pt]
    \authsym{\diamondsuit}\textmd{Stanford University,}
    \quad
    \authsym{*}\textmd{\texttt{hardym[\alpha]stanford$\cdot$edu}}
    \\
    \authsym{\clubsuit}\textmd{University of Illinois Urbana-Champaign}
    \quad
    \authsym{\dagger}\textmd{Senior Author}
  \end{tabular}%
}

\iclrfinalcopy % Uncomment for camera-ready version, but NOT for submission.
\begin{document}

\maketitle

\begin{abstract}
Agent evaluations are increasingly used to compare models, assess capabilities, and inform deployment decisions, yet observed scores can reflect not only the model but also the effects of the evaluation conditions such as the scaffolds or tasks. This makes reliability claim-dependent: an evaluation that reliably ranks deployed systems may not reliably rank underlying models or produce reliable absolute scores. We ask which conclusions current agent evaluations reliably support and what additional evaluation would actually improve them. Using Generalizability theory, we develop a Bayesian variance-decomposition framework for sparse, imbalanced agent leaderboards and apply it to 22 benchmarks from the Holistic Agent Leaderboard and Harbor Index. The framework separates \emph{signal}, performance differences relevant to the intended claim, from \emph{noise}, irrelevant variation that can still change scores or rankings. We find four practical results: (1) Reliability depends on the measurement goal. Fixed model--scaffold systems are ranked reliably ($E\rho^2=0.935$--$0.994$), while underlying-model reliability is substantially lower ($0.148$--$0.841$). (2) Scaffold choice can change evaluation conclusions. We introduce \emph{inter-scaffold reliability}, measuring whether scaffolds preserve model rankings, and show that scaffold effects vary substantially across evaluations. (3) More tasks cannot resolve all uncertainty. Even infinitely many similarly constructed tasks improve model-ranking reliability of the dataset by at most $0.097$ when uncertainty is dominated by limited scaffold coverage. (4) Pooling diverse benchmarks can improve cross-task rankings at lower cost. For rankings across diverse agentic tasks, pooling benchmarks raises projected reliability from $0.44$ to $0.75$ at the same task budget and can reduce projected cost by up to $83\%$. Evaluation design should therefore follow the intended claim: practitioners should identify what a score or ranking should mean, diagnose what limits its reliability, and spend evaluation budget on the sources of uncertainty that matter.\footnote{Code and data: \url{https://github.com/hardy-education/scaffold_eval}}

\end{abstract}

\section{Introduction}
\label{sec:introduction}

Agentic AI systems interact with software, tools, users, and environments to accomplish multi-step goals. Agent evaluations such as SWE-bench~\citep{jimenezSWEbenchCanLanguage2024a}
% , GAIA~\citep{mialonGAIABenchmarkGeneral2023}, 
and \(\tau\)-bench~\citep{yao$t$benchBenchmarkToolAgentUser2024} are increasingly used to test and rank such systems in technical reports, system cards, and policy discussions~\citep{anthropic2026claudeopus47,openai2026gpt55}. Yet, a score on such evaluations is only useful if we know what conclusions it can reliably support: Does an observed ranking reflect model performance differences that persist beyond the particular evaluation setup? Would the same score or ranking hold under a different scaffold, a different set of tasks, or a broader range of agentic settings?

These questions are particularly important for agent evaluations because performance depends on more than the underlying model. A scaffold\footnote{In the current literature, the software infrastructure and post-training methods surrounding the LLM allowing it to act is synonymous with ``scaffold'', ``harness'', and even ``agent''.} manages the state, exposes tools, and translates model outputs into actions; together, the model\footnote{Throughout, ``model'' denotes a base large language model paired with a reasoning-effort configuration; for example, \texttt{gpt-5-high} and \texttt{gpt-5-minimal} are treated as distinct, following our primary dataset.
% and common reporting practice.
} and scaffold form the system. This creates an immediate measurement choice: are we trying to evaluate the complete system as deployed, or isolate performance differences attributable to the underlying model? The answer changes what should count as signal and what should count as evaluation error.

Reliability also depends on the conclusion we want to draw. A practitioner may care whether a model ranking remains stable, whether an absolute score would remain similar, or whether performance generalizes across different agentic tasks. These are different measurement goals and need not have the same reliability. Existing agent reliability work primarily studies whether a fixed agent behaves consistently across repeated runs or prompt perturbations of the same task \citep{rabanser_towards_2026,Razavi2025BenchmarkingPS, yao$t$benchBenchmarkToolAgentUser2024}. We instead ask: \textbf{what claims do current agent evaluations reliably support, and what additional evaluation would make those claims more reliable?}

We use Generalizability Theory (G-theory) \citep{cronbach_dependability_1972,brennan_generalizability_2001} to separate variation associated with models, tasks, scaffolds, benchmarks, and their interactions. To word it differently, our framework separates \emph{signal}--performance differences that persist across the conditions an evaluation is meant to generalize over--from \emph{noise}--variation that is irrelevant to the intended claim but can still change the resulting score or ranking; reliability increases when signal dominates this noise. G-theory lets us identify not only how reliable an evaluation is for a particular claim, but also what limits that reliability and whether adding more tasks, scaffolds, or benchmark coverage would help. We develop a Bayesian variance-decomposition approach for the sparse and imbalanced designs common in agent leaderboards and apply it to $30{,}859$ rollouts from 9 benchmarks in the Holistic Agent Leaderboard (HAL) \citep{kapoor_holistic_2025} and 13 the Harbor Index \citep{shi_harbor_2026}.

Our results yield four practical findings:
\begin{itemize}[leftmargin=0.55cm,nosep]
    \item \textbf{Reliability depends on the measurement goal}. Fixed model--scaffold systems are ranked reliably $E\rho_{MA}^2\in[0.94$,$0.99]$, while rank reliability for the models is substantially lower \(E\rho_{M}^2\in[0.148,0.841]\). %; reliability of absolute scores is lower still.
    \item \textbf{Changing the scaffold can change evaluation conclusions}. We introduce \emph{inter-scaffold reliability}, which measures whether different scaffolds produce similar model rankings when evaluating the same models on the same tasks. Its posterior medians range from $0.151$ to $0.852$ across benchmarks. We further show that scaffold choice can change which tasks a model solves even without improving its overall benchmark score.
    \item \textbf{Adding more tasks does not always make an evaluation more reliable}. Under the observed scaffold coverage, even infinitely many similarly constructed tasks would improve model-ranking reliability by at most approximately $0.10$.
    \item \textbf{When the goal is to rank models across diverse agentic tasks, pooling diverse benchmarks can improve reliability at lower cost}. At the same task budget, projected ranking reliability rises from approximately $0.44$ within one benchmark to $0.75$ across the nine-benchmark battery, while reliability-aware allocation can reduce projected evaluation cost by up to $83\%$.
\end{itemize}

The broader lesson is that more evaluation is not automatically better evaluation. Practitioners should first specify what they want an evaluation result to mean, then identify which sources of uncertainty limit that claim, and allocate evaluation effort accordingly.

\section{Related Work and Background}
\label{sec:related}

\paragraph{Agent evaluation.}
Agent benchmarks span software engineering \citep{jimenezSWEbenchCanLanguage2024a}, web navigation \citep{zhouWebArenaRealisticWeb2024a,heWebVoyagerBuildingEndtoEnd2024}, general assistance \citep{mialonGAIABenchmarkGeneral2023}, and customer service \citep{yao$t$benchBenchmarkToolAgentUser2024}. Existing reliability studies primarily test whether a fixed agent behaves consistently across repeated runs, prompt perturbations, tool configurations, or environmental failures \citep{rabanser_towards_2026,Razavi2025BenchmarkingPS,wang2026agentnoisebench,kumar2025robustness}. These analyses assess robustness of a particular system. We instead study the reliability of the \emph{comparative inference}: whether the reported ordering of models or systems would persist under new tasks, scaffolds, or benchmarks. This distinction is consequential because a system can be repeatable within one harness while its rank is highly contingent on that harness.

% \paragraph{Reliability and Generalizability theory.}
% Classical reliability asks whether repeated measurements preserve distinctions among the objects being compared. Generalizability theory (G-theory) extends this idea to evaluations with multiple sources of variation, or \emph{facets}, by decomposing score variance and projecting reliability under alternative designs \citep{cronbach_dependability_1972,shavelson_generalizability_1989,brennan_generalizability_2001,cronbach_my_2004}. Related variance-decomposition methods show that benchmark conclusions can depend strongly on task sampling, evaluation conditions, and the population of models under comparison \citep{madaan_quantifying_2024,hardy_knowledge_2026,hardy_ai_2026}. We adapt G-theory to sparse, crossed agent evaluations and use it not only to diagnose unreliability, but also to identify which additional measurements would most efficiently improve it.

% \paragraph{Reliability depends on what is being ranked.} If the object of measurement is the joint model--scaffold system, then model--scaffold compatibility is part of the signal. If the LLM itself is to be ranked, the same compatibility becomes rank-relevant error. 

\paragraph{Generalizability theory separates signal from conditional advantage.}
Generalizability theory (G-theory) treats evaluation conditions as measurement \emph{facets} and decomposes score variation into their main effects and interactions \citep{cronbach_dependability_1972,shavelson_generalizability_1989,
brennan_generalizability_2001,cronbach_my_2004}. Related work shows that AI benchmark conclusions depend on task
sampling, evaluation conditions, and the model population \citep{madaan_quantifying_2024,hardy_knowledge_2026,hardy_ai_2026}. The crucial first choice is the \emph{object of measurement}. Model--scaffold compatibility is signal when selecting a deployable system, but error when ranking models independently of scaffolding. Reliability therefore belongs to an object, a generalization universe,
and an evaluation design, not to a benchmark alone.

For an object \(o\) and design \(\mathcal D\), the relative generalizability coefficient and its signal-to-noise ratio are 
\begin{equation}
E\rho_o^2(\mathcal D)
=
\frac{\sigma_o^2}
     {\sigma_o^2+\sigma_\delta^2(\mathcal D)},
\qquad
\operatorname{SNR}_o(\mathcal D)
=
\frac{\sigma_o^2}{\sigma_\delta^2(\mathcal D)}
=
\frac{E\rho_o^2}{1-E\rho_o^2}.
\label{eq:relative-reliability}
\end{equation}
Here \(\sigma_o^2\) is the variance of scores averaged over the specified universe, and \(\sigma_\delta^2\) contains variation that can change relative standing. Under \(X_o=Z_o+\delta_o\), with uncorrelated universe score and error,
\(E\rho_o^2=\operatorname{Corr}^2(X_o,Z_o)\). %This is a score-reliability coefficient, not the probability of recovering an exact ranking.

Facet main effects cancel from comparisons only when objects share the same conditions and weights. A uniformly difficult task, for example, does not change relative latent scores under a common allocation. Object-by-facet interactions do: they represent advantages that depend on the chosen task, scaffold, or benchmark. G-theory uses these components in a \emph{Decision study} (D-study) to project reliability before collecting more data.

\section{Reliability and Generalizability for Agent Leaderboards}
\label{sec:methods}

% Agent leaderboards are sparse, unbalanced, and only partially crossed: different models are evaluated on different tasks, often through different scaffolds, and most possible benchmark--model--scaffold combinations are absent.
Agent leaderboards are often sparse, imbalanced, and partially crossed: models are evaluated through different scaffolds, task counts are unequal, and many combinations are absent. 
In this section, we discuss several metrics to measure reliability in agent leaderboards given these common constraints. 
We fit a Bayesian variance decomposition to binary task outcomes, then propagate its uncertainty into reliability, design projections, and model ranks.

% Our method estimates a latent variance decomposition suited for agent evaluations with binary outcomes, then propagate posterior uncertainty into reliability, ranks, and counterfactual evaluation designs.

\subsection{Observations and Generalization Universe}
\label{sec:methods-components}

Let \(b\in\mathcal B\) index benchmarks, \(i\in\mathcal I_b\) tasks nested within benchmark \(b\), \(m\in\mathcal M\) models, and \(a\in\mathcal A\) agent scaffolds. The response
\(
y_{bima}\in\{0,1\}
\)
indicates whether model \(m\), operated through scaffold \(a\), solves task \(i\) from benchmark \(b\). Our primary object is the model \(m\). Its universe score is the component expected to persist across new tasks, benchmarks, and scaffolds resembling those represented in the leaderboard. We also consider the model--scaffold pair \((m,a)\), the relevant object when selecting a deployable system. We treat tasks as draws from benchmark-specific construction and grading processes; benchmarks as instruments drawn from a battery of contemporary agent evaluations; scaffolds as contemporary evaluation and orchestration systems; and models as the frontier-model population represented in the data. These universes delimit the inference. In particular, generalization to new benchmark-like tasks does not establish that a benchmark represents all real-world uses. %We also analyze the model--scaffold pair \((m,a)\), which is the appropriate object when users select a deployable system rather than an underlying model. High reliability for \((m,a)\) does not imply that model rankings persist after changing scaffolds.

\subsection{A Latent Variance Decomposition for Binary Outcomes}
\label{sec:methods-vardecomp}

We use Bayesian Bernoulli--logit mixed models:
% \begin{equation}
$y_{bima}\sim\operatorname{Bernoulli}(p_{bima})$,
% \qquad
$\operatorname{logit}(p_{bima})=\eta_{bima}$.
% \label{eq:bernoulli-logit}
% \end{equation}
Equivalently, \(y_{bima}=\mathbb{1}\{z_{bima}>0\}\), where
\(z_{bima}=\eta_{bima}+\varepsilon_{bima}\) and
\(\varepsilon_{bima}\sim\operatorname{Logistic}(0,1)\), giving the standard latent logistic residual variance \(\pi^2/3\). Modeling the binary responses directly avoids Gaussian approximations that are particularly misleading under floor or ceiling effects. All random effects are mean-zero Gaussian and mutually independent; for example,
\(u_m^{(M)}\sim\mathcal N(0,\sigma_M^2)\). Variance components and reliability coefficients are defined on the common latent log-odds scale (see Appendix \ref{app:latent_estimations}). From an Item Response Theory (IRT) perspective, these are random-item, many-facet Rasch models \citep{wang_exploring_2005,fox_bayesian_2001,linacre_understanding_2002}.

\subsubsection{Benchmark-level model and system reliability}
\label{sec:benchmark-model}

For each benchmark \(b\), we fit
\begin{equation}
\eta_{ima}^{(b)}
=
\beta_0^{(b)}
+u_i^{(I)}
+u_m^{(M)}
+u_a^{(A)}
+u_{im}^{(IM)}
+u_{ia}^{(IA)}
+u_{ma}^{(MA)}.
\label{eq:perbenchmark}
\end{equation}
Variance components are benchmark-specific, with the index \(b\) suppressed for readability. The decomposition separates persistent model differences from task sensitivity, scaffold differences, and model--scaffold compatibility. Because repeated observations of the same \((i,m,a)\) cell are rare, the task--model--scaffold interaction is not separately identifiable from latent response variation; we denote this terminal residual variance by \(\sigma_{IMA,e}^2=\pi^2/3\). For an equal-allocation design with \(n_i\) tasks and \(n_a\) scaffolds, model-ranking reliability is
\begin{equation}
E\rho_{M(b)}^2(n_i,n_a)
=
\frac{\sigma_M^2}
{\sigma_M^2
+\sigma_{IM}^2/n_i
+\sigma_{MA}^2/n_a
+\sigma_{IMA,e}^2/(n_i n_a)}.
\label{eq:perbench-model-reliability}
\end{equation}
Scaffold main effects do not enter the denominator because shifting every model equally does not change their ordering. In contrast, model--scaffold interactions are rank-relevant: they encode which models benefit from which scaffolds. When the object is the complete \textit{model--scaffold system},
\begin{equation}
E\rho_{MA(b)}^2(n_i)
=
\frac{\sigma_M^2+\sigma_A^2+\sigma_{MA}^2}
{\sigma_M^2+\sigma_A^2+\sigma_{MA}^2
 +(\sigma_{IM}^2+\sigma_{IA}^2+\sigma_{IMA,e}^2)/n_i}.
\label{eq:perbench-system-reliability}
\end{equation}
% The same model--scaffold variation is thus error for model-only inference but signal for system selection.

To measure whether scaffold choice preserves model ordering, we additionally define the inter-scaffold reliability for two independently sampled scaffolds evaluated on the same \(n_i\) tasks:
\begin{equation}
\rho_{AA'}^{(b)}(n_i)
=
\frac{\sigma_M^2+\sigma_{IM}^2/n_i}
{\sigma_M^2
+\sigma_{IM}^2/n_i+\sigma_{MA}^2+\sigma_{IMA,e}^2/n_i}.
\label{eq:interscaffoldrho}
\end{equation}
This is analogous to inter-rater reliability, with scaffolds acting as alternative measurement procedures. A low value means that changing the scaffold can change which model appears strongest, even when each scaffold yields internally stable scores.

\subsubsection{Persistent model differences across pooled benchmarks}
\label{sec:leaderboard-model}

Each benchmark contains too few scaffolds to estimate all scaffold-related components precisely in isolation. We therefore also fit a joint model across the nine-benchmark leaderboard:
% \begin{align}
% \eta_{bima}
% ={}&
% \beta_0
% +u_b^{(B)}
% +u_{bi}^{(I[B])}
% +u_m^{(M)}
% +u_a^{(A)}
% +u_{bm}^{(BM)}
% +u_{ba}^{(BA)}
% +u_{ma}^{(MA)}
% \nonumber\\
% &+
% u_{bim}^{(IM[B])}
% +u_{bia}^{(IA[B])}
% +u_{bma}^{(BMA)}.
% \label{eq:pooled}
% \end{align}

{\scriptsize
\begin{align}
\eta_{bima}
={}&
\beta_0
+u_b^{(B)}
+u_{bi}^{(I[B])}
+u_m^{(M)}
+u_a^{(A)}
+u_{bm}^{(BM)}
+u_{ba}^{(BA)}
+u_{ma}^{(MA)}
% \nonumber\\
+u_{bim}^{(IM[B])}
+u_{bia}^{(IA[B])}
+u_{bma}^{(BMA)}.
\label{eq:pooled}
\end{align}}
\noindent
Items are nested within benchmarks, while models and scaffolds are crossed with benchmarks wherever supported by the observed incidence graph. Shared models and scaffolds connect benchmarks and permit partial separation of their effects. 
Hierarchical priors can produce estimates for weakly supported contrasts, but cannot supply empirical identification 
% where the design provides none. The model does not identify contrasts 
between disconnected components; connectivity and estimability diagnostics are reported in Appendix~\ref{app:connectivity}.

The leaderboard universe score for model \(m\) is \(u_m^{(M)}\), the component expected to persist across sampled benchmarks, tasks, and scaffolds. Benchmark-conditioned capability \(u_{bm}^{(BM)}\) contributes to performance on benchmark \(b\), but is not assumed to transfer to a new benchmark. For a balanced design with \(n_b\) benchmarks, \(n_i\) tasks per benchmark, and \(n_a\) scaffolds,
{\scriptsize
\begin{equation}
E\rho_M^2(n_b,n_i,n_a)
=
\frac{\sigma_M^2}
{\sigma_M^2
+\sigma_{BM}^2/n_b
+\sigma_{MA}^2/n_a
+\sigma_{BMA}^2/(n_b n_a)
+\sigma_{IM[B]}^2/(n_b n_i)
+\sigma_{BIMA,e}^2/(n_b n_i n_a)}.
\label{eq:dstudy-full}
\end{equation}}
\noindent
Here \(\sigma_{BIMA,e}^2=\pi^2/3\) is terminal cell-specific variation and the logistic residual (see \S~\ref{app:olre} for separated variance). Equation~\ref{eq:dstudy-full} maps each design intervention to the uncertainty it can reduce: tasks average task-indexed error, benchmarks average benchmark-conditioned differences, and scaffolds average scaffold-conditioned differences. Table~\ref{tab:reliability-coefficients} consolidates the estimands used in the main body.

% \begin{table*}[t]
% \centering
% \small
% \caption{Reliability estimands. Each row applies
% \(E\rho_o^2=\sigma_o^2/(\sigma_o^2+\sigma_\delta^2)\), but changes the object whose ordering should generalize. The \(n_f\) terms denote equal allocation over facet \(f\).}
% \label{tab:reliability-coefficients}
% \begin{tabular}{llll}
% \toprule
% \textbf{Estimand} & \textbf{Object and inferential question}
% & \(\boldsymbol{\sigma_o^2}\) & \(\boldsymbol{\sigma_\delta^2(\mathcal D)}\) \\
% \midrule
% \(E\rho_{M(b)}^2\)
% & Does benchmark \(b\) preserve model order?
% & \(\sigma_M^2\)
% & \(\sigma_{IM}^2/n_i+\sigma_{MA}^2/n_a
%   +\sigma_{IMA,e}^2/(n_i n_a)\) \\[2pt]

% \(E\rho_{MA(b)}^2\)
% & Does benchmark \(b\) preserve system order?
% & \(\sigma_M^2+\sigma_A^2+\sigma_{MA}^2\)
% & \((\sigma_{IM}^2+\sigma_{IA}^2+\sigma_{IMA,e}^2)/n_i\) \\[2pt]

% \(\rho_{AA'}^{(b)}\)
% & Do alternative scaffolds induce the same model order?
% & \(\sigma_M^2+\sigma_{IM}^2/n_i\)
% & \(\sigma_{MA}^2+\sigma_{IMA,e}^2/n_i\) \\[2pt]

% \(E\rho_M^2\)
% & Does the leaderboard preserve model order?
% & \(\sigma_M^2\)
% & \(\sigma_{BM}^2/n_b+\sigma_{MA}^2/n_a
% +\sigma_{BMA}^2/(n_b n_a)
% +\sigma_{IM[B]}^2/(n_b n_i)
% +\sigma_{BIMA,e}^2/(n_b n_i n_a)\) \\
% \bottomrule
% \end{tabular}
% \end{table*}

\begin{table}[t]
\centering
% \small
% \footnotesize
\renewcommand{\arraystretch}{0.2}
\caption{Reliability estimands. Each row applies
\(E\rho_o^2=\sigma_o^2/(\sigma_o^2+\sigma_\delta^2)\), \(\text{SNR}_o=\sigma_o^2/\sigma_\delta^2\), but changes the object whose ordering should generalize. The \(n_f\) terms denote equal allocation over facet \(f\).}
\label{tab:reliability-coefficients}
\resizebox{\ifdim\width>\linewidth\linewidth\else\width\fi}{!}{
\begin{tabular}{llccc}
\toprule
\textbf{Estimand} 
% & \textbf{Level}
& \textbf{Object of inference} & $\boldsymbol{\sigma_o^2}$ & $\boldsymbol{\sigma_\delta^2(\mathcal{D})}$ & Eq. \\
\midrule
\makecell[l]{$E\rho^2_{M(b)}$, \\$\text{SNR}_{M(b)}$} 
% & Benchmark 
& \makecell[l]{Does benchmark \(b\)\\ preserve model order?}
& $\sigma_M^2$
& 
    $\dfrac{\sigma_{IM}^2}{n_i}+\dfrac{\sigma_{MA}^2}{n_a}+\dfrac{\sigma_{IMA,e}^2}{n_i n_a}$
& (\ref{eq:perbench-model-reliability}) \\
% & \tableeqnum\label{eq:perbench-model-reliability} \\
% \addlinespace
$E\rho^2_{MA(b)}$ 
% & Benchmark 
& \makecell[l]{Does benchmark \(b\)\\ preserve system order?}
& $\sigma_M^2+\sigma_A^2+\sigma_{MA}^2$
& 
    $\dfrac{\sigma_{IM}^2+\sigma_{IA}^2+\sigma_{IMA,e}^2}{n_i}$
& (\ref{eq:perbench-system-reliability}) \\
% & \tableeqnum\label{eq:perbench-system-reliability} \\
\addlinespace
$\rho^{(b)}_{AA'}$ 
% & Benchmark 
& \makecell[l]{Do scaffolds induce \\ the same model order?}
& $\sigma_M^2+\dfrac{\sigma_{IM}^2}{n_i}$
& $\sigma_{MA}^2+\dfrac{\sigma_{IMA,e}^2}{n_i}$ 
% & \tableeqnum\label{eq:interscaffoldrho} \\
& (\ref{eq:interscaffoldrho}) \\
\addlinespace
\makecell[l]{$E\rho^2_{M}$, $\text{SNR}_{M}$} 
% & Leaderboard 
& \makecell[l]{Does the leaderboard \\ preserve model order?}
& $\sigma_M^2$
& $\dfrac{\sigma_{BM}^2}{n_b}+\dfrac{\sigma_{MA}^2}{n_a}+\dfrac{\sigma_{BMA}^2}{n_b n_a}+\dfrac{\sigma_{IM[B]}^2}{n_b n_i}+\dfrac{\sigma_{BIMA,e}^2}{n_b n_i n_a}$ 
% & \tableeqnum\label{eq:dstudy-full} \\
& (\ref{eq:dstudy-full}) \\

\bottomrule
\end{tabular}
}
\end{table}

\subsection{What Additional Evaluations Can Resolve}
\label{sec:methods-design}

\begin{proposition}[Facet-specific replication and reliability ceilings]
\label{prop:rel-ceiling}
For %nonnegative 
variance components, the reliability in
Equations~\ref{eq:perbench-model-reliability} and~\ref{eq:dstudy-full}
is non-decreasing in each sample size. Holding \(n_b\) and \(n_a\) fixed,
{\small 
\begin{equation}
\lim_{n_i\rightarrow\infty}E\rho_{M(b)}^2(n_i,n_a)
=
\frac{\sigma_M^2}
{\sigma_M^2
+\frac{\sigma_{MA}^2}{n_a}},
\;
\lim_{n_i\rightarrow\infty}E\rho_M^2(n_b,n_i,n_a)
=
\frac{\sigma_M^2}
{\sigma_M^2
+\frac{\sigma_{BM}^2}{n_b}
+\frac{\sigma_{MA}^2}{n_a}
+\frac{\sigma_{BMA}^2}{n_b n_a}}.
\label{eq:task-ceiling}
\end{equation}} 
% \begin{align}
% \lim_{n_i\to\infty}E\rho_{M(b)}^2
% &=
% \frac{\sigma_M^2}
% {\sigma_M^2+\sigma_{MA}^2/n_a},
% \\
% \lim_{n_i\to\infty}E\rho_M^2
% &=
% \frac{\sigma_M^2}
% {\sigma_M^2+\sigma_{BM}^2/n_b
% +\sigma_{MA}^2/n_a
% +\sigma_{BMA}^2/(n_b n_a)}.
% \end{align}
\end{proposition}

\begin{proof}
Increasing \(n_i\) monotonically decreases only denominator terms indexed by tasks. Those terms converge to zero, while benchmark- and scaffold-indexed terms remain.
\end{proof}

% \begin{proposition}[Task replication has a structural ceiling]
% \label{prop:rel-ceiling}
% Assume \(\sigma_M^2>0\) and nonnegative variance components.
% Reliability in Equations~\ref{eq:perbench-model-reliability}
% and~\ref{eq:dstudy-full} is non-decreasing in each design size.
% Holding benchmark and scaffold coverage fixed,
% \begin{align}
% \lim_{n_i\to\infty}E\rho_{M(b)}^2
% &=
% \frac{\sigma_M^2}
% {\sigma_M^2+\sigma_{MA}^2/n_a},
% \label{eq:task-ceiling}\\
% \lim_{n_i\to\infty}E\rho_M^2
% &=
% \frac{\sigma_M^2}
% {\sigma_M^2+\sigma_{BM}^2/n_b
% +\sigma_{MA}^2/n_a+\sigma_{BMA}^2/(n_b n_a)}.
% \label{eq:pooled-task-ceiling}
% \end{align}
% \end{proposition}

% \begin{proof}
% Increasing a design size decreases only error terms indexed by that facet. As \(n_i\to\infty\), task-indexed terms vanish; benchmark- and scaffold-indexed terms remain.
% \end{proof}

% \noindent
Thus, arbitrarily many tasks cannot overcome model--benchmark heterogeneity or model--scaffold coupling. This result motivates evaluating diversity, rather than task count alone, as a design resource.

% The ceiling is a property of the intended generalization, not necessarily a failure to estimate the observed system accurately. More tasks can precisely establish an advantage that remains conditional on one benchmark or scaffold.

\begin{corollary}[Benchmark breadth at a fixed task budget]
\label{cor:fixed-budget}
Fix \(n_a\) and the total number \(N=n_b n_i\) of tasks per scaffold.
Under the balanced, exchangeable-facet model,
\begin{align}
\sigma_\delta^2
={}&
\frac{\sigma_{MA}^2}{n_a}
+\frac{\sigma_{BM}^2+\sigma_{BMA}^2/n_a}{n_b}+
% \nonumber\\
% &+
\frac{\sigma_{IM[B]}^2+\sigma_{BIMA,e}^2/n_a}{N}.
\label{eq:fixed-budget}
\end{align}
% Distributing the same task budget across benchmarks
% increases reliability when \(\sigma_{BM}^2+\sigma_{BMA}^2/n_a>0\).
Thus, distributing the same task budget across more benchmarks
increases reliability whenever \(\sigma_{BM}^2+\sigma_{BMA}^2/n_a>0\).
\end{corollary}

\begin{proof}
Use \(n_i=N/n_b\) in
Eq. ~\ref{eq:dstudy-full}.
Only the benchmark-conditioned term changes with \(n_b\).
\end{proof}

Breadth does not create additional persistent model signal; it reduces contamination by condition-specific advantages.
The corollary assumes similarly informative benchmark draws, adequate crossing, and no additional benchmark setup cost. It does not imply that arbitrary new benchmarks outperform more tasks, or that benchmarks always offer greater value than scaffolds.

\subsection{Bayesian Estimation and Design Studies}
\label{sec:methods-estimation}
We use Bayesian estimation with regularizing priors because sparse crossed designs can yield unstable variance estimates, especially near floor and ceiling performance. For every posterior draw, we compute reliability, SNR, and the task-only ceilings. D-study projections therefore retain uncertainty in the variance decomposition rather than substituting point estimates. Partial pooling allows connected
observations to inform common variance components while retaining uncertainty where scaffold or cross-benchmark replication is limited.

Primary D-studies use common equal allocations; analyses reproducing the observed imbalance appear in
Appendix~\ref{app:methods-details}. We measure evaluation volume in model--scaffold--task trials and use dashboard prices for dollar-cost
projections. Repeated task subsampling checks whether reduced designs retain the ranking information predicted by the D-study (see \S~\ref{app:subsampling}).

\subsection{Posterior Capability, Ranks, and Transportability}
\label{sec:methods-ranks}

For benchmark \(b\), the scaffold-marginalized latent capability of model \(m\) in posterior draw \(s\) is
% \begin{equation}
$\theta_{mb}^{(s)}
=
u_m^{(M,s)}+u_{bm}^{(BM,s)}$.
% \end{equation}
% Ranking models within each draw yields posterior rank distributions and pairwise ordering probabilities
% \(
% \Pr(\theta_{mb}>\theta_{m'b}\mid y).
% \)
% These quantities distinguish an estimated ordering from evidence that two models are meaningfully distinguishable. We compare posterior and published rankings using Spearman correlation and ranked unbiased squared distance correlation, \(\operatorname{dCor}_n^2\); the latter remains informative in the presence of extensive ties.
To test whether the estimated shared model component is transportable rather than an artifact of the nine-benchmark panel, we evaluate \(\widehat\theta_m =u_m^{(M)}\) on four contemporaneous benchmarks excluded from model estimation. Among overlapping models, we compare its association with external performance against that of the conventional in-panel mean-accuracy aggregate. This is an out-of-panel test of convergent predictive validity, not evidence of universal deployment validity nor internal construct validity. Finally, we assess sensitivity to the link function, estimation method, variance-component estimator, prior specification, and leave-one-benchmark-out refits. The latter analysis also identifies benchmarks that disproportionately contribute signal or connectivity to the leaderboard. 
Appendices have %HAL and external dataset details (\S~\ref{app:data}), 
full computational estimation details (\S~\ref{app:methods-details}), robustness checks and sensitivity analyses (\S~\ref{app:robustness}), %connectivity and identifiability analyses (\S~\ref{app:connectivity}), 
and methodological comparisons (\S~\ref{app:method_comps}).
% Appendix~\ref{app:robustness} has full computational and robustness details, and Appendix \ref{app:rank_stability} has rank stability estimates.% appear in Appendix~\ref{app:robustness}.

% \paragraph{External evidence for the shared model component.}
% We test whether the component treated as model signal corresponds
% to recurring performance outside the fitted benchmark panel.
% Among overlapping LLMs, we compare the posterior-median shared
% model effect with an unweighted mean-score baseline on four
% external agent benchmarks. For SWE-bench Verified and
% \(\tau^2\)-bench, we exclude the corresponding internal benchmark
% before estimating either score.

% The benchmark-specific comparisons average across the included
% SWE-bench Verified leaderboards. A complementary multilevel
% analysis retains those leaderboard-specific observations and
% includes benchmark and leaderboard random effects, yielding
% \(52\) model--source observations. These are repeated observations,
% not \(52\) distinct models. Full aggregation and dependence
% specifications appear in Appendix~\ref{app:external_valid_data}.

\section{Data}
\label{sec:data}
We analyze agent rollouts from all nine benchmarks distributed via HAL~\cite{kapoor_ai_2024}: AssistantBench~\citep{yoran2024assistantbench}, CoreBench Hard~\citep{siegelCOREBenchFosteringCredibility2024}, GAIA~\citep{mialonGAIABenchmarkGeneral2023}, Online-Mind2Web \citep{xue_illusion_2025}, SciCode~\citep{tianSciCodeResearchCoding2024}, ScienceAgentBench \citep{chen2024scienceagentbench}, SWE-bench Verified Mini~\citep{jimenezSWEbenchCanLanguage2024a},
$\tau$-bench Airline~\citep{yao$t$benchBenchmarkToolAgentUser2024}, and USACO~\citep{shiCanLanguageModels2024}, covering web navigation, scientific programming tasks, multi-step and user assistance, software engineering, customer-service interaction, and competitive programming (details in Appendix~\ref{app:hal_data}). From each of AssistantBench, GAIA, SciCode and SWE-bench Verified Mini, the HAL leaderboard uses a subset of tasks from the full benchmark \citep{kapoor_holistic_2025}.

HAL dataset covers $29{,}923$ total agent rollouts across 54 models and 13 scaffolds; 9 models appear on every benchmark each of which span at least 69\% of the available scaffolds. Each rollout is over a single task, model, reasoning-effort, and scaffold. Summary statistics are found in Table \ref{tab:bench-summary}. All outcomes are task-level binary scores.
% Because our analysis is mostly interested in the relationship between the model-variant, scaffold and benchmarks, we join the reasoning-effort with the model to be our model-facet. 
% For our analysis, following the leaderboard construction, each time ``model'' is stated it refers to the model at a specific-reasoning effort (i.e. `gpt-5high' and `gpt-5minimal' are separated). 
% Task counts range from $33$ (AssistantBench) to $307$ (USACO), model-scaffold counts from $14$ to $56$, and each benchmark has $2$ or $3$ distinct scaffolds. % ($\text{score}_{ijk} \in \{0,1\}$). 
%The following benchmarks use a subset of tasks from the full benchmark, based on the available on the HAL leaderboard~\cite{kapoor_holistic_2025}: AssistantBench, GAIA, SciCode and SWE-bench Verified Mini. Table~\ref{tab:bench-summary} summarizes the design. 
%The data contain 54 LLM models, of which nine appear on every benchmark and use at least 69\% of the 13 scaffolds. 
The incidence structure is incomplete at every level, a common problem in leaderboards \citep{singh_leaderboard_2026}. One scaffold is shared across eight benchmarks; AssistantBench connects the remaining benchmark through an additional shared scaffold; all other benchmarks contain at least one additional benchmark-specific scaffold. Models and scaffolds are neither fully crossed nor evenly replicated. Nevertheless, shared models and scaffolds connect the observation graph, permitting partial separation of model, benchmark, and scaffold effects. 

Additionally, to corroborate our claims and provide external convergent validity, we use two supplemental datasets. The Harbor Index dataset is a meta-benchmark containing task level scores from 29 agent benchmarks for 9 LLMs and 4 scaffolds. The second dataset consists of LLM-level scores for four benchmarks that are contemporaneous with and include LLMs specified in the HAL data. Appendices more detail for the datasets (\S~\ref{app:data}) and
% full computational estimation details (\S~\ref{app:methods-details}), robustness checks and sensitivity analyses (\S~\ref{app:robustness}), 
data connectivity analysis (\S~\ref{app:connectivity}). %, and methodological comparisons (\S~\ref{app:method_comps}).
% For the analyses of this study, Appendix~\ref{app:robustness} reports robustness and ablation analyses and Appendix \ref{app:connectivity} the corresponding identifiability of the estimations given the characteristics of the data.
These data represent current practice: because agent evaluations are expensive,%, these conditions (different models, scaffolds, and tasks) are often sparsely observed in composite leaderboards: 
many model--scaffold--benchmark combinations absent entirely.

% , propagating error appropriately

\begin{table}[t]
\centering
\caption{HAL dataset design summary. Counts denote unique levels of each facet.}\label{tab:bench-summary}
\tiny
\begin{tabular}{lccccc}
\toprule
Benchmark & Models & Tasks & Agent Scaffolds & Model--Scaffold Pairs \\
\midrule
AssistantBench   & 18 &  33 & 2 & 30  \\
CORE-Bench Hard   & 34 &  45 & 3 & 56 \\
GAIA             & 21 & 165 & 2 & 34 \\
Online-Mind2Web & 13 & 300 & 2 & 23  \\
SciCode          & 17 &  65 & 3 & 37 \\
ScienceAgentBench & 19 & 102 & 2 & 25 \\
SWE-bench Verified Mini & 24 &  50 & 2 & 26 \\
$\tau$-bench Airline & 23 & 50 & 3 & 42\\
USACO            & 13 & 307 & 2 & 14 \\
\bottomrule
\end{tabular}
\end{table}

\section{Results and Practical Recommendations}
\label{sec:results}

% We ask three questions. First, do agent benchmarks support rankings of base models, or only of particular model--scaffold systems? Second, which additions to an evaluation design---tasks, benchmarks, or scaffolds---most improve reliability? Third, does reliability adjustment recover a model ordering that transports beyond the benchmarks used to estimate it? Together
%Responding to our three questions, the results show that current leaderboards often measure fixed systems reproducibly but isolate persistent model differences imprecisely.
The results identify a structural limitation of task-only scaling:
current evaluations can distinguish fixed systems precisely while
leaving persistent model differences unresolved. The useful response
is to change the measurement design, not merely enlarge it.

\subsection{Reliability depends on the measurement goal}

% There is no single answer to whether an evaluation is reliable. 
Reliability depends on the measurement claim a practitioner wants to make. In agent evaluations, two choices are relevant here: whether the object of measurement is the underlying model or the model--scaffold system, and whether the goal is relative (rank) or absolute interpretation of scores. % to compare systems or interpret their absolute performance. 

\begin{figure}[h!]
    \centering
    \includegraphics[width=\linewidth]{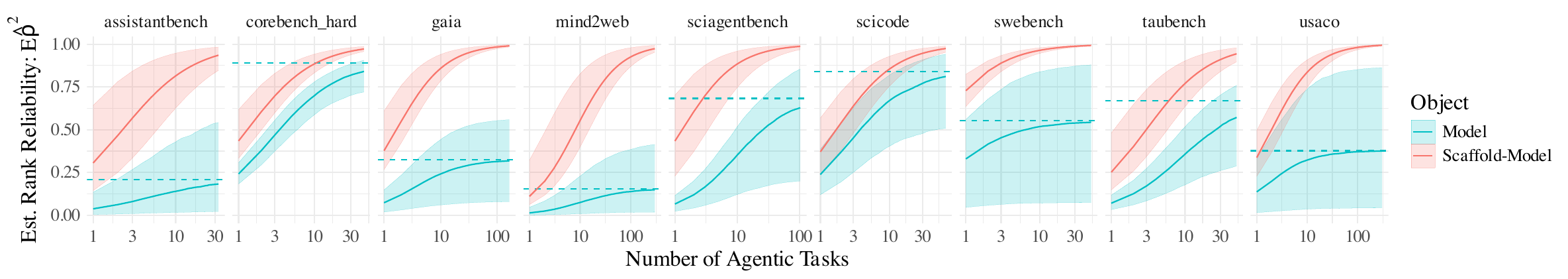}
    \caption{\textbf{Reliability depends on the object being ranked \& task replication cannot remove scaffold-dependent model advantages.}
    D-study reliability as tasks are added for base models and fixed model--scaffold systems. Dashed lines are task-only asymptotes for model rankings. Estimates are posterior medians with \(68\%\) standard error HDIs. See also Table \ref{tab:bench_stats}.}
    \label{fig:mod_v_agent_cis}
\end{figure}

\paragraph{Model and system reliability can differ substantially.} Across the nine benchmarks, estimated system reliability (i.e., ranking model--scaffold pairs) is \(E\rho_{MA(b)}^2\in[0.935,0.994]\), whereas model reliability is \(E\rho_{M(b)}^2\in[0.148,0.841]\) (Figure \ref{fig:mod_v_agent_cis}). These are not conflicting assessments of the same score. System reliability counts scaffold differences and compatibility as signal; model reliability requires differences that persist after averaging over scaffolds. A leaderboard can therefore be reliable for ranking model-scaffold pairs but unreliable for solely ranking the underlying model.

\begin{comment}
\begin{figure}[]
    \centering
    \includegraphics[width=\linewidth]{figs/gbayes_rel_limit_SE_per_bench_all.pdf}
    \caption{\textbf{Task replication cannot remove scaffold-dependent
    model advantages.}
    D-study reliability for base models and fixed model--scaffold systems as task counts increase. Dashed lines show task-only ceilings. Estimates are posterior medians with \(68\%\) HDIs.}
    \label{fig:mod_v_agent_cis}
\end{figure}
\end{comment}

%Posterior mean inter-scaffold reliability ranges from \(0.205\) on OnlineMind2Web to \(0.828\) on CORE-Bench Hard (Figure~\ref{fig:interscaffold}). Thus, scaffold dependence varies substantially across benchmarks; it is not a uniform property of an agent evaluation. At the low end, using the same models and tasks does not ensure comparable relative scores after changing the scaffold.

% \paragraph{Reliable \textit{rankings} do not imply reliable \textit{scores}.} Relative reliability (\(E\rho^2\)) asks whether the ranking of models stays similar across evaluation conditions, while absolute reliability (\(\Phi\)) asks whether their reported scores stay similar. Figure~\ref{fig:absolute-relative} shows that this distinction matters substantially in practice: absolute reliability is consistently lower across the nine benchmarks, and for several benchmarks remains near zero even as ranking reliability increases with additional tasks. An evaluation can therefore support a stable ranking without supporting equally strong claims about a model's absolute score or whether it exceeds a fixed threshold.

\paragraph{Recommendation.} Specify whether the intended object of measure is the model or a model--scaffold system, especially in leaderboards using multiple agentic benchmarks, and whether the intended interpretation is rank comparison (or whether absolute scores are intended to carry meaning, such as capability thresholds.\footnote{There may be situations where reliability of actual numeric scores (not just ranking) is needed. We illustrate this separate estimation on the observed scale in \S~\ref{app:abs_reliab}, which shows that score reliability is substantially lower for all benchmarks. Reliable \textit{rankings} do not imply reliable \textit{scores}.} 
% ; see \S~\ref{app:abs_reliab} for estimation procedure on observed scale). 
Provide rankings, calculate scores, and estimate reliability for that specific claim rather than reporting a single generic reliability statistic.

\subsection{Changing the scaffold can change evaluation conclusions}

\begin{figure}[h!]
    \centering
    \includegraphics[width=\linewidth]{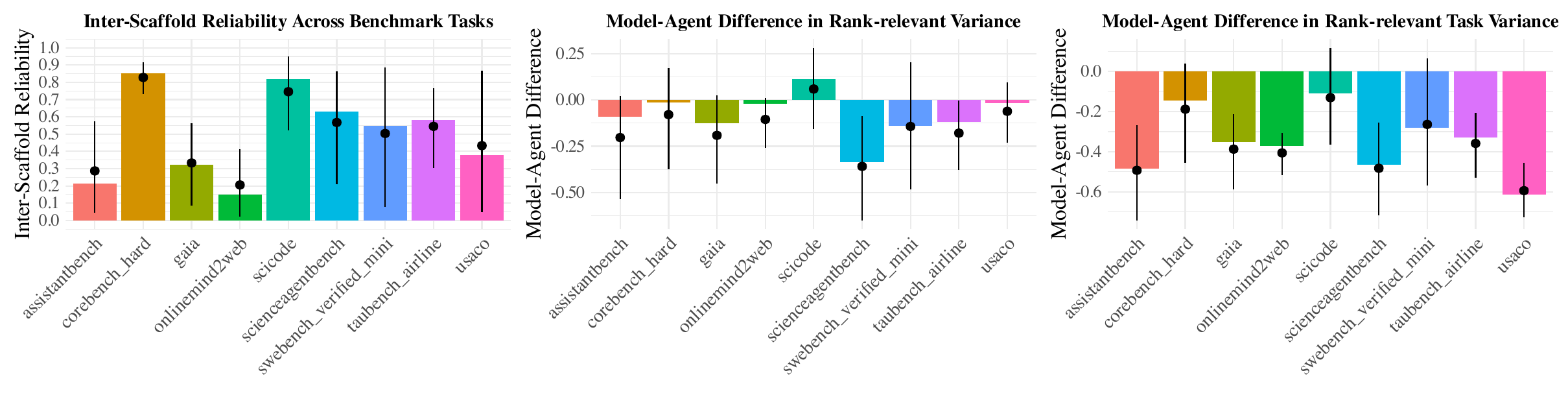}
    \caption{\textbf{Scaffolds are rank-relevant and result in distinct task-level consequences.}
    \textit{Left}: posterior inter-scaffold reliability \(\rho_{AA'}^{(b)}\).
    \textit{Middle} and \textit{Right}: model-minus-scaffold system rank-relevant variance contrasts (over denominator from Eq. \ref{eq:perbench-system-reliability}) at the benchmark and task levels, respectively; positive values favor the model contribution. Task-interaction contrasts measure
task-specific sensitivity. %, not the probability of a model-rank reversal. 
Points and bars denote posterior means and medians, respectively, with \(68\%\) standard error HDIs.}
    \label{fig:interscaffold}
\end{figure}

\paragraph{How much the scaffold matters depends on the benchmark.} Inter-scaffold reliability (Eq. \ref{eq:interscaffoldrho}) measures whether different scaffolds produce similar rankings when evaluating the same models on the same tasks ranging (with 95\% HDI) from 0.151 [0.004,0.577] on OnlineMind2Web to 0.852 [0.631,0.947] on CORE-Bench Hard (Figure~\ref{fig:interscaffold}). At the low end, changing only the scaffold can substantially change which models appear to perform best on a benchmark, even when the models and tasks remain fixed. %[Add absolute inter-scaffold reliability ref to appendix.]

\paragraph{Scaffolds can change which tasks get solved without improving the overall benchmark score.} 
The pooled decomposition (Eq. \ref{eq:pooled}) distinguishes persistent differences from task-specific sensitivity. In the pooled analysis, the contrast \(\sigma_M^2-\sigma_A^2\) favors neither direction (posterior directional probability approximately \(50\%\)) \citep{makowski_indices_2019}. However, task-specific variation across scaffolds is larger than task-specific variation across models with posterior probability \( \Pr\!( \sigma_{IA[B]}^2>\sigma_{IM[B]}^2 \mid y )=84\%\) (see Figure~\ref{fig:vardiff_levels}). Benchmark specific decompositions (Eq. \ref{eq:perbenchmark}) show the same pattern (Figure \ref{fig:interscaffold}, middle and right) %, although the (95\%) HDI includes equality (Figure~\ref{fig:vardiff_levels}). 
This suggests that scaffold choice can strongly affect success on individual tasks, even without making one scaffold producing uniformly better on a benchmark as a whole. Similarly, A scaffold can change which tasks a system solves without producing a uniformly stronger system.

\paragraph{Recommendation.} If the goal is to evaluate the underlying model, test the same models across multiple scaffolds rather than relying on a single implementation. Report how much conclusions change across scaffolds, and avoid attributing scaffold-specific advantages to the model itself. If the deployed model--scaffold system is the intended object of evaluation, scaffold variation can instead be treated as part of the system being measured.

\subsection{More tasks cannot always resolve evaluation uncertainty}
%\subsection{More Tasks Leave Most Model-Ranking Error Intact}
\label{sec:results-task-ceiling}

\paragraph{More tasks only help when the main uncertainty comes from differences across tasks.} Adding tasks helps when a model’s measured performance changes substantially depending on which tasks it is tested on (Proposition~\ref{prop:rel-ceiling}). It does not fix uncertainty caused by other facets, such as the choice of scaffold. In our data, even infinitely many similarly constructed tasks would improve model-ranking reliability by at most \(0.10\). Only CORE-Bench Hard and SciCode can exceed \(E\rho^2=0.75\) through task scaling alone; for OnlineMind2Web, reliability increases only from \(0.148\) to \(0.153\).

\paragraph{A benchmark can run out of useful information before it runs out of tasks.} Adding tasks repeatedly measures the same model--scaffold and benchmark-specific effects. Once these sources of uncertainty dominate, more tasks make the existing evaluation setup more precise without making the broader model claim substantially more reliable. Low reliability does not mean that a benchmark measures an unimportant capability; it means that, for the models being compared, its scores do not reliably distinguish the quantity of interest. Thus, reliability should be re-estimated as the competitor population changes.

\paragraph{Recommendation.} Estimate how much reliability can improve from adding tasks before expanding a benchmark. Add tasks when task sampling is the main source of uncertainty; otherwise, spend evaluation budget on relevant facets that drive uncertainty, such as broader scaffold coverage. 

\begin{comment}
The fitted task-only ceilings explain why high system reliability coexists with uncertain model rankings
(Figure~\ref{fig:mod_v_agent_cis}). Under the observed scaffold coverage, adding infinitely many similarly constructed tasks improves estimated model reliability by at most approximately \(0.10\). Only CORE-Bench Hard and SciCode have task-only ceilings above \(E\rho^2=0.75\). For OnlineMind2Web, for example, estimated reliability changes
from \(0.148\) to a ceiling of \(0.153\).

Proposition~\ref{prop:rel-ceiling} identifies the mechanism: additional tasks average task sensitivity but repeatedly measure the same model--scaffold compatibility. The remaining limitation is therefore not simply too few task observations. It is insufficient coverage of the conditions over which the model claim is meant to hold.

Low reliability does not imply that a benchmark lacks scientific value. It means that its scores weakly resolve persistent differences within the evaluated population. A benchmark can cover an important capability yet provide little
ranking information when nearly all frontier models succeed, fail, or perform similarly. Reliability should therefore be
re-estimated as the competitor population changes.

\paragraph{Design implication.}
Estimate the task-only ceiling before extending a benchmark. Add tasks while task-indexed error dominates; expand benchmark or scaffold coverage when the corresponding interactions set the limit.

\end{comment}

\begin{figure}[ht!]
    \centering
    \begin{subfigure}{0.30\linewidth}
        \centering
        \includegraphics[width=\linewidth]{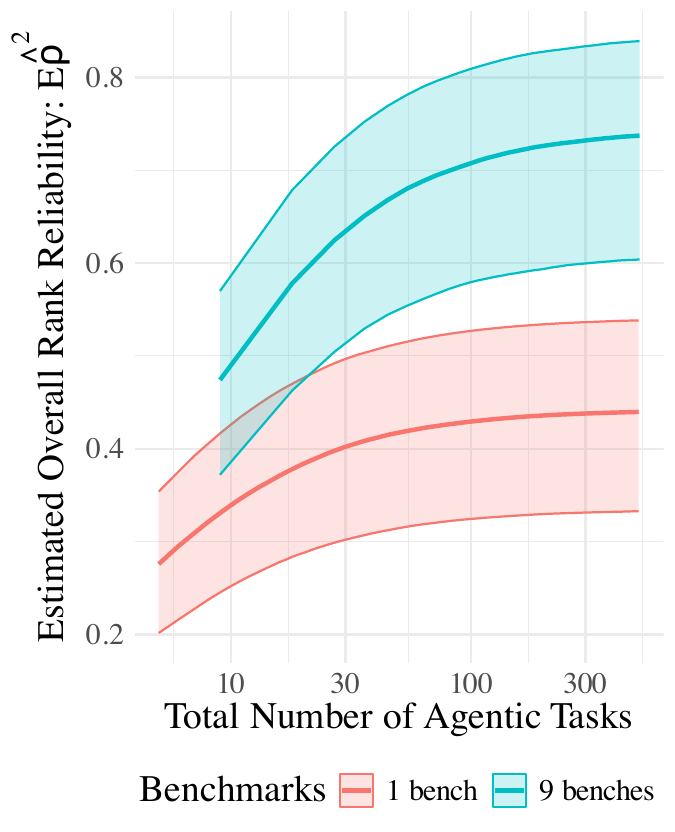}
        \caption{Model reliability.}
        \label{fig:pooled_reliability}
    \end{subfigure}
    \hfill
    \begin{subfigure}{0.69\linewidth}
        \centering
        \includegraphics[width=\linewidth]{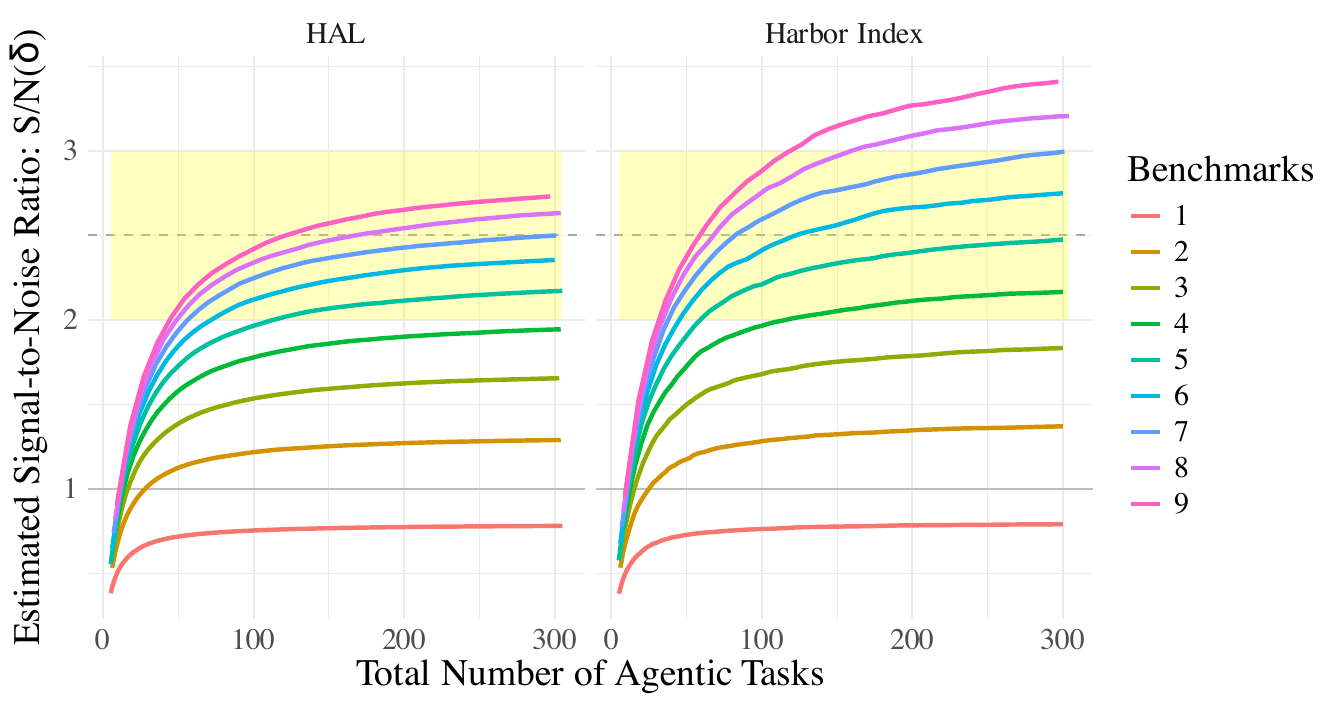}
        \caption{Rank-relevant signal-to-noise ratio.}
        \label{fig:pooled_snr}
    \end{subfigure}
    \caption{\textbf{Broader measurement conditions reduce error that
    an increase in the number of test items alone cannot.} Pooled D-studies compare concentration within one benchmark against allocation across multiple benchmarks. Estimates are posterior medians; panel (a) includes \(68\%\) standard error HDIs. HAL and Harbor are fitted separately. The reference bands \([2,3]\) represent minimal detection limits used in laboratory measurement \citep{cder_q2r2_2024,sheehan_whats_2026,taleuzzaman_limit_2018}, and are descriptive rather than benchmark calibrated thresholds.}
    \label{fig:ep2_and_snr}
\end{figure}

\subsection{Pooling diverse benchmarks can make model rankings more reliable}\label{sec:pooling_reliability_increase}

Agentic leaderboards often consist of multiple benchmarks from which model agentic capability is to be inferred. If the goal is to rank models on their ability to perform on diverse agentic tasks, pooling benchmarks that test different kinds of tasks provides more information about which performance differences persist across settings. At a fixed task budget, benchmark breadth averages condition-specific model advantages that within-benchmark replication leaves untouched (Corollary~\ref{cor:fixed-budget}).

\begin{table}[h!]
\centering
\caption{\textbf{Agreement with external agent benchmarks.}
Kendall's \(\tau\) for the unweighted HAL mean and reliability-adjusted latent ($\widehat{\theta}$) model effect (median posterior) with \([95\%]\) CIs.}
\label{tab:external-rank-validity}
\resizebox{\linewidth}{!}{
\begin{tabular}{lrrrc}
\toprule
External Benchmark
& Mean score
& \(\widehat{\theta}\)
& Difference
& LLM Observations \\
\midrule
BFCL v4
& \(0.429\,[-0.282,\,0.836]\)
& \(\mathbf{0.714}\,[0.147,\,0.928]\)
& \(+0.285\) & 7 \\
Terminal-Bench 2.0
& \(0.524\,[-0.165,\,0.869]\)
& \(\mathbf{0.810}\,[0.361,\,0.954]\)
& \(+0.286\) & 7 \\
SWE-bench Verified
& \(0.565\,[0.306,\,0.746]\)
& \(\mathbf{0.765}\,[0.595,\,0.870]\)
& \(+0.200\) & 20 \\
\(\tau^2\)-bench Core
& \(0.333\,[-0.515,\,0.852]\)
& \(\mathbf{0.867}\,[0.383,\,0.977]\)
& \(+0.534\) & 6 \\
\midrule
Mean $\tau$
& \(0.463\)
& \(\mathbf{0.789}\)
& \(+0.326\) & --- \\
% Multilevel $\tau_{\text{partial}}$
% & \(0.430\,[ 0.266,\, 0.570]\) 
% & \(\mathbf{0.632}\,[0.507,\,0.732]\)   
% & \(+0.202\) & 52 \\
\bottomrule
\end{tabular}}
% \caption*{\footnotesize Multilevel $\tau_{\text{partial}}$ retains those leaderboard-specific observations and
% includes benchmark and leaderboard random effects, yielding
% \(52\) model--source observations. Full aggregation and
% specifications are in \S~\ref{app:external_valid_data}. }
\end{table}

% The pooled D-study projects a single-benchmark model-reliability ceiling near \(0.44\). Distributing evaluations across the nine-benchmark battery raises projected reliability to approximately \(0.75\) under the studied allocation (Figure~\ref{fig:ep2_and_snr}). This is the empirical counterpart of Corollary~\ref{cor:fixed-budget}: 

\paragraph{Pooling benchmarks can achieve more reliable model rankings at lower cost.}
At the same task budget, distributing evaluations across the nine-benchmark battery raises projected model-ranking reliability from approximately 0.44
% [0.27,0.60] 
for a single benchmark to 0.75 
% [0.52,0.89] 
(Figure~\ref{fig:ep2_and_snr}). A complementary pooled analysis of the Harbor Index \citep{shi_harbor_2026} (Appendix~\ref{app:harborindex}) supports the same qualitative pattern. Moreover, the full HAL battery costs more than (\$47{,}000), while a balanced allocation achieves comparable reliability for approximately (\$19{,}000). For an illustrative target of ($SNR = 2.5, E\rho^2\approx0.71$), approximately (14) tasks per benchmark cost (\$8{,}144), an estimated (83\%) reduction. In Appendix Table~\ref{tab:k_subsamples}, we corroborate the diminishing returns using task subsampling.

\paragraph{Task diversity leads to better model rank generalization.} 
% If the goal is to rank models on their ability to perform on diverse agentic tasks, pooling benchmarks that test different kinds of tasks provides more information about which performance differences persist across settings. 
% At the same task budget, 
% projected model-ranking reliability increases from approximately \(0.44\) within a single benchmark to \(0.75\) across the nine-benchmark battery (Figure~\ref{fig:ep2_and_snr}). The resulting shared 
Model effect estimates $\hat\theta$ also agree more closely with all four held-out agent benchmarks than a simple average of benchmark scores (mean Kendall's \(\tau\): \(0.789\) vs.\ \(0.463\); Table~\ref{tab:external-rank-validity}; see \S~\ref{app:external_valid_data} for data and contamination prevention). Together, these results suggest that pooling diverse benchmarks can help separate model differences that recur across settings from advantages specific to a particular benchmark or evaluation condition. These results provide convergent predictive evidence, not proof of a universal one-dimensional agent capability.\footnote{Internally, the pooled estimation strongly outperforms per-benchmark estimations in correlations of posterior rankings with benchmark-wise observed scores, [0.62,0.96] and [0.08,0.25] respectively (Table \ref{tab:rank-robustness}, \S~\ref{app:ranking-sensitivity}). Yet posterior rank intervals remain wide; many neighboring models have ordering probabilities near \(1/2\) (Figure~\ref{fig:pr_theta_dominance}), with some models moving across rank quartiles between reported and adjusted rankings (Figures~\ref{fig:old_new_rank_apx} and~\ref{fig:rank_instability_appendix}). 
}

\paragraph{Breadth is most useful when the design is connected with informative benchmarks.} For these data, CORE-Bench Hard contains the greatest SNR (see Figure \ref{fig:mod_snr}, \S~\ref{app:loo-benchmark}), particularly with its own scaffold, CORE Agent (\S~\ref{app:loo-benchmark-agent}). Removing this strongest discriminating signal increases pooled uncertainty more than any other LOO removals. Thus, having a benchmark with high model signal can support anchoring a common scale through sufficient connectivity to other benchmarks.

\paragraph{Recommendation.}
When the intended claim concerns performance across different kinds of agentic tasks, pool benchmarks that represent that range rather than evaluating each benchmark exhaustively. Use the reliability analysis to determine how much evaluation is needed within each benchmark while preserving enough task diversity to support the broader cross-task claim. For imbalanced designs, prioritize identifying informative anchors to bridge conditions.

\section{Conclusion}\label{sec:conclusion}

Agent evaluations should not be treated as having a single, intrinsic level of reliability. Reliability depends on what practitioners want to learn from them: the underlying model or a complete model--scaffold system, a ranking or an absolute score, and performance on one benchmark or across a broader range of agentic tasks. Our results show that these distinctions matter in practice and hence that evaluation design should follow the intended claim: Practitioners should first specify what they want a score or ranking to mean, then identify which sources of uncertainty prevent the evaluation from supporting that claim. More evaluation is useful only when it addresses those sources. Reliability analysis can therefore guide not only how evaluation results are interpreted, but also where additional evaluation effort is most valuable for differentiating a target model population.

\subsection*{Reproducibility statement}
To facilitate reproducibility, the code and data are available online,\footnote{\url{https://github.com/hardy-education/scaffold_eval}} original data sources are linked in Appendix \ref{app:data}, and computation and estimation details are found in Appendix \ref{app:methods-details}.

\subsection*{AI use statement}\label{app:ai_use}

AI was used during the writing phase of this study. GPT 5.6 Sol was used after initial drafts to reduce the text length of several sections in the main body, all of which required further editing after use. Paragraphs describing tabular results in Appendices \ref{app:loo-benchmark}--\ref{app:loo-model} were revised using GPT 5.5. Drafts of several sections of Appendix \ref{app:connectivity} were created from research notes using GPT 5.5, which were revised and then rewritten using human-edited combinations of text from GPT 5.5, GPT 5.6 Sol, and Claude Opus 4.8. Finally, the 100\% human-written code used in this study was refactored with expanded comments and validated against paper findings (under subsampling) to support reproducibility using Claude Opus 4.8 with Claude Code.

\bibliographystyle{iclr2027_conference}
\bibliography{references.bib, ruhana.bib}

\clearpage

\appendix

\section*{Appendix}

\section*{Limitations and Future Work}
\label{sec:limitations}

\paragraph{Reliability is necessary but not sufficient for valid evaluation.} The bounds we estimate describe how consistently a benchmark differentiates the objects it ranks, not whether the rankings correspond to underlying capability or to any external ground truth: a benchmark can be reliable but invalid. The object-of-measurement problem we document is in this sense a construct validity problem in disguise, since reliability against an undefined construct cannot be assessed in principle~\citep{cronbach_construct_1955, salaudeen_measurement_2025}. 
% Our bounds are also within-benchmark: whether any object of measurement supports generalization \emph{across} benchmarks, e.g., whether a model that ranks first on SWE-bench Verified Mini also ranks first on $\tau$-bench Airline, is a separate validity question our framework does not directly answer. 
Future work should complement the reliability framework with validity studies, including external-criterion validation and cross-benchmark transfer analyses.

\paragraph{Scope conditions on the headline ceiling.} The reliability bounds we report are joint properties of the benchmark, the scaffold sample, the task sample, and the competitor set being ranked. A clustered frontier-model pool depresses $\sigma^2_M$ regardless of benchmark quality, lowering reliability not because the benchmark is poorly designed but because it is being used to distinguish systems too close for its resolution; hence, the  ceiling we find should be read as a statement about current leaderboards for the current frontier-model competitor set. Scaffolds are selected artifacts, often co-developed with the benchmarks they evaluate, and per-benchmark counts ($n_a \in \{2, 3\}$) reflect standard practice across the field rather than a feature of HAL; pooling across benchmarks in Equation \ref{eq:pooled} partially addresses but per-benchmark scaffold-variance claims remain weaker, and pooling itself depends on \textsc{HAL-Generalist} as the cross-benchmark harness anchor (Appendix~\ref{app:connectivity}). Future work should construct evaluation datasets that systematically span larger scaffold libraries, and should track how the ceiling shifts as the model pool evolves.

\paragraph{Scope of Agent Scaffolds and Systems.} These results are conditional on the sampled model pool
and scaffold set. A compressed frontier depresses $\sigma_M^2$ and thus reliability
independent of benchmark quality; a richer scaffold sample would sharpen the model--scaffold
decomposition. Our claims are therefore about \emph{these leaderboards for this competitor
set}, and generalize to new models and scaffolds only insofar as the sampled conditions
represent the intended universe. We view stating this scope explicitly as part of reliable
practice rather than a caveat to it. Inference remains conditional on the represented evaluation population; partial pooling does not by itself correct selective evaluation or reporting.

\paragraph{Latent-scale reliability.} We estimate reliability on the latent logit scale, while leaderboards report observed proportions. The two scales do not translate one-to-one, so our numbers should be read as approximate bounds on observed-scale rank stability. Future work could entail a sensitivity analysis comparing the scales.

%\paragraph{Latent-scale vs.\ observed-scale reliability.} We estimate reliability on the latent logit scale, the natural scale for variance decomposition under a Bernoulli likelihood, while leaderboards report observed proportions. Variance components on the logit scale do not translate one-to-one into variance in percent-correct scores, and a reliability of $0.45$ on the logit scale need not correspond to the same level of rank instability one would compute from observed-scale variance. Readers interpreting our numbers as bounds on observed-scale rank stability should treat the correspondence as approximate; a sensitivity analysis comparing the two scales is a useful extension.

\paragraph{Raising reliability bounds.} The empirical scope of this work is the characterization of reliability bounds rather than their displacement. Validating interventions that raise the ceiling is the subject of subsequent work, since each candidate intervention introduces its own measurement-design tradeoffs. Candidate interventions include task selection using item discrimination methods, scaffold sampling under a defined scaffold family, and task-quality auditing; empirical evaluation of these, alongside methods that reduce the measurement cost of the recommendations, are the most direct extensions.
 
\section{Dataset Descriptions}\label{app:data}

\subsection{HAL Dataset}\label{app:hal_data}
The Holistic Agent Leaderboard (HAL) dataset\footnote{\url{https://hal.cs.princeton.edu/}} is a standardized, cost-aware, and third-party evaluation platform and dataset initiative developed by the SAgE (Science of Agent Evaluation) research group at Princeton University. The formatted and cleaned tasks-level of this dataset is publicly released \footnote{\url{https://huggingface.co/datasets/razam2/hal-response-matrix}}.

\subsubsection{Benchmarks}\label{app:benchmarks}

\paragraph{AssistantBench}
Tasks consist of time-consuming, busy-work tasks that an average person may face, seeking information from the web.

\begin{tcolorbox}[examplebox]
Which gyms near Tompkins Square Park ($<$200m) have fitness classes before 7am?
\end{tcolorbox}

\paragraph{CORE-Bench Hard}
Tasks ask an agent to computationally reproduce specific quantitative results from a published scientific paper, given a code repository, dataset, and research paper.

\begin{tcolorbox}[examplebox]
Run the main.py file three times. First, with \texttt{config/uci.json}, the preprocessing task, and the CTGCN-C method. Second, with \texttt{config/uci.json}, the embedding task, and the CTGCN-C method. Third, using Python3 with \texttt{config/uci.json} and the link-pred task.
\end{tcolorbox}

\paragraph{GAIA}
GAIA consists of tasks requiring multi-step tool use --- including web browsing, code execution, file reading (PDFs, spreadsheets, audio), and multimodal understanding.

\begin{tcolorbox}[examplebox]
(Given an Excel file) The attached Excel file contains the sales of menu items for a local fast-food chain. What were the total sales that the chain made from food (not including drinks)? Express your answer in USD with two decimal places.
\end{tcolorbox}

\paragraph{OnlineMind2Web} 
Online Mind2Web is the live, online version of Mind2Web. It does not rely on cached pages and allows for real-time testing against dynamic, evolving web interfaces. All tasks are sourced from 136 popular websites to reflect authentic user workflows.

\begin{tcolorbox}[examplebox]
Browse used Audi cars made before 2015 and sort by lowest price on KBB. Website: \texttt{https://www.kbb.com/}
\end{tcolorbox}

\paragraph{SciCode}
Tasks consist of research-level coding problems decomposed into subproblems drawn from actual scientific work across physics, chemistry, biology, math, and materials science.

\begin{tcolorbox}[examplebox]
Main problem: Reproduce the Chern number phase diagram of the Haldane model on a hexagonal lattice. Subproblem 1.1: Write a Haldane model Hamiltonian on a hexagonal lattice, given: wavevector components $k_x$ and $k_y$, lattice spacing $a$, nearest-neighbor coupling constant $t_1$, next-nearest-neighbor coupling constant $t_2$, phase $\varphi$ for next-nearest-neighbor hopping, and on-site energy $m$. Output: a $2\times2$ matrix. Subproblem 1.2: Calculate the Chern number using the Haldane Hamiltonian, given the grid size $\delta$ for discretizing the Brillouin zone\ldots
\end{tcolorbox}

\paragraph{ScienceAgentBench} ScienceAgentBench is a benchmark for evaluating the ability of language agents to conduct data-driven scientific discovery. % \\

\begin{tcolorbox}[examplebox]
	(Given separate training and test datasets) Train a multitask model on the Clintox dataset to predict a drug's toxicity and FDA approval status. Save the test set predictions, including the SMILES representation of drugs and the probability of positive labels, to ``\texttt{pred\_results/clintox\_test\_pred.csv}''.
\end{tcolorbox}

\paragraph{SWE-bench Verified Mini}
A subset of SWE-bench Verified where each task gives the agent a real GitHub issue description and a full Python repository, and requires the agent to produce a code patch that makes failing unit tests pass.

\begin{tcolorbox}[examplebox]
(Given a repo and issue description) Navigate the repository, identify the relevant code path in \texttt{django/db/models/query.py}, and produce a \texttt{.patch} file that makes the FAIL\_TO\_PASS unit tests pass without breaking existing PASS\_TO\_PASS tests.
\end{tcolorbox}

\paragraph{$\tau$-bench Airline}
Tasks simulate a realistic airline customer service scenario in which a human user (played by another LLM) contacts an agent with requests like rebooking a flight, adding a passenger, or canceling a reservation. The agent must follow a detailed airline policy document and a limited set of callable functions.

\begin{tcolorbox}[examplebox]
(Given the airline's policy and pre-defined functions) Your user id is \texttt{daiki\_muller\_1116}. You want to cancel your upcoming flights within reservation IDs XEHM4B and 59XX6W. If the agent says either reservation has basic economy flights, ask to upgrade them to economy first and then cancel them. You are very persistent and terse but clear. After the third agent message, also ask whether you have any other upcoming flights and what their total cost is.
\end{tcolorbox}

\paragraph{USACO}
Tasks are problems from the USA Computing Olympiad, spanning four difficulty tiers (Bronze through Platinum).

\begin{tcolorbox}[examplebox]
Farmer John has $N$ cows ($2 \leq N \leq 10^5$), each liking exactly one type of hay $h_i$. He can host focus groups over contiguous ranges of cows --- if more than half the cows in a group prefer the same type, all cows switch to that type. He wants to know which types of hay can become universally liked. Given $T$ test cases, each with $N$ and a list of $h_i$ values, output all achievable universal hay types in increasing order, or $-1$.
\end{tcolorbox}

\subsubsection{Scaffolds}\label{app:hal_scaffolds}
The scaffolds included in the HAL benchmark dataset are in Table \ref{tab:hal_agent_names}. For each benchmark, the scaffold consist of a contrast between a generalist scaffold (e.g., HAL generalist, Claude Code) and and a specialist scaffold tailored to the specific benchmark (e.g., SWE agent, $\tau$-bench tool-calling). %The imbalanced patterns of scaffold overlap result in high levels  
For most benchmarks, the majority of LLMs have scores across multiple scaffolds, with the exceptions of USACO, SWE-bench Verified Mini, and ScienceAgentBench. USACO in particular has very little model--scaffold diversity, with uncertainty visible in its very large HDI intervals when trying to generalize scaffolds.

\subsection{Harbor Index Dataset for Decision Study Corroboration}\label{app:harborindex}
Harbor-Index\footnote{\url{https://harbor-index.org/}} is a curated meta-dataset and benchmark containing 82 difficult and diverse tasks designed for evaluating AI language model agents.  Distilled from a pool of over 6,000 candidate tasks across 54 benchmarks, the final dataset features 82 high-quality tasks spanning 29 benchmarks and seven domains (including software engineering, scientific research, tool use, mathematics, data analytics, and security). It was built by passing candidate tasks through difficulty filtering, automated AI audits, human reviews, and iterative audit-and-fix loops to weed out structurally broken or flawed tasks. 

The final dataset lacks sufficient per-benchmark task representation that prevent inclusion in all the analyses of this paper. Descriptive statistics are in Table \ref{tab:harborindex}. Thus, to estimate both task and benchmark effects, we take the subset of benchmarks with at least three items. The cross-benchmark model of Equation \ref{eq:pooled} is fit and produces the sample the posterior draws used in Figure \ref{fig:pooled_snr}. For our analyses, we use their verified and judged outcome classifications as the measure of task success (i.e., if the verifier marked a reward as a false negative and the judge determined it was a ``true solve'' it was marked as successful). Because the dataset suffers from a small quantity of total tasks, both the proportion of variance due to persistent differences in LLM and overall reliability begins to drop more sharply if we take the subset of benchmarks with at least four items, decreasing further with the subset that has five items. We conjecture this pattern may be in part due to the nonrandom nature of the item selection process used to represent each benchmark. %, resulting in patterns of behavior that cluster in facets of variation that do not easily reduce with changes to evaluation design.

% We also note that Figure \ref{fig:pooled_snr} shows a hypothetical 300 items across similm

\begin{table}[!h]
\centering
\caption{Descriptive statistics from the complementary Harbor Index dataset. \textbf{Bolded} datasets have at least three tasks and were used in the analysis of benchmark and item effects on agent system signal.}
\label{tab:harborindex}
\resizebox{\ifdim\width>\linewidth\linewidth\else\width\fi}{!}{
\begin{tabular}{lrrrrr}
\toprule
Benchmark & Models & Tasks & Scaffold & Model-Scaffold Pair & Mean Score \\
\midrule
\textbf{algotune} & 9 & 5 & 4 & 18 & 0.078\\
\textbf{arcagi2} & 9 & 5 & 4 & 18 & 0.067\\
bigcodebench & 9 & 1 & 4 & 18 & 0.222\\
\textbf{bixbench} & 9 & 5 & 4 & 18 & 0.122\\
codepde & 9 & 1 & 4 & 18 & 0.000\\
\addlinespace
cybergym & 9 & 2 & 4 & 18 & 0.1 39\\
dacode & 9 & 1 & 4 & 18 & 0.111\\
\textbf{featurebench} & 9 & 4 & 4 & 18 & 0.042\\
\textbf{gaia} & 9 & 3 & 4 & 18 & 0.185\\
\textbf{gaia2} & 9 & 5 & 4 & 18 & 0.051\\
\addlinespace
gpqadiamond & 9 & 1 & 4 & 18 & 0.056\\
\textbf{gso} & 9 & 7 & 4 & 18 & 0.024\\
\textbf{hle} & 9 & 8 & 4 & 18 & 0.079\\
\textbf{labbench} & 9 & 4 & 4 & 18 & 0.182\\
omnimath & 9 & 2 & 4 & 18 & 0.278\\
\addlinespace
qcircuitbench & 9 & 1 & 4 & 18 & 0.000\\
replicationbench & 9 & 1 & 4 & 18 & 0.444\\
\textbf{scicode} & 9 & 3 & 4 & 18 & 0.019\\
skillsbench & 9 & 2 & 4 & 18 & 0.167\\
sldbench & 9 & 1 & 4 & 18 & 0.056\\
\addlinespace
spider2 & 9 & 2 & 4 & 18 & 0.000\\
\textbf{swebenchpro} & 9 & 4 & 4 & 18 & 0.139\\
\textbf{swebenchverified} & 9 & 5 & 4 & 18 & 0.067\\
swelancer & 9 & 2 & 4 & 18 & 0.139\\
swesmith & 9 & 1 & 4 & 18 & 0.111\\
\addlinespace
swtbenchverified & 9 & 1 & 4 & 18 & 0.333\\
\textbf{tb} & 9 & 3 & 4 & 18 & 0.241\\
usaco & 9 & 1 & 4 & 18 & 0.000\\
widesearch & 9 & 1 & 4 & 18 & 0.000\\
\bottomrule
\end{tabular}}
\end{table}

\subsection{External Validation Datasets for Model Rankings}\label{app:external_valid_data}
% \paragraph{External evidence for the shared model component.}
% We test whether the component treated as model signal corresponds to recurring performance outside the fitted benchmark panel. Among overlapping LLMs, we compare the posterior-median shared model effect with an unweighted mean-score baseline on four external agent benchmarks. For SWE-bench Verified and \(\tau^2\)-bench, we exclude the corresponding internal benchmark before estimating either score.

% The benchmark-specific comparisons average across the included
% SWE-bench Verified leaderboards. 

We compare the posterior-median model effect \(\widehat{\theta}_m\) with four contemporaneous external benchmarks containing at least six overlapping LLMs and which report individual task-level performance by model. The main body reports concordance of the observed rankings with \(\widehat{\theta}_m\) and mean score with these external benchmarks in Table \ref{tab:external-rank-validity}; \S~\ref{app:benchmark_rank_agreement} analyzes the differences between the two HAL ranking mechanisms.
To prevent data leakage for the external SWE-bench and \(\tau^2\)-bench, the semantically corresponding HAL benchmark is excluded before estimating both \(\widehat{\theta}_m\) and the raw-score baseline (see \S~\ref{app:loo-benchmark}), as described in the details of the data and collection below.
% BFCL v4 , Terminal-Bench 2.0 \citep{merrill_terminal-bench_2026}, SWE-bench Verified, and the core domains of \(\tau^2\)-bench \citep{barres_tau2-bench_2025}. 

\begin{itemize}
    \item \textbf{The Berkeley Function Calling Leaderboard} (BFCL) V4 \citep{patil_berkeley_2025} is a benchmark designed to evaluate the tool use, API execution, and agentic capabilities of large language models.\footnote{\url{https://gorilla.cs.berkeley.edu/leaderboard.html}}. We use BFCL's holistic score for function-calling models.

    \item \textbf{Terminal-Bench 2.0} \citep{merrill_terminal-bench_2026} is an evaluation suite and benchmark designed to test how well AI agents perform complex, multi-step tasks inside a real command-line interface (CLI).\footnote{\url{https://hub.harborframework.com/datasets/terminal-bench/terminal-bench-2/latest?tab=leaderboard&leaderboard=2-0}}. To compare LLMs, we average each model's score across reported agent runs.

    \item \textbf{SWE-bench Verified} \citep{jimenezSWEbenchCanLanguage2024a} is the superset of tasks that are associated with the HAL dataset SWE-bench Verified Mini. The data of several third-party external leaderboard providers\footnote{\url{https://www.swebench.com/}, \url{https://www.vals.ai/benchmarks/swebench}, \url{https://llm-stats.com/benchmarks/swe-bench-verified}} were collected. Accuracy scores were averaged across leaderboards for overlapping models for the standard Kendall's $\tau$, and where left separate for the multilevel partial correlation (Appendix \ref{app:multilevelpartial}). When performing analyses with this data, such as in Table \ref{tab:external-rank-validity}, the external correlations were calculated with estimates from the HAL dataset after removing SWE-bench Verified Mini from the analysis. The description of this process based on the pooled model (Eq. \ref{eq:pooled}) can be found in Appendix \ref{app:loo-benchmark}. 

    \item \(\mathbf{\tau^2}\)\textbf{-bench Core} \citep{barres_tau2-bench_2025} is an evaluation framework and benchmark for conversational AI and customer service agents that tests performance in a ``dual-control'' environment where both the AI agent and the user take active actions. It contains tasks that evaluate LLM agents on retail, airline (of which HAL's $\tau$-bench Airline is a subset), and telecom customer-service tasks, where the agent and the user both act on the world.\footnote{\url{https://taubench.com/leaderboard?benchmark=core}} We use the overall score across all of these domains and, as with SWE-bench Verified, we only calculate correlations with HAL data after removing $\tau$-bench Airline. The description of this process based on the pooled model (Eq. \ref{eq:pooled}) can be found in Appendix \ref{app:loo-benchmark}.
\end{itemize}

\subsubsection{Differences in concordance with external performance}
\label{app:benchmark_rank_agreement}
In addition to Table \ref{tab:external-rank-validity}, we provide additional analyses to explore the differences between transportable ranking methods. This section supplements the differences in correlations. We compared the rank agreement of the more traditional aggregated mean score $\bar x$ and the latent estimated value $\hat\theta$ with external performance separately for each benchmark. For each model, the external reference was its mean reported benchmark score across the available sources and evaluation conditions. Both scoring mechanisms were evaluated on the same models within each benchmark. We used Kendall's $\tau_b$, defining the paired difference as
\[
\Delta_b
=
\tau_b(\bar x,Z_b)
-
\tau_b(\hat\theta,Z_b),
\]
where $Z_b$ denotes the aggregated external reference for benchmark $b$. Thus, $\Delta_b<0$ favors $\hat\theta$.

To characterize uncertainty, we constructed nominal 95\% percentile intervals using a paired bootstrap within each benchmark (with 20,000 bootstraps). Each resampled unit was a complete model-level observation containing mean score, $\hat\theta$, and the external reference, preserving the dependence between the two estimated correlations. Given the small numbers of models, particularly for benchmarks with six or seven observations, these intervals were interpreted as exploratory rather than as reliably calibrated confidence intervals. We also recomputed
each difference after removing each model in turn to assess sensitivity to individual observations. The resulting leave-one-out ranges are influence diagnostics, not confidence intervals.

\begin{table}[htbp]
\centering
\caption{Within-benchmark agreement with the aggregated external
reference. Negative differences favor $\hat\theta$. Bootstrap
intervals are exploratory, marginal 95\% percentile intervals and
are not adjusted for multiple comparisons.}
\label{tab:benchmark_rank_agreement}
\footnotesize
\begin{tabular}{lrrrrll}
\hline
Benchmark & $n$
& $\tau_b(\bar x,Z)$
& $\tau_b(\hat\theta,Z)$
& $\Delta_b$
& Bootstrap interval
& LOO range \\
\hline
BFCL
& 7  & 0.429 & 0.714 & $-0.286$
& $[-1.111,\phantom{-}0.556]$
& $[-0.533,\phantom{-}0.000]$ \\
SWE-bench
& 20 & 0.533 & 0.755 & $-0.222$
& $[-0.503,\phantom{-}0.045]$
& $[-0.258,-0.129]$ \\
$\tau^2$-core
& 6  & 0.333 & 0.867 & $-0.533$
& $[-1.231,\phantom{-}0.000]$
& $[-0.800,-0.400]$ \\
Terminal-Bench 2.0
& 7  & 0.524 & 0.810 & $-0.286$
& $[-0.824,\phantom{-}0.000]$
& $[-0.400,-0.133]$ \\
\hline
\end{tabular}
\end{table}

The observed correlations consistently favored $\hat\theta$ (Table~\ref{tab:benchmark_rank_agreement}). Its advantage in
$\tau_b$ ranged from 0.222 on SWE-bench to 0.533 on $\tau^2$-core. Both mechanisms were positively associated with the external
reference in every benchmark, but $\hat\theta$ exhibited stronger agreement throughout.

The direction of the comparison was also stable under single-model deletion. Every leave-one-out difference remained
strictly negative for SWE-bench, $\tau^2$-core, and Terminal-Bench~2.0. For BFCL, deleting an individual model could reduce the difference to zero, but no deletion reversed its sign. Consequently, the observed direction was not dependent on retaining
any single model, although this diagnostic does not establish stability to broader changes in the model sample.

As expected, uncertainty remained substantial due to the constraint of having few candidate contemporaneous benchmarks that correspond with HAL benchmark models. Thus predictably for small sample sizes, none of the bootstrap intervals excluded zero: those for BFCL and SWE-bench extended above zero, whereas those for $\tau^2$-core and Terminal-Bench~2.0 ended at zero. The latter endpoints should not be interpreted as precise significance thresholds, given the discrete rank statistic and small samples. Interval bounds below $-1$ are permissible because a difference between two Kendall correlations lies in $[-2,2]$. Undefined bootstrap differences were absent except for $\tau^2$-core, where they occurred in 0.01\% of resamples; its interval was calculated from the defined replicates.

These difference analyses provide consistent descriptive evidence that $\hat\theta$ ranks the observed models more closely to their
aggregated external performance than the mean score does. They do not, however, establish a statistically conclusive advantage within
individual benchmarks. Moreover, the analysis treats each model's aggregated external score as its reference value: it does not
separately propagate uncertainty in that score or adjust for unequal coverage of sources and evaluation conditions. The results therefore characterize agreement with the observed mean references, rather than with context-adjusted latent performance.

\subsubsection{Combined External Data Multilevel Partial Correlation}\label{app:multilevelpartial}
In addition to traditional Kendall's $\tau_b$ correlations, a multilevel partial correlation was calculated to clarify the relationships with greater statistical power. For LLM \(m\) evaluated on benchmark \(b\) and external leaderboard providers \(\ell\), let \(R_{mb\ell}\) denote its observed performance rank and \(P_{mb\ell}\) the rank induced by the calculated proxy. Benchmark- and leaderboard-provider-level heterogeneity can be modeled through the rank-based mixed-effects specifications
\[
R_{mb\ell}=\mathbf{x}_{mb\ell}^{\top}\boldsymbol{\beta}_{R}+u^{R}_{b}+v^{R}_{\ell}+\varepsilon^{R}_{mb\ell},
\qquad
P_{mb\ell}=\mathbf{x}_{mb\ell}^{\top}\boldsymbol{\beta}_{P}+u^{P}_{b}+v^{P}_{\ell}+\varepsilon^{P}_{mb\ell},
\]
where \(u_b^{(\cdot)}\) and \(v_\ell^{(\cdot)}\) are random effects for benchmarks and leaderboards, respectively, and \(\mathbf{x}_{mb\ell}\) contains any control variables. The inclusion of benchmark and leaderboard random effects removes the need to average across the SWE-bench Verified leaderboards, increasing the overall number of observations to 52 for this analysis. The multilevel partial Kendall correlation is then defined as the concordance between the adjusted residual ranks,
\[
\tau_{\mathrm{partial}}
=
\mathbb{E}\!\left[
\operatorname{sgn}\!\left(\tilde R_a-\tilde R_{a'}\right)
\operatorname{sgn}\!\left(\tilde P_a-\tilde P_{a'}\right)
\right],
\]
where \(\tilde R_{mb\ell}=R_{mb\ell}-\mathbf{x}_{mb\ell}^{\top}\hat{\boldsymbol{\beta}}_R-\hat u_b^R-\hat v_\ell^R\) and \(\tilde P_{mb\ell}=P_{mb\ell}-\mathbf{x}_{mb\ell}^{\top}\hat{\boldsymbol{\beta}}_P-\hat u_b^P-\hat v_\ell^P\). Thus, \(\tau_{\mathrm{partial}}\) measures agreement between the LLM rankings and the proxy rankings after accounting for observed covariates and clustering attributable to benchmarks and leaderboards. The results are in Table \ref{tab:external-rank-multilevel}.

\begin{table}[h!]
\centering
\caption{\textbf{Hierarchical concordance with external agent benchmarks.}
Kendall's \(\tau\) for the unweighted HAL mean and reliability-adjusted latent ($\widehat{\theta}$) model effect (median posterior) with \([95\%]\) CIs.}
\label{tab:external-rank-multilevel}
\resizebox{\linewidth}{!}{
\begin{tabular}{lrrrc}
\toprule
External Benchmark
& Mean score
& \(\widehat{\theta}\)
& Difference
& LLM Observations \\
\midrule
% BFCL v4
% & \(0.429\,[-0.282,\,0.836]\)
% & \(\mathbf{0.714}\,[0.147,\,0.928]\)
% & \(+0.285\) & 7 \\
% Terminal-Bench 2.0
% & \(0.524\,[-0.165,\,0.869]\)
% & \(\mathbf{0.810}\,[0.361,\,0.954]\)
% & \(+0.286\) & 7 \\
% SWE-bench Verified
% & \(0.565\,[0.306,\,0.746]\)
% & \(\mathbf{0.765}\,[0.595,\,0.870]\)
% & \(+0.200\) & 20 \\
% \(\tau^2\)-bench Core
% & \(0.333\,[-0.515,\,0.852]\)
% & \(\mathbf{0.867}\,[0.383,\,0.977]\)
% & \(+0.534\) & 6 \\
% \midrule
% Mean $\tau$
% & \(0.463\)
% & \(\mathbf{0.789}\)
% & \(+0.326\) & --- \\
Multilevel $\tau_{\text{partial}}$
& \(0.430\,[ 0.266,\, 0.570]\) 
& \(\mathbf{0.632}\,[0.507,\,0.732]\)   
& \(+0.202\) & 52 \\
\bottomrule
\end{tabular}}
% \caption*{\footnotesize Multilevel $\tau_{\text{partial}}$ retains those leaderboard-specific observations and
% includes benchmark and leaderboard random effects, yielding
% \(52\) model--source observations. Full aggregation and
% specifications are in \S~\ref{app:external_valid_data}. }
\end{table}

% To prevent data leakage for the external SWE-bench and \(\tau^2\)-bench, the semantically corresponding HAL benchmark is excluded before estimating both \(\widehat{\theta}_m\) and the raw-score baseline (see \S~\ref{app:loo-benchmark}). 

% BFCL is represented by its holistic function-calling score, and Terminal-Bench by each model's mean across reported agent runs. For SWE-bench, we average the accuracies reported by \texttt{swebench.com}, Vals AI, and \texttt{llm-stats.com}. To make the two closest comparisons strictly out-of-benchmark, we re-estimate both \(\widehat{\theta}\) and the overall-score baseline after removing SWE-bench Verified Mini for the external SWE-bench comparison and \(\tau\)-bench Airline for the external \(\tau^2\)-bench comparison.

\section{HAL Coverage Matrix}
\label{app:coverage}

\begin{sidewaysfigure}
    \centering
    \includegraphics[width=\textheight, height=\linewidth, keepaspectratio]{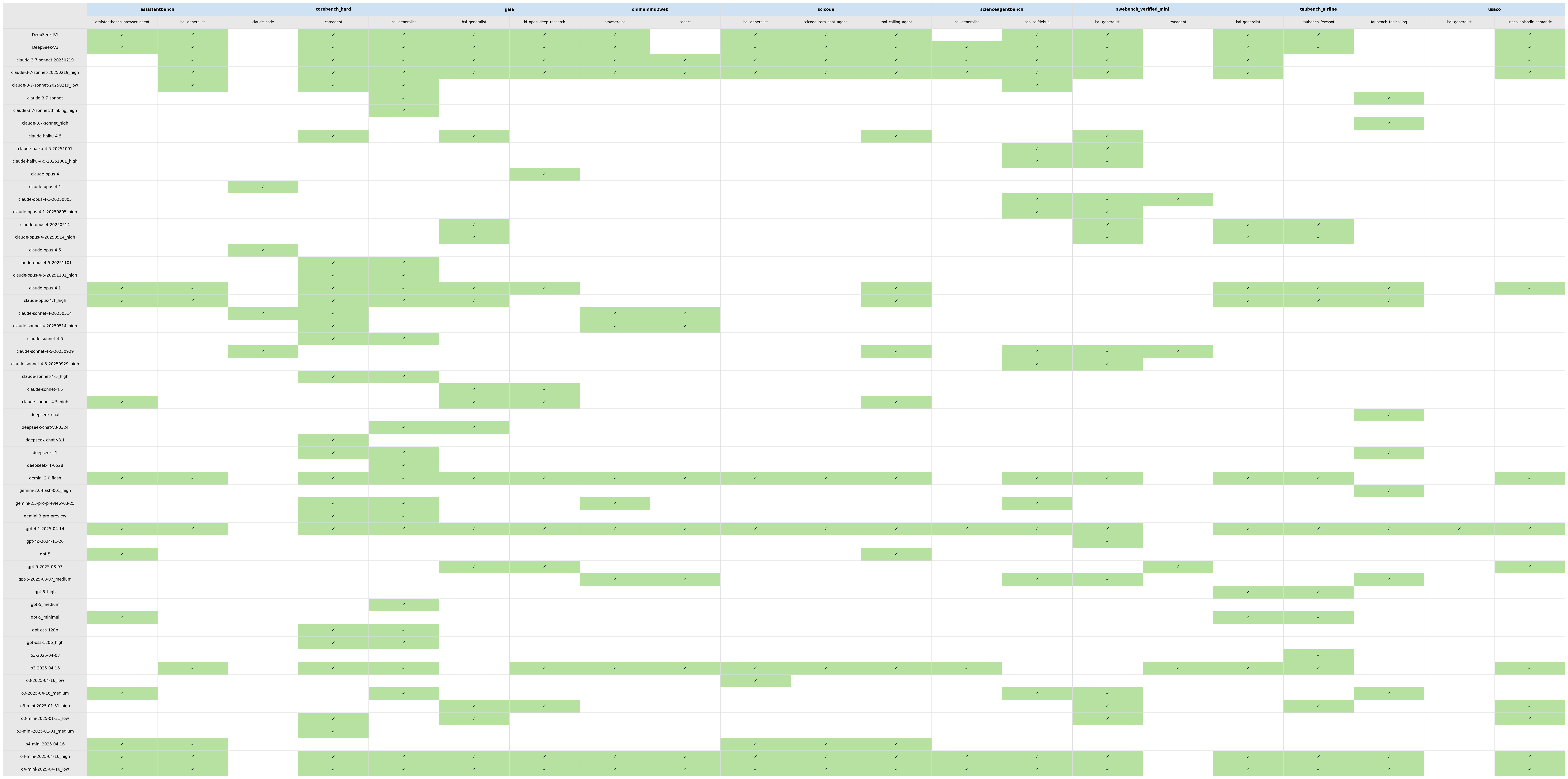}
    \caption{Matrix indicating all the variations of benchmarks, scaffold and models (with reasoning-effort) that were included in our dataset.}
    \label{fig:coverage_matrix}
\end{sidewaysfigure}

Below is the list of agents used to test each of the benchmarks. Each model was not uniformly tested with each benchmark and agent. As shown in Table~\ref{tab:hal_agent_names}, about nearly all (11/13) agents have only been tested on a single benchmark. Similarly, many models (12/54) are tested with a single agent. The full breakdown of which models, agents and benchmarks have been tested can be seen in Figure \ref{fig:coverage_matrix}.

\begin{table}[ht]
\centering
\caption{Agent harness/scaffold by benchmark.}\label{tab:hal_agent_names}
\begin{tabular}{ll}
\toprule
\textbf{Benchmark} & \textbf{Agent Names} \\
\midrule
AssistantBench & hal\_generalist, browser-use \\% assistantbench\_browser\_agent \\
CORE-Bench Hard & hal\_generalist, coreagent, claude\_code \\
GAIA  & hal\_generalist, hf\_open\_deep\_research \\
OnlineMind2Web & browser-use, seeact \\
SciCode & hal\_generalist, tool\_calling\_agent, scicode\_zero \\
ScienceAgentBench & hal\_generalist, sab\_selfdebug \\
SWE-bench Verified Mini & hal\_generalist, sweagent \\
$\tau$-bench Airline & hal\_generalist, taubench\_tool\_calling, taubench\_fewshot \\
usaco & hal\_generalist, usaco\_episodic\_semantic \\
\bottomrule
\end{tabular}

\end{table}

\begin{figure}[ht]
    \centering
    \includegraphics[width=0.7\linewidth]{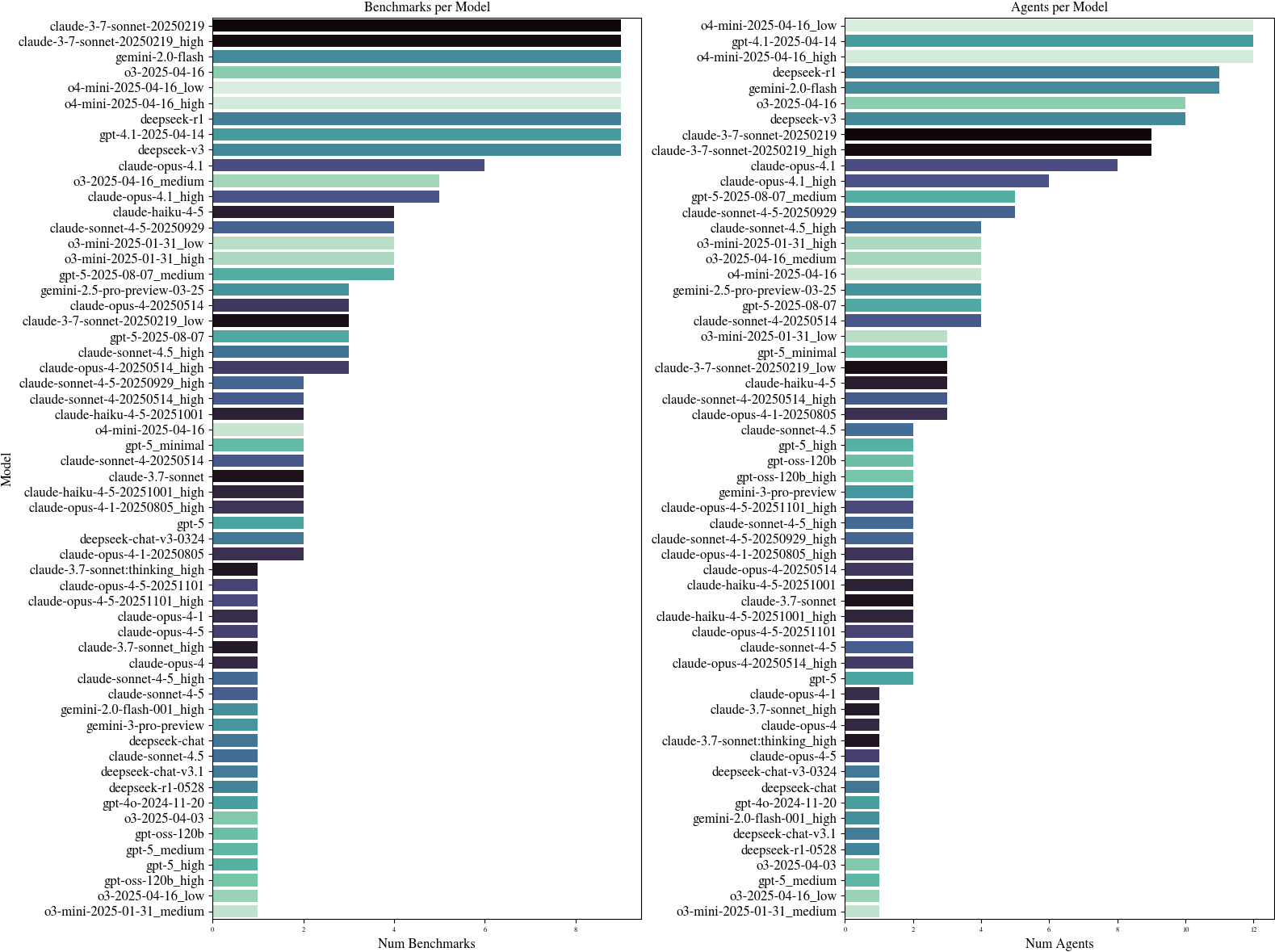}
    \caption{Number of benchmarks tested (left) and agents tested (right) per model.}
    \label{fig:model_test_distribution}
\end{figure}

% Todo: make vertical and full page
\newpage

% ============================================================
\section{Methods, Estimation, and Computational Details}\label{app:methods-details}
% ============================================================

% \subsection{Model Specifications in Detail}\label{sec:app-specs}
This appendix reports prior distributions, parameterization, sampler settings, convergence diagnostics, posterior predictive checks, and the treatment of repeated observations. It also gives reliability expressions for the observed
unbalanced allocation. For each posterior draw, design-specific error is computed using the realized cell weights rather than the equal-allocation approximations used for the main D-studies.

We provide the full random-effects structure for each model using standard mixed-model notation. 
For the main body studies, we estimate all variance components jointly using Bayesian generalized linear mixed models, which provide partial pooling for sparsely observed cells and propagate uncertainty into the resulting generalizability coefficients.

%% OLD Naive decomposition baselines REMOVED
%All models assume Bernoulli outcomes with logit link for Methods 2--4, and Gaussian identity link for Decomposition~1. We denote indices as: $i$ for task (``item'' in benchmarking literature), $j$ for model (model name $\times$ reasoning-effort), $k$ for agent scaffold, and $b$ for benchmark.

% \subsubsection{Decomposition 1: Na\"ive}

% \begin{equation}
% \text{logit}\,\Pr(Y_{ij}=1) = \mu + t_i + m_j, \quad t_i \stackrel{\text{iid}}{\sim} \mathcal{N}(0,\sigma^2_t),\; m_j \stackrel{\text{iid}}{\sim} \mathcal{N}(0,\sigma^2_m).
% \end{equation}
% Fit per benchmark. The residual on the logit scale is the standard logistic variance $\pi^2/3 \approx 3.29$.

% \medskip\noindent\texttt{brms} specification:
% \begin{lstlisting}[language=R]
% brm(score ~ 1 + (1|task_id) + (1|model),
%     family = bernoulli("logit"), ...)
% \end{lstlisting}

% \subsubsection{Decomposition 2: Model--Scaffold Pair}

% \begin{equation}
% \text{logit}\,\Pr(Y_{i,jk}=1) = \mu + t_i + p_{jk}, \quad p_{jk} \stackrel{\text{iid}}{\sim} \mathcal{N}(0,\sigma^2_p).
% \end{equation}
% Fit per benchmark. The model$\times$agent interaction is absorbed into the pair term $p_{jk}$.

% \medskip\noindent\texttt{brms} specification:
% \begin{lstlisting}[language=R]
% brm(score ~ 1 + (1|task_id) + (1|model:agent_name),
%     family = bernoulli("logit"), ...)
% \end{lstlisting}

\subsection{Estimation Details}\label{sec:estimation}

\subsubsection{Leaderboard-level pooled model}\label{sec:bayes}

Let \(y_{bima}\in\{0,1\}\) denote success on benchmark \(b\), task \(i\), model \(m\), and scaffold \(a\). We assume
\[
y_{bima}\mid p_{bima}\sim\operatorname{Bernoulli}(p_{bima}),
\qquad
\operatorname{logit}(p_{bima})=\eta_{bima},
\]
with latent linear predictor (from Equation \ref{eq:pooled})
\begin{align*}
\eta_{bima}
={}&\mu
+u_b+u_{bi}+u_m+u_a
+u_{bim}+u_{bia}+u_{ma} \nonumber\\
&+u_{bm}+u_{ba}+u_{bma}.
% \label{eq:leaderboard-model}
\end{align*}
Here \(bi\) identifies a task nested within its benchmark. Each term is a mean-zero random intercept,
\[
u_{F,j}\mid \sigma_F \sim \mathcal N(0,\sigma_F^2),
\qquad
F\in\{b,bi,m,a,bim,bia,ma,bm,ba,bma\},
\]
and random-effect families are conditionally independent. This decomposition separates persistent model and scaffold differences from variation attributable to benchmark choice, task composition, and model--scaffold compatibility. In particular, \(u_{bm}\) captures benchmark-dependent model performance, while \(u_{bma}\) captures benchmark-specific compatibility between models and scaffolds; both can alter rankings even when average model effects are unchanged.

The model was fitted to \(29{,}923\) observations spanning \(9\) benchmarks, \(1{,}117\) benchmark--task units, \(54\) models, and \(13\) scaffolds. Because the likelihood is Bernoulli-logit, all variance components are defined on the latent log-odds scale. When an observation-level residual is required for a generalizability coefficient, the conventional logistic variance \(\pi^2/3\) is used. Reliability quantities are computed separately for every posterior draw, rather than from ratios of posterior mean variances, thereby preserving uncertainty in nonlinear variance decompositions.

\medskip\noindent\texttt{brms} specification:
\begin{lstlisting}[language=R]
brm(score ~ 1 + (1|benchmark) + (1|benchmark:task_id)
    + (1|model_name) + (1|agent_name)
    + (1|benchmark:task_id:model_name)
    + (1|benchmark:task_id:agent_name)
    + (1|model_name:agent_name)
    + (1|benchmark:model_name)
    + (1|benchmark:agent_name)
    + (1|benchmark:model_name:agent_name),
    family = bernoulli("logit"), ...)
\end{lstlisting}

\subsubsection{Benchmark-level decomposition models}

To quantify the signal supplied by each benchmark, we also fit a separate model within every benchmark using Equation \ref{eq:perbenchmark}:
\begin{align*}
y_{ima}&\sim\operatorname{Bernoulli}(p_{ima}),\\
\operatorname{logit}(p_{ima})
&=\mu+u_i+u_m+u_a+u_{im}+u_{ia}+u_{ma}.
% \label{eq:within-benchmark-model}
\end{align*}
These fits distinguish stable model variation from task-, scaffold-, and interaction-driven variation within an instrument. Thus, a benchmark with many observations need not be highly informative for ranking models: its effective signal depends on the posterior magnitude of model-related variance relative to the variance induced by tasks, scaffolds, and their interactions.

\medskip\noindent\texttt{brms} specification:
\begin{lstlisting}[language=R]
brm(score ~ 1 + (1|task_id) + (1|model_name) + (1|agent_name)
    + (1|task_id:model_name) + (1|task_id:agent_name)
    + (1|model_name:agent_name),
    family = bernoulli("logit"), ...)
\end{lstlisting}

\subsubsection{Priors and posterior computation}\label{sec:app-bayes}

We used the default weakly informative \texttt{brms} priors. For every group-level standard deviation,
\[
\sigma_F\sim \operatorname{Student}\text{-}t^{+}(3,0,2.5),
\]
where \(t^{+}\) denotes truncation to \(\sigma_F\geq 0\). Equivalently, the estimated variance component is \(\sigma_F^2\). The prior is concentrated near modest latent-scale heterogeneity while retaining sufficiently heavy tails for large benchmark or task effects. The intercept used the corresponding weakly informative \(\operatorname{Student}\text{-}t(3,0,2.5)\) prior. These priors regularize components supported by few levels---notably benchmark and scaffold effects---without forcing them toward equality.

Posterior sampling used Stan's Hamiltonian Monte Carlo implementation through \texttt{brms} \citep{burkner_bayesian_2021}, with the conservative target average proposal acceptance probability during the adaptation period for the sampler $\texttt{adapt\_delta} = 0.95$ without any resultant divergent transitions. The full model used six chains of \(9{,}000\) iterations, including \(4{,}000\) warm-up iterations, with thinning by eight, yielding \(3{,}750\) retained draws. Each benchmark-specific model used four chains of \(2{,}000\) iterations, including \(1{,}000\) warm-up iterations, with thinning by three. On an Apple M1 Max, the full fit required approximately \(5.5\) hours, while a benchmark-specific fit required approximately \(6.5\) minutes on average.

For the full model, all reported split-\(\widehat R\) values rounded to \(1.00\); bulk effective sample sizes for variance components ranged from \(1{,}559\) to \(3{,}743\), and tail effective sample sizes ranged from \(2{,}062\) to \(3{,}783\). Benchmark-specific fits were also inspected individually. For example, the largest split-\(\widehat R\) in the USACO fit was \(1.03\), with uncertainty retained in all downstream posterior summaries rather than suppressed through plug-in estimates.

% \subsubsection{Method C: Bayesian Generalized Linear Mixed Effects}\label{sec:bayes}

% We fit the same logistic specification as Method~B within a fully Bayesian framework using Hamiltonian Monte Carlo (HMC) via \texttt{brms}/Stan \citep{burkner_bayesian_2021}. Weakly informative half-Student-$t$ priors are placed on all standard-deviation parameters. Four chains of 2{,}000 (Models~1--3) or 6 chains, each with iter = 9000; warmup = 4000; thin = 8; total post-warmup draws = 3750  chains of 9{,}000 (Model~4) iterations are run with thinning factors of 2 or 8, respectively, and the conservative target average proposal acceptance probability during the adaptation period for the Hamiltonian Monte Carlo sampler $\texttt{adapt\_delta} = 0.95$ without any resultant divergent transitions. Convergence is assessed by $\hat{R} < 1.01$ for all parameters. Posterior means, medians, and highest-density intervals (HDIs) of variance proportions are reported.

\paragraph{Convergence.} All parameters across all models fit and all estimates derived from posterior draws throughout this study achieve $\hat{R} < 1.05$, the standard threshold for adequate mixing. For the 95\% of estimates \(\widehat R\) values rounded to \(1.00\). For the leaderboard-level model, complete convergence information is in Table \ref{tab:full-model-posterior}

% Required packages:
% \usepackage{booktabs}

\begin{table*}[t]
\centering
\caption{\textbf{Full posterior summary.} Bayesian fit estimates for the full leaderboard-level model (Eq. \ref{eq:pooled}). Group-level
entries are standard deviations of random intercepts on the latent log-odds
scale. Intervals are equal-tailed 95\% posterior credible intervals.}
\label{tab:full-model-posterior}
\footnotesize
\setlength{\tabcolsep}{2.25pt}
\begin{tabular}{@{}lrrrrrrr@{}}
\toprule
Effect & Levels & Estimate & SD & 95\% CrI
       & $\widehat{R}$ & Bulk ESS & Tail ESS \\
\midrule
\multicolumn{8}{@{}l}{\textit{Group-level standard deviations}} \\[2pt]
Scaffold, $\sigma_{a}$                             
    & 13     & 0.79 & 0.44 & $[0.06,\ 1.74]$ & 1.00 & 3454 & 3360 \\
Benchmark, $\sigma_{b}$                            
    & 9      & 2.45 & 0.73 & $[1.39,\ 4.21]$ & 1.00 & 3743 & 3783 \\
Benchmark $\times$ scaffold, $\sigma_{ba}$         
    & 21     & 0.69 & 0.36 & $[0.06,\ 1.43]$ & 1.00 & 3580 & 3615 \\
Benchmark $\times$ model, $\sigma_{bm}$            
    & 182    & 0.52 & 0.17 & $[0.12,\ 0.80]$ & 1.00 & 1771 & 2110 \\
Benchmark $\times$ model $\times$ scaffold, $\sigma_{bma}$
    & 287    & 0.55 & 0.14 & $[0.24,\ 0.79]$ & 1.00 & 1559 & 2062 \\
Task within benchmark, $\sigma_{bt}$               
    & 1117   & 2.41 & 0.09 & $[2.25,\ 2.58]$ & 1.00 & 3309 & 3350 \\
Task $\times$ scaffold in benchmark, $\sigma_{bta}$
    & 2394   & 1.03 & 0.05 & $[0.93,\ 1.14]$ & 1.00 & 3445 & 3540 \\
Task $\times$ model in benchmark, $\sigma_{btm}$
    & 18856  & 0.88 & 0.07 & $[0.75,\ 1.01]$ & 1.00 & 2513 & 3241 \\
Model, $\sigma_{m}$                                
    & 54     & 0.78 & 0.14 & $[0.52,\ 1.07]$ & 1.00 & 3152 & 3550 \\
Model $\times$ scaffold, $\sigma_{ma}$             
    & 209    & 0.36 & 0.14 & $[0.05,\ 0.61]$ & 1.00 & 1852 & 2376 \\
\addlinespace
\multicolumn{8}{@{}l}{\textit{Population-level coefficient}} \\[2pt]
Intercept, $\mu$                                   
    & ---    & $-1.95$ & 0.88 & $[-3.62,-0.17]$ & 1.00 & 3693 & 3574 \\
\bottomrule
\end{tabular}

\vspace{2pt}
\begin{minipage}{0.98\textwidth}
\footnotesize
\textit{Note.} ``Levels'' denotes the number of observed levels of each
grouping factor. Estimate and SD are the posterior mean and posterior
standard deviation, respectively. For group-level effects, the estimate
summarizes the random-intercept standard deviation $\sigma$; for the
intercept, it summarizes the coefficient itself. ESS denotes effective sample
size. ``Scaffold'' corresponds to \texttt{agent\_name} in the estimation data.
\end{minipage}
\end{table*}

\subsubsection{Leave-one-benchmark-out stability}

To assess whether the leaderboard-level decomposition was dominated by any single instrument, we refit Equation~\ref{eq:pooled} after removing each benchmark in turn. These are full Bayesian refits, not importance-sampling approximations. Each refit used four chains of \(4{,}000\) iterations, including \(2{,}000\) warm-up iterations, with thinning by five, yielding \(1{,}600\) retained draws. Stability was evaluated by comparing the posterior distributions of variance components and derived reliability quantities with those from the complete-data fit. This analysis directly tests whether conclusions about model, scaffold, and benchmark contributions persist under changes to the benchmark ecosystem.

\subsection{Bayesian Estimation for Linear Mixed Effect Models}\label{app:bayes_lme}
To establish methodological sensitivity, Equation \ref{eq:pooled} was estimated using a Bayesian linear mixed effect model on the observation scale. Additional detail about the differences in methods and results can be found in Appendix \ref{app:method_comps}. All linear estimations of this use the same hyperparameters as above, except that the family is Gaussian with an identity link.

\subsection{Frequentist Linear Mixed Effects (LME)}\label{sec:lme}

We fit a Gaussian identity-link model via restricted maximum likelihood (REML). Although the binary outcome violates normality, LME provides a familiar baseline and is the most commonly used variance-component estimator in G-theory applications. Variance components are extracted directly from the REML fit.

\subsection{Frequentist Generalized Linear Mixed Effects (GLME)}\label{sec:glme}

A logistic mixed-effects model with a Bernoulli likelihood and logit link is fit via Laplace approximation to the marginal likelihood, with parameter estimation by penalized iteratively reweighted least squares (PIRLS). This respects the binary nature of the data but estimates are on the logit scale; we convert variance proportions by computing the share of total variance (including the logistic residual variance $\pi^2/3$) attributable to each component. Both the Linear and Generalized Linear models were estimated using \texttt{lme4} \citep{bates_fitting_2015}.

\subsection{Nonparametric Distance Components (DISCO)}\label{sec:disco}

We estimate a nonparametric variance decomposition (see Appendix \ref{sec:app-disco}) using distance components \citep{rizzo_disco_2010,szekely_energy_2017} from the energy-statistics literature. DISCO decomposes total dispersion---measured by pairwise Euclidean distances---into between- and within-group components without distributional assumptions. For a single facet with $K$ groups,
\begin{equation}\label{eq:disco}
    \mathcal{S}_{\text{total}} \;=\; \mathcal{S}_{\text{between}} + \mathcal{S}_{\text{within}},
\end{equation}
where $\mathcal{S}$ denotes the energy-based dispersion statistic.  The G-coefficient analog is $E\rho^2_{\textsc{disco}} = \mathcal{S}_p / (\mathcal{S}_p + \mathcal{S}_{\text{within}})$.  DISCO captures nonlinear relationships and is robust to the heavy skewness observed in the Bayesian posteriors, particularly for facets with few levels (e.g., agents). However, it does not guarantee a positive $\sigma^2_\epsilon/\mathcal{S}_{\text{global within}}$ term, so its reported ``variance''/dispersion shares sum to one across the total explained dispersion.

% \subsection{Estimation Methods}

% Each of the four models is estimated by four methods, chosen to span the space from conventional to robust. In the main body, we report Method C. Findings are robust to estimation method.

\subsection{Details On Compute-Usage}\label{app:compute}
Estimations of the main variance decomposition models used in the body of the paper took 8 total hours on an Apple M1 Max. LOO ablation studies in Appendix \ref{app:robustness} took 57 hours. Methodological contrasts reported in Appendix \ref{app:method_comps} took 2 hours for frequentist estimations (both linear mixed effect models and generalized mixed effect models), Bayesian linear mixed effect models took 8 hours, and nonparametric distance components estimates took 18 hours.

\subsection{Posterior Estimands}
Table \ref{tab:reliability-coefficients-full} contains the set of estimands found in the paper.

\begin{table}[t]
\centering
% \small
% \footnotesize
\renewcommand{\arraystretch}{0.2}
\caption{Reliability estimands. Each row applies
\(E\rho_o^2=\sigma_o^2/(\sigma_o^2+\sigma_\delta^2)\), \(\text{SNR}_o=\sigma_o^2/\sigma_\delta^2\), but changes the object whose ordering should generalize. The \(n_f\) terms denote equal allocation over facet \(f\).}
\label{tab:reliability-coefficients-full}
\resizebox{\ifdim\width>\linewidth\linewidth\else\width\fi}{!}{
\begin{tabular}{llccc}
\toprule
\textbf{Estimand} 
% & \textbf{Level}
& \textbf{Object of inference} & $\boldsymbol{\sigma_o^2}$ & $\boldsymbol{\sigma_\delta^2(\mathcal{D})}$ & Eq. \\
\midrule
\makecell[l]{$E\rho^2_{M(b)}$, \\$\text{SNR}_{M(b)}$} 
% & Benchmark 
& \makecell[l]{Does benchmark \(b\)\\ preserve model order?}
& $\sigma_M^2$
& 
    $\dfrac{\sigma_{IM}^2}{n_i}+\dfrac{\sigma_{MA}^2}{n_a}+\dfrac{\sigma_{IMA,e}^2}{n_i n_a}$
& (\ref{eq:perbench-model-reliability}) \\
% & \tableeqnum\label{eq:perbench-model-reliability} \\
% \addlinespace
$E\rho^2_{MA(b)}$ 
% & Benchmark 
& \makecell[l]{Does benchmark \(b\)\\ preserve system order?}
& $\sigma_M^2+\sigma_A^2+\sigma_{MA}^2$
& 
    $\dfrac{\sigma_{IM}^2+\sigma_{IA}^2+\sigma_{IMA,e}^2}{n_i}$
& (\ref{eq:perbench-system-reliability}) \\
% & \tableeqnum\label{eq:perbench-system-reliability} \\
\addlinespace
$\rho^{(b)}_{AA'}$ 
% & Benchmark 
& \makecell[l]{Do scaffolds induce \\ the same model order?}
& $\sigma_M^2+\dfrac{\sigma_{IM}^2}{n_i}$
& $\sigma_{MA}^2+\dfrac{\sigma_{IMA,e}^2}{n_i}$ 
% & \tableeqnum\label{eq:interscaffoldrho} \\
& (\ref{eq:interscaffoldrho}) \\
\addlinespace
\makecell[l]{$E\rho^2_{M}$, $\text{SNR}_{M}$} 
% & Leaderboard 
& \makecell[l]{Does the leaderboard \\ preserve model order?}
& $\sigma_M^2$
& $\dfrac{\sigma_{BM}^2}{n_b}+\dfrac{\sigma_{MA}^2}{n_a}+\dfrac{\sigma_{BMA}^2}{n_b n_a}+\dfrac{\sigma_{IM[B]}^2}{n_b n_i}+\dfrac{\sigma_{BIMA,e}^2}{n_b n_i n_a}$ 
% & \tableeqnum\label{eq:dstudy-full} \\
& (\ref{eq:dstudy-full}) \\

$\displaystyle \lim_{n_i\rightarrow\infty}E\rho_{M(\cdot)}^2$
& \makecell[l]{Does having unlimited tasks \\ preserve model order?} 
& $\sigma_M^2$
& $\dfrac{\sigma_{MA}^2}{n_a}$\; or \;  $\frac{\sigma_{BM}^2}{n_b}
+\frac{\sigma_{MA}^2}{n_a}
+\frac{\sigma_{BMA}^2}{n_b n_a}$
& (\ref{eq:task-ceiling}) \\

\makecell[l]{${\Delta\rho^{2}_{(M-A)(b)}}$} 
% & Leaderboard 
& \makecell[l]{Do models have more benchmark-\\relevant signal than scaffolds?}
& $\sigma_M^2-\sigma_A^2$
& $\sigma_{MA}^2+\frac{\sigma_{IM}^2+\sigma_{IA}^2+\sigma_{IMA,e}^2}{n_i}$ 
& \tableeqnum\label{eq:diff_variance_main_bench} \\

\makecell[l]{$\Delta\rho^2_{I(M-A)(b)}$} 
% & Leaderboard 
& \makecell[l]{Do models have more task-\\level signal than scaffolds?}
& $\sigma_M^2+\frac{\sigma_{IM}^2}{n_i}-\sigma_A^2-\frac{\sigma_{IA}^2}{n_i}$
& $\sigma_{MA}^2+\frac{\sigma_{IM}^2+\sigma_{IA}^2+\sigma_{IMA,e}^2}{n_i}$ 
& \tableeqnum\label{eq:diff_variance_task_bench} \\
% & (\ref{eq:dstudy-full}) \\

\makecell[l]{$\Delta\rho^2_{M-A}$} 
% & Leaderboard 
& \makecell[l]{Do models have more leaderboard-\\relevant signal than scaffolds?}
& $\sigma_M^2-\sigma_A^2$
& \makecell{$\sigma_{MA}^2+\frac{\sigma_{BM}^2}{n_b}+\frac{\sigma_{BA}^2}{n_b}+\frac{\sigma_{BMA}^2}{n_b}+ \frac{\sigma_{IM[B]}^2}{n_b n_i}+\frac{\sigma_{IA[B]}^2}{n_b n_i}+\frac{\sigma_{BIMA,e}^2}{n_b n_i}$ }
& \tableeqnum\label{eq:diff_variance_main_lead} \\

\makecell[l]{$\Delta\rho^2_{B(M-A)}$} 
% & Leaderboard 
& \makecell[l]{Do models have more benchmark-\\level signal than scaffolds?}
& $\sigma_M^2+\frac{\sigma_{BM}^2}{n_b}-\sigma_A^2-\frac{\sigma_{BA}^2}{n_b}$
& \makecell{$\sigma_{MA}^2+\frac{\sigma_{BM}^2}{n_b}+\frac{\sigma_{BA}^2}{n_b}+\frac{\sigma_{BMA}^2}{n_b}+ \frac{\sigma_{IM[B]}^2}{n_b n_i}+\frac{\sigma_{IA[B]}^2}{n_b n_i}+\frac{\sigma_{BIMA,e}^2}{n_b n_i}$ }
& \tableeqnum\label{eq:diff_variance_bench_lead} \\

\makecell[l]{$\Delta\rho^2_{I[B](M-A)}$} 
% & Leaderboard 
& \makecell[l]{Do models have more task-\\level signal than scaffolds?}
& \makecell{$\sigma_M^2+\frac{\sigma_{BM}^2}{n_b}+\frac{\sigma_{IM}^2}{n_b n_i}-\sigma_A^2-\frac{\sigma_{BA}^2}{n_b}-\frac{\sigma_{IA}^2}{n_b n_i}$}
& \makecell{$\sigma_{MA}^2+\frac{\sigma_{BM}^2}{n_b}+\frac{\sigma_{BA}^2}{n_b}+\frac{\sigma_{BMA}^2}{n_b}+ \frac{\sigma_{IM[B]}^2}{n_b n_i}+\frac{\sigma_{IA[B]}^2}{n_b n_i}+\frac{\sigma_{BIMA,e}^2}{n_b n_i}$ }
& \tableeqnum\label{eq:diff_variance_task_lead} \\

\bottomrule
\end{tabular}
}
\end{table}

\section{Ablations and Sensitivity Analyses}
\label{app:robustness}

The benchmark ecosystem is sparse, unbalanced, and only partially connected. Tasks are nested within benchmarks, most model--scaffold combinations are absent, and many interaction levels are observed only once. Consequently, a fully crossed variance decomposition could in principle be driven by a small number of influential observations or by a particularly informative benchmark. We therefore examine robustness at four complementary levels: pointwise predictive stability, stability of latent rankings, leave-one-benchmark-out sensitivity, and leave-one-benchmark--scaffold-out sensitivity.

\subsection{Reference indexed variance decomposition}
\label{app:robustness:model}

For observation \(n\), let \(b[n]\), \(i[n]\), \(m[n]\), and \(a[n]\) denote its benchmark, task, model, and scaffold. The reference model with additional indexing from \eqref{eq:pooled} is
\begin{align*}
Y_n &\sim \operatorname{Bernoulli}(p_n), \nonumber\\
\operatorname{logit}(p_n)
&=
\mu
+u^{B}_{b[n]}
+u^{I}_{b[n],i[n]}
+u^{M}_{m[n]}
+u^{A}_{a[n]} \nonumber\\
&\quad
+u^{BM}_{b[n],m[n]}
+u^{BA}_{b[n],a[n]}
+u^{MA}_{m[n],a[n]} \nonumber\\
&\quad
+u^{IM}_{b[n],i[n],m[n]}
+u^{IA}_{b[n],i[n],a[n]}
+u^{BMA}_{b[n],m[n],a[n]},
% \label{eq:robustness-full-model}
\end{align*}
where every random effect is independently distributed as
\(
u^F_{\ell}\sim\mathcal N(0,\sigma_F^2)
\)
for facet or interaction \(F\). On the latent logistic scale, the observation-level residual variance is \(\pi^2/3\). The reference fit contains \(N=29{,}923\) binary observations, 54 models, 13 scaffolds, nine benchmarks, and 1,117 benchmark-specific tasks.

The central object of measurement is the model. For a fixed evaluation design \(D\), the relative generalizability coefficient has the schematic form from Equation \ref{eq:relative-reliability}
\begin{equation}
\rho_D^2
=
\frac{\sigma_M^2}
{\sigma_M^2+\sigma_{\delta,D}^2},
\label{eq:relative-generic}
\end{equation}
where \(\sigma_{\delta,D}^2\) contains only those components that change model contrasts under repeated realizations of \(D\), divided by their effective replication counts. Components that shift all models equally under a shared condition do not enter the relative error variance.

\paragraph{Cancellation of common condition effects.}
For two models \(m\) and \(m'\) evaluated under the same benchmark and scaffold,
\[
\eta_{bima}-\eta_{bim'a}
\]
eliminates the benchmark main effect \(u^B_b\), scaffold main effect \(u^A_a\), benchmark--scaffold effect \(u^{BA}_{ba}\), task--scaffold effect \(u^{IA[B]}_{bia}\), and task main effect \(u^I_{bi}\). Thus, these components can affect absolute scores but not the ordering of models evaluated under identical conditions. This distinction is important below: instability in \(\sigma_A^2\) or \(\sigma_{BA}^2\) need not imply instability in relative model reliability.

All posterior chains for the reference model mixed adequately: the reported variance parameters had \(\widehat R=1.00\), with bulk effective sample sizes between 1,559 and 3,743. The following analyses address robustness to the data and specification rather than only Monte Carlo convergence.

\subsection{Sensitivity of model rankings}
\label{app:ranking-sensitivity}

Standard leaderboards rank systems by empirical mean accuracy,\footnote{e.g., for HAL see ``Accuracy'' at \url{https://hal.cs.princeton.edu/reliability/}\, for HELM see ``mean score'' at \url{https://crfm.stanford.edu/helm/}, etc. }
\[
\bar y_{bm}
=
\frac{1}{N_{bm}}\sum_{n:\,b[n]=b,m[n]=m}y_n,
\]
possibly aggregating over different scaffolds and nonidentical task sets. Such rankings treat all observed variation as evidence about model capability. The mixed models instead estimate latent model effects after separating task difficulty, scaffold effects, and their interactions.

For each benchmark, we formed rankings from posterior summaries of the relevant latent model effects. We considered two estimators:

\begin{enumerate}
    \item \textbf{Full-model latent ranking}, obtained from the joint model in Equation~\ref{eq:pooled}, which partially pools information through models and scaffolds appearing elsewhere in the benchmark ecosystem.
    \item \textbf{Per-benchmark latent ranking}, obtained by fitting, within each benchmark,
    \[
    \operatorname{logit}(p_{ima})
    =
    \mu+u^I_i+u^M_m+u^A_a
    +u^{IM}_{im}+u^{IA}_{ia}+u^{MA}_{ma}.
    \]
    This estimator uses only the information available within one benchmark.
\end{enumerate}

These posterior conditional means are the Bayesian analogue of best linear unbiased predictors (BLUPs). We compared their induced rankings with the rankings from observed mean scores using Kendall's \(\tau\), Spearman's \(\rho\), and bias-corrected squared distance correlation \(\text{dCor}^2_n\) computed on rank vectors. Kendall's \(\tau\) measures pairwise ordering agreement; Spearman's \(\rho\) measures monotone rank association; and corrected squared distance correlation can detect more general dependence between the rank assignments.

\begin{table*}[t]
\centering
\caption{Association between observed rankings and latent-effect rankings. ``Full'' uses the ecosystem model (Eq. \ref{eq:pooled}); ``Per-benchmark'' uses an independently fitted model for each benchmark (Eq. \ref{eq:perbenchmark})}
\label{tab:rank-robustness}
\resizebox{\textwidth}{!}{
\begin{tabular}{llrrrrrrrrr}
\toprule
Estimator & Metric
& AssistantBench & CORE-Hard & GAIA & Mind2Web & SciCode
& ScienceAgentBench & SWE-mini & \(\tau\)-bench & USACO\\
\midrule
% Full & dCor
% & .668 & .955 & .895 & .930 & .848 & .822 & .917 & .919 & .759\\
% % & .871 & .904 & .762 & .514 & .930 & .451 & .877 & .530 & .898\\
% Per-benchmark & dCor
% & .129 & .352 & .256 & .254 & .186 & .171 & .255 & .266 & .240\\
% \addlinespace

Full & dCor${}^2_n$
& .404 & .899 & .780 & .872 & .703 & .651 & .833 & .828 & .430\\
% & .871 & .904 & .762 & .514 & .930 & .451 & .877 & .530 & .898\\
Per-benchmark & dCor${}^2_n$
& -.005 & .107 & .046 & .039 & .013 & .007 & .046 & .051 & .032\\
\addlinespace

Full & Kendall \(\tau\)
& .478 & .859 & .771 & .744 & .731 & .582 & .764 & .787 & .590\\
% & .683 & .748 & .448 & .385 & .775 & .218 & .691 & .421 & .769\\
Per-benchmark & Kendall \(\tau\)
& .081 & .248 & .196 & .151 & .128 & .101 & .167 & .180 & .142\\
\addlinespace
Full & Spearman \(\rho\)
& .619 & .961 & .900 & .896 & .867 & .795 & .916 & .930 & .731\\
% & .859 & .909 & .688 & .363 & .918 & .361 & .855 & .523 & .901\\
Per-benchmark & Spearman \(\rho\)
& .111 & .340 & .271 & .211 & .171 & .137 & .236 & .255 & .199\\
\bottomrule
\end{tabular}}
% \caption*{\small For most benchmarks we see that general rank order (Spearman) and distribution are preserved (dCor) Rank and Distance Correlations between Original Rankings and Best Linear Unbiased Predictor (BLUP) Rankings taken from the posterior.}
\end{table*}

Two findings are salient. First, the jointly estimated rankings retain substantial agreement with observed leaderboards for CORE-Bench Hard, GAIA, Online-Mind2Web, SciCode, SWE-bench Verified Mini, and $\tau$-Bench Airline. For example, full-model Spearman correlations are at least \(0.85\) on these benchmarks. Their empirical rankings therefore contain recoverable cross-system signal even after nuisance variation is separated.

% Second, agreement is appreciably weaker for Online-Mind2Web, ScienceAgentBench, and \(\tau\)-bench Airline. ScienceAgentBench has a full-model Kendall correlation of only \(0.218\), implying that many pairwise model comparisons change after partial pooling and adjustment for evaluation conditions. These benchmarks should not be interpreted as producing precise total orders: the data more strongly support uncertainty sets or posterior pairwise comparisons than a single deterministic ranking.

Second, the independently fitted per-benchmark models exhibit weaker correspondence with raw rankings on every benchmark. This result is not evidence that the per-benchmark models are necessarily ``less accurate.'' Instead, it exposes an information and estimand mismatch. Within one sparse benchmark, model effects must be separated from model--scaffold and task--model interactions using few connected observations. Strong posterior shrinkage is therefore appropriate, but it compresses distinctions that empirical average accuracy treats as model signal. Moreover, the raw ranking may combine deployable model--scaffold performance, whereas the latent model ranking intentionally removes scaffold-specific contributions. The joint model can recover more stable model effects because shared models and scaffolds connect otherwise isolated benchmark-specific designs. In other words, the combined model, after accounting for sources of variation, provides more stable and reliable estimates by taking advantage of shared facet variation.

Accordingly, rank correlation with the observed leaderboard is a sensitivity diagnostic, not a ground-truth accuracy measure. High agreement indicates that adjustment for known nuisance facets preserves the reported ordering. Low agreement indicates that the ordering depends on whether task and scaffold variation are treated as capability or as measurement error. It does not establish which ordering is externally valid without an independent criterion.

\subsection{Pointwise predictive stability with PSIS-LOO}
\label{app:psis-loo}

We first assessed whether individual observations exert disproportionate influence on the posterior predictive distribution. Exact leave-one-out cross-validation would require \(N\) refits. We instead used Pareto-smoothed importance-sampling leave-one-out cross-validation (PSIS-LOO), with moment matching for observations having unstable importance ratios \citep{vehtari_practical_2017, paananen_implicitly_2021,vehtari_pareto_2024}. For observation \(n\), PSIS approximates
\[
p(y_n\mid y_{-n})
=
\int p(y_n\mid\theta)\,p(\theta\mid y_{-n})\,d\theta
\]
using draws from the full posterior \(p(\theta\mid y)\). The generalized Pareto shape diagnostic \(\widehat k_n\) measures the tail behavior of the resulting importance ratios. Values \(\widehat k_n\leq 0.7\) generally indicate a reliable approximation, whereas larger values identify observations for which deleting the point substantially changes its posterior predictive distribution.

\begin{table}[t]
\centering
\caption{PSIS-LOO diagnostics for the full Bayesian logistic mixed model.}
\label{tab:psis-loo}
\begin{tabular}{lrr}
\toprule
Quantity & Estimate & Standard error\\
\midrule
\(\operatorname{elpd}_{\mathrm{loo}}\) & \(-11{,}217.1\) & \(94.2\)\\
\(p_{\mathrm{loo}}\) & \(3{,}032.0\) & \(32.0\)\\
\(\operatorname{LOOIC}\) & \(22{,}434.2\) & \(188.5\)\\
\midrule
Pareto diagnostic & Count & Percentage\\
\(\widehat k\leq 0.7\) & \(29{,}902\) & \(>99.9\%\)\\
\(0.7<\widehat k\leq 1\) & \(21\) & \(<0.1\%\)\\
\(\widehat k>1\) & \(0\) & \(0.0\%\)\\
\bottomrule
\end{tabular}
\end{table}

The diagnostics in Table~\ref{tab:psis-loo} are favorable for a model of this complexity. The effective number of parameters is substantially smaller than both \(N\) and the nominal number of coefficients induced by the random effects, demonstrating strong regularization through partial pooling. No observation has \(\widehat k>1\), and only 21 of 29,923 observations have \(\widehat k>0.7\).

The few warnings are explained by the connectivity of the design rather than by broad model failure. Approximately \(32\%\) of observed benchmark--task--model groups are singletons, a known point of sensitivity for PSIS-LOO \citep{bindoff_partial_2026}. Of the 21 observations with \(\widehat k>0.7\), 19 (\(90.5\%\)) belong to such singleton groups. The remaining two form the only observations in their group: SciCode task 14 evaluated with GPT-4.1 under two scaffolds.

This behavior follows directly from the pointwise LOO target. If \(y_n\) is the only observation informing a random-effect level \(u_\ell\), deletion of \(y_n\) makes the leave-one-out distribution of \(u_\ell\) approximately prior-predictive. The full-data posterior, by contrast, has adapted \(u_\ell\) to \(y_n\). Importance sampling must therefore bridge two meaningfully different distributions, producing a heavy-tailed importance ratio. A large \(\widehat k_n\) in this setting primarily says that the outcome of an isolated, data-defined group cannot be predicted after removing its only observation. It does not, by itself, imply that population-level variance components are determined by that point.

\subsubsection{Observation-level random-effect ablation.}\label{app:olre}
As an additional check, we augmented Equation~\ref{eq:pooled} with
\[
u^{IMA}_{b,i,m,a}\sim\mathcal N(0,\sigma_{IMA}^2).
\]
Because benchmark--task--model--scaffold cells generally lack within-cell replication in the HAL dataset, this term acts as an observation-level random effect (OLRE) \citep{harrison_comparison_2015}. In a Bernoulli-logit model, it competes with the fixed latent logistic residual rather than identifying a conventionally replicated three-way interaction. Including it did not materially change the reliability estimates: it absorbed variation previously treated as observation-level noise, and its contribution is attenuated by task replication in the D-study. The principal generalizability conclusions are therefore not an artifact of omitting the highest-order cell term.

Importantly, this additional term should be used for studies where output stability is of interest where benchmark--task--model--scaffold levels have multiple runs. This variation can support measuring the stochasticity of task scores under the same conditions.

\subsection{Leave-one-benchmark-out sensitivity}
\label{app:loo-benchmark}

Pointwise LOO evaluates interpolation within the observed ecosystem. A stronger test removes an entire benchmark and therefore deletes all of its tasks, score distribution, and benchmark-specific interactions. For each \(b^\star\in\mathcal B\), we refitted the complete variance decomposition to
\[
\mathcal D_{-b^\star}
=
\{y_n:b[n]\neq b^\star\}
\]
and recomputed the posterior variance decomposition and D-study reliability curves. This procedure asks whether conclusions about the benchmark battery are broadly distributed across benchmarks or are driven by a single test.

\begin{figure}
    \centering
    \includegraphics[width=1\linewidth]{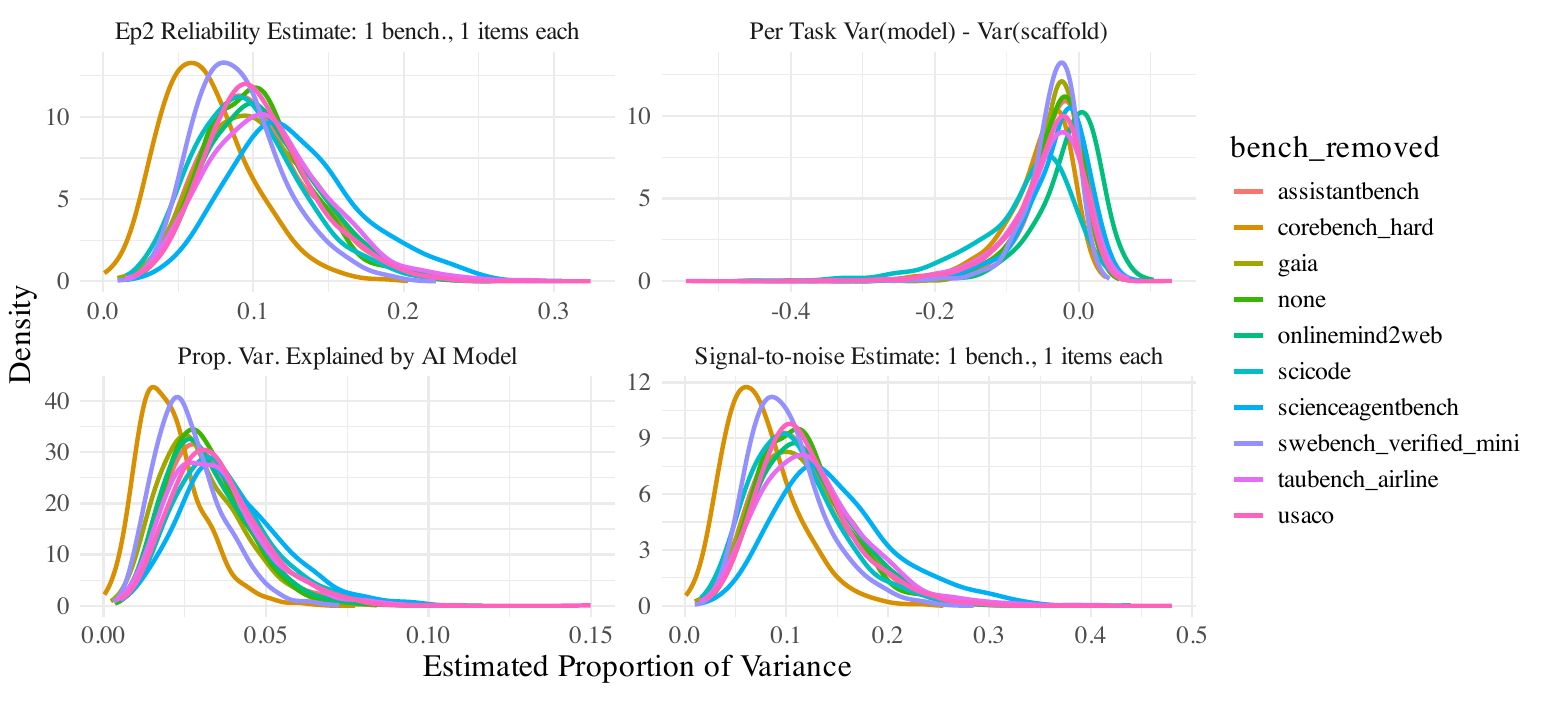}
    \caption{Posterior distributions for benchmark-level LOO ablations for the paper's quantities of interest}
    \label{fig:looposterior}
\end{figure}

The reliability estimations, proportions of variance attributable to models, and the principal interaction facets were generally stable across these refits as shown in Fig. \ref{fig:looposterior}. Sensitivity nevertheless depended on which benchmark was withheld. Removing CORE-Bench Hard caused the largest reduction in estimated reliability. This is substantively expected: CORE-Bench Hard supplies the most runs, spans the widest range of observed accuracies, and has the highest estimated signal-to-noise ratio among the included benchmarks. It therefore contributes unusually strong information both for distinguishing models and for connecting model performance to the rest of the ecosystem.

Without CORE-Bench Hard, the posterior signal-to-noise ratio from the D-study no longer reaches the prespecified limit-of-detection band. This finding should not be read merely as ``more observations are better.'' A large but noisy benchmark can contribute little to the reliability of a battery. CORE-Bench Hard is influential because it combines replication with discrimination. Thus, increasing the number of benchmarks is not equivalent to increasing effective measurement information.

More generally, for a battery aggregating conditionally independent benchmark-level measurements with signal \(S_b\) and error \(E_b\), the information supplied by benchmark \(b\) is governed by its discrimination relative to error, not simply its task count. Although the fully crossed design includes interactions and is more complicated than this schematic case, the same principle applies: benchmarks with substantial model variation and controlled task- and harness-dependent error contribute disproportionately to stable ecosystem-level rankings.

This ablation also clarifies the intended scope of generalization. The leave-one-benchmark-out results do not claim that the fitted model can predict an arbitrary future benchmark with no shared structure. Rather, they test whether the estimated variance decomposition and design recommendations survive removal of one observed measurement instrument. The sensitivity to CORE-Bench Hard indicates that the present nine-benchmark ecosystem contains limited redundancy at the high-signal end.

\subsection{Leave-one-benchmark--scaffold-out sensitivity}
\label{app:loo-benchmark-agent}

Because scaffold coverage is also unbalanced, we conducted a complementary sensitivity analysis that removed one observed benchmark--scaffold combination at a time and re-estimated the full decomposition. Table~\ref{tab:loo-ba} summarizes the resulting frequentist proportions of total variance. This ablation is especially stringent for benchmark-specific scaffolds, whose removal can eliminate nearly all direct information about a scaffold level. 

\begin{table}[t]
\centering
\caption{Leave-one-benchmark--scaffold-out proportions of variance. SE is the standard error across refits, and CV is the coefficient of variation.}
\label{tab:loo-ba}
\resizebox{\columnwidth}{!}{
\begin{tabular}{lrrrrlrr}
\toprule
Facet & Mean & Median & Min. & Max. &  SD&SE & CV\\
\midrule
Residual & .237 & .235 & .226 & .267 &  .008&.002 & .035\\
Scaffold (\(A\)) & .020 & .019 & .008 & .037 &  .007&.002 & .338\\
Benchmark (\(B\)) & .258 & .259 & .229 & .298 &  .015&.003 & .060\\
Benchmark--scaffold (\(BA\)) & .026 & .028 & .000 & .035 &  .009&.002 & .345\\
Benchmark--model (\(BM\)) & .020 & .020 & .009 & .031 &  .004&.001 & .208\\
Benchmark--model--scaffold (\(BMA\)) & .014 & .013 & .010 & .019 &  .003&.001 & .180\\
Task within benchmark (\(I\)) & .328 & .329 & .282 & .365 &  .015&.003 & .047\\
Task--scaffold (\(IA\)) & .052 & .053 & .028 & .066 &  .009&.002 & .168\\
Task--model (\(IM\)) & .003 & .004 & .000 & .006 &  .001&\(<.001\) & .426\\
Model (\(M\)) & .033 & .033 & .026 & .044 &  .004&.001 & .110\\
Model--scaffold (\(MA\)) & .008 & .009 & .003 & .012 &  .002&\(<.001\) & .249\\
\bottomrule
\end{tabular}}
\end{table}

\begin{table}[!h]
\centering
\caption{Leave-one-LLM-out proportions of variance. SD is the standard deviation across refits, SE is the standard error across refits, and CV is the coefficient of variation.}
\label{tab:loo-m}
\resizebox{\ifdim\width>\linewidth\linewidth\else\width\fi}{!}{
\begin{tabular}{lrrrrrrr}
\toprule
Facet & Mean & Median & Min. & Max. & SD & SE & CV\\
\midrule
Residual & 0.236 & 0.236 & 0.231 & 0.241 & 0.002 & 0 & 0.008\\
Scaffold (\(A\)) & 0.020 & 0.020 & 0.016 & 0.029 & 0.002 & 0 & 0.098\\
Benchmark (\(B\)) & 0.259 & 0.259 & 0.252 & 0.268 & 0.002 & 0 & 0.009\\
Benchmark--scaffold (\(BA\)) & 0.026 & 0.026 & 0.022 & 0.030 & 0.001 & 0 & 0.056\\
Benchmark--model (\(BM\)) & 0.020 & 0.020 & 0.015 & 0.022 & 0.001 & 0 & 0.065\\
Benchmark--model--scaffold (\(BMA\)) & 0.014 & 0.014 & 0.008 & 0.016 & 0.001 & 0 & 0.092\\
Task within benchmark (\(I\)) & 0.328 & 0.328 & 0.320 & 0.334 & 0.002 & 0 & 0.006\\
Task--scaffold (\(IA\))  & 0.052 & 0.052 & 0.049 & 0.053 & 0.001 & 0 & 0.018\\
Task--model (\(IM\)) & 0.004 & 0.004 & 0.001 & 0.004 & 0.001 & 0 & 0.166\\
Model (\(M\))  & 0.033 & 0.033 & 0.028 & 0.036 & 0.002 & 0 & 0.048\\
Model--scaffold (\(MA\)) & 0.009 & 0.009 & 0.006 & 0.012 & 0.001 & 0 & 0.107\\
\bottomrule
\end{tabular}}
\end{table}

The dominant components are stable. Task difficulty accounts for approximately \(32.8\%\) of variance across refits, benchmark differences for \(25.8\%\), and residual variation for \(23.7\%\). Their coefficients of variation are only \(0.047\), \(0.060\), and \(0.035\), respectively. The model component is also stable in absolute terms, ranging from \(0.026\) to \(0.044\).

The largest relative variation occurs for the scaffold main effect, the benchmark--scaffold interaction, and the task--model interaction. These cases require different interpretations. The first two are weakly identified because most scaffolds occur in only one benchmark; removing a benchmark--scaffold cell can therefore remove much of the relevant connectivity. However, common scaffold and benchmark--scaffold shifts cancel from same-condition model contrasts and consequently do not enter the relative model reliability estimates used in the primary D-study.

The task--model component has the largest coefficient of variation (\(0.426\)), but its estimated proportion is always between \(0\) and \(0.006\). Its high relative variability is therefore primarily a small-denominator effect: even its maximum is smaller than the typical contribution of every other reported component. Coefficients of variation should not be interpreted without the corresponding absolute scale. Additionally, both this component and the second largest CV value, the benchmark--scaffold component, are the only components that contain minimum values equal to zero; these sensitivities, paired with extreme ablation minima, are likely elevated due to the nature of frequentist estimations, which can result in singular values in highly unbalanced designs (see Appendix \ref{app:estim_choices}). These estimates may be more stable under a full Bayesian ablation where CV estimates would be less likely to decrease the mean in the denominator.

The model-related components that directly govern ranking stability remain comparatively well behaved. The model proportion has CV \(0.110\), while the benchmark--model and model--scaffold components remain small across all deletions. Hence, no individual benchmark--scaffold cell appears to create the principal conclusion that bare-model rankings are less reliable than rankings would appear under a decomposition that treats all observed model--scaffold performance as signal. However, it is worth noting that removing AssistantBench during the LOO ablation disconnects the OnlineMind2Web scaffold from the broader network, which means that, for that connection, identifiability conditions found in Proposition \ref{prop:suff_estim} would not be complete. Nevertheless, the estimates remain stable, even with this disconnection.

Importantly, the contribution to overall reliability of CORE-Bench Hard discussed in \S~\ref{app:loo-benchmark} can be narrowed further to the contribution of the CORE Agent, which shows the most between-LLM discrimination on its benchmark. Removing this particular benchmark--scaffold accounts for the large increase in uncertainty across the pooled model. As this scaffold is less connected than the HAL Generalist scaffold, this reaffirms that having a strong discriminating signal is critical to pooled reliability, potentially more than large connectivity.

\subsection{Leave-one-LLM-out sensitivity}
\label{app:loo-model}

We next test whether the estimated measurement structure is driven by a single LLM and whether the variance decomposition generalizes to new LLMs. For each LLM model \(m^\star\in\mathcal M\), we removed all of its observations,
\[
\mathcal D_{-m^\star}
=
\{y_n:m[n]\neq m^\star\},
\]
and refit the variance-decomposition model. This is a stronger perturbation than deleting one response: it simultaneously removes a model main-effect level and every observed benchmark--model, model--scaffold, task--model, and benchmark--model--scaffold cell involving that LLM. Because evaluation coverage differs substantially across models---only nine of the 54 models appear in every benchmark---the ablation also tests sensitivity to the most highly connected models in the design. The LOO ablations for this and those of Appendix \ref{app:loo-benchmark-agent} were conducted in a frequentist framework to reduce practical computation costs of hundreds of Bayesian refittings (see Appendix \ref{app:compute}).

Table~\ref{tab:loo-m} shows that the decomposition is remarkably insensitive to the removal of any LLM. The three largest components are almost invariant: task-within-benchmark variance has mean proportion \(0.328\), range \(0.320\)--\(0.334\), and CV \(0.006\); residual variance has mean \(0.236\), range \(0.231\)--\(0.241\), and CV \(0.008\); and benchmark variance has mean \(0.259\), range \(0.252\)--\(0.268\), and CV \(0.009\). Thus, the conclusion that variation is dominated by differences among tasks and benchmarks, together with substantial unexplained response variation, is not attributable to the performance profile of a particular model.

More importantly for relative reliability, the model variance is also stable. Its proportion remains between \(0.028\) and \(0.036\), with mean \(0.033\) and CV \(0.048\). The benchmark--model component ranges from \(0.015\) to \(0.022\), while the benchmark--model--scaffold component ranges from \(0.008\) to \(0.016\). Removing even a highly connected or unusually capable LLM therefore does not qualitatively change the estimated amount of systematic between-model variation or the extent to which model performance depends on the benchmark and scaffold.

As in the benchmark--scaffold ablation, the largest relative variability occurs in small components. The task--model interaction has CV \(0.166\), but its variance share is only \(0.001\)--\(0.004\). The model--scaffold interaction has CV \(0.107\) and remains between \(0.006\) and \(0.012\). Their relative sensitivity should consequently not be confused with a large contribution to total variance. In absolute terms, deletion-induced changes in both components are small.

The leave-one-model results are also more stable than the leave-one-benchmark results. This asymmetry reflects the structure of the available evidence. Each LLM contributes another sample from the population of systems, and its outcomes are partially pooled with those of the remaining 53 models. By contrast, removing a benchmark deletes an entire measurement instrument, all of its unique tasks, and its characteristic signal-to-noise ratio. The ecosystem therefore has greater redundancy across models than across high-quality benchmarks. Adding another LLM primarily improves estimation of the distribution of model capability, whereas adding a discriminating benchmark can alter the quality of the measurement battery itself.

\paragraph{Implication for reliability.}
For a fixed D-study design, relative reliability depends on the model variance and on model-dependent error components such as \(BM\), \(MA\), \(IM\), and \(BMA\), after scaling by their effective replication counts. The stability of these components under model deletion implies that the reported reliability conclusions are not generated by one extreme or unusually well-connected LLM. Nevertheless, this ablation evaluates \emph{influence on the population variance decomposition}, not the ability to predict the performance or rank of a previously unseen model. Generalization to a new LLM additionally requires that it be exchangeable with the sampled model population and evaluated on conditions that connect it to the existing design.

No single LLM acts as a leverage point for the principal variance decomposition. The greater sensitivity to removing an informative benchmark than to removing an LLM reinforces a central design recommendation: once a reasonably diverse model sample has been obtained, additional evaluation resources may yield greater reliability gains by improving benchmark quality, task replication, and cross-scaffold connectivity than by adding sparsely evaluated models.

\subsection{Methodological Robustness}
We also estimate each model using five different approaches. A separate Appendix \ref{app:method_comps} explains and discusses each method and reports the full results.

\subsection{Rank stability across posterior}\label{app:rank_stability}

Ranking models within each draw yields posterior rank distributions and pairwise ordering probabilities
\(
\Pr(\theta_{mb}>\theta_{m'b}\mid y).
\)
These quantities distinguish an estimated ordering from evidence that two models are meaningfully distinguishable. We compare posterior and published rankings using Spearman correlation and ranked unbiased squared distance correlation, \(\operatorname{dCor}_n^2\); the latter remains informative in the presence of extensive ties.

We compare posterior ranks with published ranks using Spearman correlation and ranked unbiased squared distance correlation, dCor${}^2_n$ (Table \ref{tab:corrs}). The latter is useful for benchmarks with extensive ties and provides a direct diagnostic of whether estimated capability ranking is statistically associated with the reported ordering. 

Posterior capability estimates differ from ranks obtained by sorting raw percent-correct scores. Raw scores credit the model for every condition with which it happens to be paired. The pooled decomposition instead estimates the portion of performance that persists after averaging over the specified benchmark and scaffold universes.

\begin{table}[!h]
\centering
\caption{Median correlation of reported and estimated ranks across posterior draws}
\label{tab:corrs}
\resizebox{\ifdim\width>\linewidth\linewidth\else\width\fi}{!}{
\begin{tabular}{lrrrrrrrrr}
\toprule
 Corr.&AssistantBench & CORE-Bench & GAIA & Mind2web & SciCode & Sci.AgentBench & SWEbench & $\tau$-bench & USACO\\
\midrule
 Spearman&0.442 & 0.824 & 0.79 & 0.621 & 0.662 & 0.611 & 0.808 & 0.683 & 0.692\\
 dCor${}^2_n$&0.141 & 0.64 & 0.576 & 0.33 & 0.36 & 0.323 & 0.633 & 0.408 & 0.395\\
\bottomrule
\end{tabular}}
\end{table}

\subsection{Implications for benchmark design}
\label{app:robustness-implications}

These checks support five practical conclusions.

\begin{enumerate}
    \item \textbf{Sparse cells principally limit local prediction.}
    PSIS warnings are almost entirely confined to singleton or doubleton interaction levels. The model cannot predict an isolated cell after its sole observation is removed, but the population-level variance decomposition is stable to these observations.

    \item \textbf{Partial pooling is necessary for ecosystem-level ranking.}
    Per-benchmark data are generally insufficient to cleanly distinguish model capability from scaffold and task interactions. Cross-benchmark connectivity substantially stabilizes latent model rankings, although the resulting rankings remain uncertain on low-signal benchmarks.

    \item \textbf{Benchmark quality is not interchangeable with benchmark quantity.}
    The leave-one-benchmark-out analysis identifies CORE-Bench Hard as a high-information anchor. A useful test battery should include multiple independently constructed benchmarks with high discrimination and controlled error, rather than merely adding more noisy tasks or near-duplicate benchmarks.

    \item \textbf{Absolute-score instability and relative-rank instability are distinct.}
    Scaffold and benchmark--scaffold effects can materially shift reported accuracies while canceling from same-condition model comparisons. Benchmark reports should therefore state whether reliability concerns absolute deployment performance, model ranking, or model--scaffold system ranking; these are different objects of measurement and induce different error terms.
    \item \textbf{Representative LLM panel improves benchmark diagnostics.} once a reasonably diverse model sample has been obtained, additional evaluation resources may yield greater reliability gains by improving benchmark quality, task replication, and cross-scaffold connectivity than by adding sparsely evaluated models.
\end{enumerate}

The robustness analyses do not imply that the sparse design is harmless. Rather, they localize its consequences. The main variance and reliability conclusions are not driven by a handful of observations or benchmark--scaffold cells, but the precision of individual rankings remains strongly dependent on cross-benchmark connectivity and on the inclusion of at least one high-signal measurement instrument. This distinction is essential for designing future agentic benchmark batteries: additional evaluation should be allocated to conditions that improve connectivity and reduce model-dependent error, not only to increasing the nominal number of tasks or sparsely connecting models.

\section{Identifiability and connectivity of the leaderboard decomposition}
\label{app:connectivity}

The leaderboard data are sparse by construction: models are submitted selectively, scaffolds are not used uniformly, and tasks are unique to benchmarks. Such missingness is common in public leaderboards~\citep{singh_leaderboard_2026}. A fully crossed experiment would simplify estimation, but it is not necessary for the questions studied here. What is required is sufficient overlap to distinguish persistent model variation from variation associated with benchmarks, scaffolds, and their interactions.

This appendix establishes that the observed design provides the connectivity needed for those contrasts. We distinguish three concepts that are often conflated:

\begin{enumerate}[leftmargin=0.6cm,nosep]
    \item \textbf{Connectivity:} whether observed model, benchmark, and scaffold levels belong to a common comparison network.
    \item \textbf{Structural identifiability:} whether distinct variance components imply distinct distributions over the observed responses.
    \item \textbf{Practical estimability:} whether the finite data determine those components precisely.
\end{enumerate}

Connectivity and structural identifiability are design properties; practical estimability also depends on sample size, outcome variation, and prior regularization. Our Bayesian estimator can produce a proper posterior for a weakly informed component, but a proper posterior alone is not evidence that the component is strongly data-identified. We therefore use posterior uncertainty and sensitivity analyses, rather than existence of an estimate, to characterize practical estimability.

\subsection{Incidence-Graph Connectivity and Identifiability}
% \label{app:connectivity}

Represent the observed design as a multipartite incidence graph whose vertices
are benchmarks, models, and scaffolds, with edges induced by observed
evaluations. Shared models and scaffolds connect benchmark-specific
observations and support estimation of common variance components. Contrasts
within a connected component are informed by observed paths; contrasts across
disconnected components are not identified without additional assumptions.
We report component membership, articulation vertices, benchmark degrees, and
the change in connectivity produced by removing each benchmark. Posterior
regularization stabilizes weakly supported components but does not create
evidence for disconnected contrasts.

\subsection{Observed incidence structure}
\label{app:connectivity-design}

The full dataset contains nine benchmarks, 54 LLMs, and 13 scaffolds. Nine models appear on every benchmark and span at least \(69\%\) of the scaffold set. One scaffold is shared by eight benchmarks. AssistantBench connects the remaining benchmark through an additional shared scaffold, and every other benchmark contains at least one additional benchmark-specific scaffold. Models and scaffolds are consequently neither fully crossed nor evenly replicated.

Let
\[
\Omega
=
\left\{(b,i,m,a):
y_{bima}\ \text{is observed}\right\}
\]
denote the observed response cells. The pooled decomposition is \eqref{eq:pooled}:
\begin{align*}
\eta_{bima}
={}&
\beta_0
+u_b^{(B)}
+u_{bi}^{(I[B])}
+u_m^{(M)}
+u_a^{(A)}
+u_{bm}^{(BM)}
+u_{ba}^{(BA)}
+u_{ma}^{(MA)}
\nonumber\\
&+u_{bim}^{(IM[B])}
+u_{bia}^{(IA[B])}
+u_{bma}^{(BMA)},
\qquad (b,i,m,a)\in\Omega .
% \label{eq:connectivity-pooled}
\end{align*}
Each random effect has mean zero and a component-specific variance. The Bernoulli--logit likelihood fixes the latent scale through the standard logistic residual variance \(\pi^2/3\).

It is useful to represent \(\Omega\) as a multipartite incidence graph. Let
\[
G=(V,E),\qquad
V=\mathcal{B}\cup\mathcal{M}\cup\mathcal{A},
\]
where an edge joins two levels when they co-occur in at least one observed response. For example, \(m\)--\(b\) is an edge if model \(m\) is evaluated on benchmark \(b\), and \(a\)--\(b\) is an edge if scaffold \(a\) is used on benchmark \(b\). Items need not connect across benchmarks because they are intentionally nested within benchmark.

The nine models observed on every benchmark form model-side anchors. The scaffold used on eight benchmarks forms a scaffold-side anchor, while the second shared scaffold connects AssistantBench to the rest of the graph. Benchmark-specific scaffolds are leaves or local branches attached to this connected core. Thus, all benchmarks and their associated observations belong to one comparison network.

\paragraph{Lemma 1 (connected additive contrasts).}

\begin{lemma}[connected additive contrasts]\label{lem:connectadditive}
    Consider an additive model on an observed incidence graph,
\[
\eta_{ma}=\mu+\alpha_m+\gamma_a,
\]
with one centering constraint per facet. If the model--scaffold incidence graph is connected, all estimable contrasts
\(\alpha_m-\alpha_{m'}\) and \(\gamma_a-\gamma_{a'}\) are identified.
\end{lemma}

% \paragraph{Proof.}
\begin{proof}
    Suppose two parameterizations produce the same linear predictor on every observed edge. Their differences satisfy
\(\Delta\alpha_m+\Delta\gamma_a=0\) on each edge. Along any path in a connected bipartite graph, these equalities imply that all model differences equal a common constant and all scaffold differences equal its negative. Centering removes this remaining additive degree of freedom.
\end{proof}
 % \(\square\)

Lemma~\ref{lem:connectadditive} concerns fixed additive effects, whereas
Equation~\ref{eq:pooled} uses random effects and interactions. It nevertheless provides the relevant intuition: an observation from a model or scaffold that is disconnected from the remainder of the leaderboard cannot support a common ranking. Shared models and scaffolds create paths along which relative effects can be compared.

\subsection{Identifiability of random-effect variances}
\label{app:variance-identifiability}

For observations ordered as a vector, write the latent linear predictor as
\[
\boldsymbol{\eta}
=
\beta_0\mathbf{1}
+\sum_{k=1}^{K} Z_k\mathbf{u}_k,
\qquad
\mathbf{u}_k\sim
\mathcal{N}(\mathbf{0},\sigma_k^2 I),
\]
where \(Z_k\) is the incidence matrix for component \(k\). On the latent scale, the random effects induce covariance
\begin{equation}
\operatorname{Cov}(\boldsymbol{\eta})
=
\sum_{k=1}^{K}\sigma_k^2 K_k,
\qquad
K_k=Z_kZ_k^\top .
\label{eq:covariance-kernels}
\end{equation}
The entries of \(K_k\) indicate which pairs of observations share a level of facet or interaction \(k\). For example, two observations share the model kernel \(K_M\) when they use the same model, and share \(K_{BM}\) only when they use both the same benchmark and model.
\begin{proposition}[variance-component criterion]\label{prop:vc_criterion}
A collection of latent variance components
\(\{\sigma_k^2:k\in\mathcal{K}\}\) is structurally identifiable from the observed design if its covariance kernels
\(\{K_k:k\in\mathcal{K}\}\), restricted to \(\Omega\), are linearly independent after removing components that are deterministically confounded with the terminal residual.
\end{proposition}

% \paragraph{Proposition 1 (variance-component criterion).}
% A collection of latent variance components
% \(\{\sigma_k^2:k\in\mathcal{K}\}\) is structurally identifiable from the observed design if its covariance kernels
% \(\{K_k:k\in\mathcal{K}\}\), restricted to \(\Omega\), are linearly independent after removing components that are deterministically confounded with the terminal residual.

% \paragraph{Proof.}

\begin{proof}
    If the restricted kernels are linearly independent, equality of two induced covariance matrices implies
\[
\sum_{k\in\mathcal{K}}
(\sigma_k^2-\widetilde{\sigma}_k^2)K_k=0,
\]
which has only the trivial solution
\(\sigma_k^2=\widetilde{\sigma}_k^2\) for every \(k\). If the kernels are linearly dependent, a nonzero perturbation of their coefficients leaves the induced covariance unchanged, so the corresponding components cannot be separated from the observed design.
\end{proof}
 % \(\square\)

For a Bernoulli GLMM, Proposition \ref{prop:vc_criterion} is most directly interpreted as a design criterion on the latent scale. The nonlinear likelihood can affect the amount of information, especially under floor effects, but it cannot create distinctions absent from the incidence matrices.

A practical interpretation is the following: to distinguish two variance components, the design must contain pairs of observations that share the grouping represented by one component without always sharing the grouping represented by the other. Complete crossing is sufficient for this condition but is not necessary.

\subsection{How the observed design separates the required components}
\label{app:component-connectivity}

Table~\ref{tab:component-identifiability} summarizes the principal replication patterns. These conditions concern variance components, not estimation of every individual random-effect realization.

\begin{table}[t]
\centering
\small
\caption{Observed comparisons supporting the pooled variance decomposition. ``Separation'' identifies the principal pair of components distinguished by each overlap pattern.}
\label{tab:component-identifiability}
\begin{tabular}{p{0.23\linewidth}p{0.34\linewidth}p{0.33\linewidth}}
\toprule
Separation & Required comparison & Support in the observed design \\
\midrule
\(M\) versus \(BM\)
&
The same model appears on multiple benchmarks.
&
Nine models appear on all nine benchmarks; additional models provide partial cross-benchmark replication.
\\[2pt]

\(A\) versus \(BA\)
&
The same scaffold appears on multiple benchmarks.
&
One scaffold spans eight benchmarks; an additional shared scaffold connects AssistantBench.
\\[2pt]

\(MA\) versus \(BMA\)
&
The same model--scaffold pair recurs across benchmarks.
&
Cross-benchmark models evaluated through shared scaffolds create repeated model--scaffold pairs.
\\[2pt]

\(M\) versus \(MA\)
&
Models are observed under multiple scaffolds, with overlap across models.
&
The cross-benchmark models span at least nine of the 13 scaffolds, and shared scaffolds provide common comparison conditions.
\\[2pt]

\(BM\) versus \(BMA\)
&
Within a benchmark, models are observed under more than one scaffold.
&
Each benchmark contains shared or locally replicated scaffold conditions in addition to benchmark-specific scaffolds.
\\[2pt]

\(I[B]\) versus \(IM[B]\)
&
Each item is attempted by multiple models.
&
Benchmark items are repeatedly scored across leaderboard models.
\\[2pt]

\(I[B]\) versus \(IA[B]\)
&
Each item is attempted under multiple scaffold conditions.
&
Scaffold replication within benchmarks provides item--scaffold contrasts where observed.
\\
\bottomrule
\end{tabular}
\end{table}

\subsubsection{Model and benchmark--model variation}

The distinction between persistent model capability and benchmark-specific performance is central to the leaderboard reliability coefficient. If every model appeared on only one benchmark, then \(u_m^{(M)}\) and \(u_{bm}^{(BM)}\) would be inseparable: ``model'' would be nested within benchmark. That failure does not occur here. Nine models appear on every benchmark, producing observations that share \(M\) while differing in \(BM\).

Consequently, the model kernel \(K_M\) and benchmark--model kernel \(K_{BM}\) have different support. The former links observations from the same model across benchmarks; the latter links them only within a benchmark. This overlap identifies the contrast between globally persistent model variation \(\sigma_M^2\) and benchmark-conditioned model variation \(\sigma_{BM}^2\).

\subsubsection{Scaffold and benchmark--scaffold variation}

Benchmark-specific scaffolds alone would not distinguish a scaffold main effect from a benchmark--scaffold interaction. For a scaffold used on exactly one benchmark, its \(A\) and \(BA\) columns coincide. The design avoids complete confounding because one scaffold spans eight benchmarks and a second shared scaffold connects AssistantBench. Observations using a shared scaffold have the same \(A\) level but different \(BA\) levels, providing the comparisons required to distinguish \(\sigma_A^2\) from \(\sigma_{BA}^2\).

Benchmark-specific scaffolds remain less individually informed than shared scaffolds. Their effects are estimated through the exchangeability assumptions of the hierarchical model, with uncertainty propagated into the posterior. The shared scaffolds identify the population-level separation; partial pooling regularizes levels with little direct replication.

\subsubsection{Model--scaffold and benchmark--model--scaffold variation}

Separating \(MA\) from \(BMA\) requires recurrence of a model--scaffold pair across benchmarks. The cross-benchmark models and shared scaffolds provide such recurrence: observations can share a model and scaffold while differing in benchmark. Within-benchmark scaffold variation then supplies the complementary comparisons needed to estimate how model--scaffold compatibility changes by benchmark.

This component is more weakly informed than the model and benchmark--model components because repeated model--scaffold pairs are less frequent than repeated models. Our estimands therefore integrate over its posterior uncertainty. The leave-one-benchmark-out and alternative-estimator analyses in Appendix~\ref{app:robustness} assess whether substantive conclusions depend on a small number of these bridges.

\subsubsection{Item-indexed variation}

Items are nested within benchmarks and are not expected to recur across benchmarks. Their main effects are identified because the same item is attempted by multiple models and scaffold conditions. Item--model variation is informed when a benchmark item is attempted by multiple models, while item--scaffold variation is informed by scaffold replication within benchmark.

No assumption is made that an item from one benchmark is exchangeable with an identically labeled item from another benchmark. Pooling occurs at the variance-component level: the model estimates the typical magnitude of item-conditioned interactions across the observed benchmark panel.

\subsection{What is not separately identifiable}
\label{app:connectivity-limitations}

The design does not contain independent repeated executions of every
\((b,i,m,a)\) cell. A four-way benchmark--item--model--scaffold interaction is therefore observationally confounded with cell-level execution variation and the Bernoulli residual for most of such interactions. We combine these sources into the terminal component
\[
\sigma_{BIMA,e}^2.
\]
This is not a limitation for the reported relative reliability coefficients: all unresolved cell-specific variation belongs in the error term because it can change model ordering across repeated evaluation conditions. Separating execution stochasticity, grading instability, and four-way interaction would require repeated runs under the same recorded cell and, ideally, explicit grading replicates.

More generally, the present design does not support:

\begin{itemize}[leftmargin=0.5cm,nosep]
    \item an unregularized estimate for every unobserved model--scaffold cell;
    \item precise scaffold-specific effects for scaffolds used on only one benchmark;
    \item causal claims about replacing one scaffold with another; or
    \item generalization to scaffold or benchmark populations unrelated to the connected evaluation panel.
\end{itemize}

These quantities are unnecessary for our research questions. We require population-level variance components and scaffold-marginalized model contrasts, not predictions for every missing factorial cell.

\subsection{Identifiability of the reliability estimands}
\label{app:estimand-identifiability}

The principal leaderboard coefficient is Equation \ref{eq:dstudy-full}:
\begin{equation*}
E\rho_M^2(n_b,n_i,n_a)
=
\frac{\sigma_M^2}
{\sigma_M^2
+\frac{\sigma_{BM}^2}{n_b}
+\frac{\sigma_{MA}^2}{n_a}
+\frac{\sigma_{BMA}^2}{n_b n_a}
+\frac{\sigma_{IM[B]}^2}{n_b n_i}
+\frac{\sigma_{BIMA,e}^2}{n_b n_i n_a}}.
% \label{eq:connectivity-reliability}
\end{equation*}
This coefficient depends on a small set of variance components and their sums, not on every random effect in
Equation~\ref{eq:pooled}. The design requirements are correspondingly weaker than those needed to reconstruct the complete model--benchmark--scaffold response tensor.

\begin{proposition}[sufficiency for the leaderboard estimand]\label{prop:suff_estim}
    Suppose that:

\begin{enumerate}[label=(\roman*),leftmargin=0.7cm,nosep]
    \item the benchmark--model incidence graph is connected and contains models observed on multiple benchmarks;
    \item the benchmark--scaffold incidence graph is connected and contains scaffolds observed on multiple benchmarks;
    \item at least some model--scaffold pairs recur across benchmarks; and
    % \item benchmark items are attempted by multiple models;
    \item benchmark items are attempted by multiple models and scaffolds.
\end{enumerate}

Then the covariance kernels corresponding to
\(M\), \(BM\), \(MA\), \(BMA\), and \(IM[B]\) have distinct observed replication patterns. The variance combinations entering
Equation~\ref{eq:dstudy-full} are therefore structurally distinguishable, up to the explicitly combined terminal component
\(\sigma_{BIMA,e}^2\).
\end{proposition}
% \paragraph{Proposition 2 (sufficiency for the leaderboard estimand).}

% \paragraph{Justification.}
 \textit{Justification}  
   Conditions (i)--(iv) provide observation pairs that share, respectively:
\(M\) but not \(BM\);
\(A\) but not \(BA\);
\(MA\) but not \(BMA\); and
\(IM[B]\) and not only \(I[B]\).
These yield distinct covariance kernels under
Proposition \ref{prop:vc_criterion}. The cell-specific remainder is intentionally aggregated because no additional replication distinguishes its constituents. \(\square\)

The observed leaderboard satisfies these conditions. In particular, the nine cross-benchmark models identify persistent versus benchmark-conditioned model variation, while shared scaffolds and repeated model--scaffold pairs identify the scaffold-related components that determine the task-only reliability ceiling.

\begin{corollary}[identifiability of D-study ceilings]\label{cor:ceiling}
    Under Proposition~\ref{prop:suff_estim}, the task-only limit
\[
\lim_{n_i\rightarrow\infty}
E\rho_M^2(n_b,n_i,n_a)
=
\frac{\sigma_M^2}
{\sigma_M^2
+\sigma_{BM}^2/n_b
+\sigma_{MA}^2/n_a
+\sigma_{BMA}^2/(n_bn_a)}
\]
is identified without separately decomposing terminal item-level residual variation.
\end{corollary}

% \paragraph{Corollary 1 (identifiability of D-study ceilings).}

Corollary \ref{cor:ceiling} is important for the paper's central design conclusion. The claim that adding tasks cannot eliminate benchmark- and scaffold-conditioned error does not depend on a precise decomposition of every item-level noise source. It depends on the persistent components whose identification is supported by cross-benchmark models and scaffolds.

\subsection{Bayesian regularization and evidential scope}
\label{app:bayesian-identifiability}

All variance components are estimated with proper, weakly informative priors. The posterior is therefore proper even when a component is weakly informed near the boundary. We use priors to stabilize finite-sample estimation, not to substitute for disconnected comparisons. Three features constrain interpretation:

\begin{enumerate}[leftmargin=0.6cm,nosep]
    \item \textbf{Data-supported contrasts.}
    Model rankings are inferred through the connected observation graph. We do not report comparisons between disconnected components because none exist in the observed design.
    
    \item \textbf{Posterior rather than plug-in reliability.}
    Equation~\ref{eq:dstudy-full} is evaluated within each posterior draw. Weak separation among related components therefore appears as uncertainty in reliability and its D-study projection.
    
    \item \textbf{Sensitivity to bridges.}
    Leave-one-benchmark-out refits test whether conclusions rely on a particular benchmark or scaffold connection. Prior and estimator comparisons test whether weak components are driving the reported variance ordering. These analyses are reported in Appendix~\ref{app:robustness}.
\end{enumerate}

The distinction between identifiability and precision is especially important for agent leaderboards. Sparse designs can be connected enough to estimate whether scaffold variation is non-negligible while still being too sparse to estimate its exact magnitude narrowly. Large credible intervals in this setting are not model failure; they reveal that the leaderboard contains too few common evaluation conditions to support precise claims.

\subsection{Connectivity Summary}
\label{app:connectivity-summary}

The full design is incomplete but connected. Nine models evaluated across every benchmark anchor the model scale; shared scaffolds connect the benchmark panel; and recurring model--scaffold pairs distinguish general scaffold compatibility from benchmark-specific compatibility. These overlaps are sufficient for the variance combinations underlying model-ranking reliability, signal-to-noise ratios, scaffold-versus-model comparisons, and D-study ceilings.

The design does not identify every possible cell effect, nor is that required. Unreplicated terminal variation is assigned to relative error, and uncertainty from sparsely observed scaffold interactions is propagated through the Bayesian posterior. The resulting claims are therefore appropriately scoped: they characterize the reliability supported by this connected leaderboard ecosystem, rather than asserting a complete factorial decomposition of all possible models, scaffolds, tasks, and benchmarks.

\section{Additional Tables and Figures}\label{app:add_plots}

The coefficients of Table~\ref{tab:reliability-coefficients} isolate \emph{where} a benchmark's signal lives and \emph{which} facet dissipates it. Because each coefficient shares a common numerator--denominator structure (signal variance over signal plus generalizing error), differences across coefficients within a single benchmark are directly interpretable: they reveal how much of the apparent capability signal survives when we generalize over tasks, over scaffolds, or over both. Table~\ref{tab:coef_results} reports posterior means from the per-benchmark fit of Eq.~\ref{eq:perbenchmark}.

\begin{table}[!h]
\centering
\caption{\textbf{Reliability depends on the object being ranked.}
    Reliability estimates for each benchmark under model and model--scaffold systems as measurement systems, including the task limit for models as the object of measurement.}
    \label{tab:bench_stats}
\resizebox{\ifdim\width>\linewidth\linewidth\else\width\fi}{!}{
\begin{tabular}{lrrrrrrrrr}
\toprule
Estimand 
& \textsc{Assistant}
 & \textsc{CORE-Hard}
 & \textsc{GAIA}
 & \textsc{Mind2Web}
 & \textsc{SciCode}
 & \textsc{ScienceAgent}
 & \textsc{SWE-mini}
 & \(\tau\)\textsc{-Airline}
 & \textsc{USACO} \\
\midrule
$\lim_{n_i\rightarrow\infty}E\rho_{M}^2$ (\ref{eq:task-ceiling}) & 0.210 & 0.892 & 0.325 & 0.153 & 0.840 & 0.683 & 0.552 & 0.669 & 0.377\\
$E\rho_{M}^2$ (\ref{eq:perbench-model-reliability}) & 0.182 & 0.841 & 0.318 & 0.148 & 0.811 & 0.628 & 0.543 & 0.572 & 0.375\\
$E\rho_{MA}^2$ (\ref{eq:perbench-system-reliability}) & 0.935 & 0.972 & 0.990 & 0.974 & 0.975 & 0.987 & 0.993 & 0.944 & 0.994\\
\bottomrule
\end{tabular}}
\end{table}

\begin{figure}[t]
    \centering
    \includegraphics[width=\linewidth]{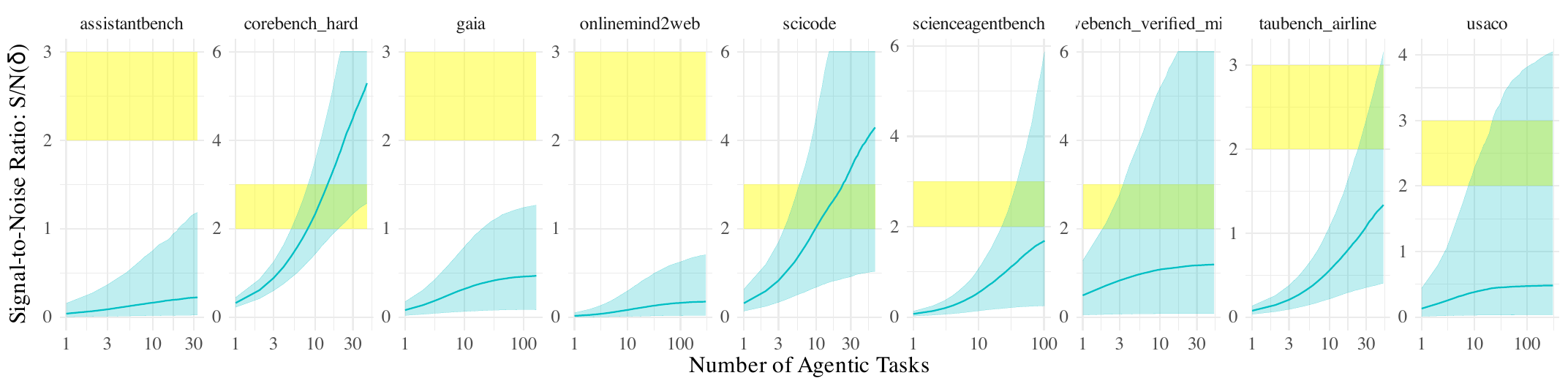}
    \caption{\textbf{Benchmarks differ sharply in ranking signal.}
    Posterior model-ranking signal-to-noise ratio as tasks are added. Shaded bands give external detection-limit reference values. Points and intervals are posterior medians and \(68\%\) standard error HDIs.}
    \label{fig:mod_snr}
\end{figure}

\begin{table}[t]
\centering
\small
\caption{Posterior-mean reliability and generalizability coefficients per benchmark. $E\rho^2_{M(i)}$ and $E\rho^2_{A(i)}$ are task reliabilities ($\sigma^2_o/(\sigma^2_o+\sigma^2_{io})$) for model and scaffold rankings% (Eqs.~\ref{eq:trm}--\ref{eq:tra})
; $\rho^{(b)}_{AA'}$ is inter-scaffold reliability (Eq.~\ref{eq:interscaffoldrho}); $E\rho^2_{M(b)}$, $E\rho^2_{A(b)}$, and $E\rho^2_{MA(b)}$ are G-coefficients by object of measurement (Eqs.~\ref{eq:perbench-model-reliability}--\ref{eq:perbench-system-reliability}); the last column is the task-saturated model ceiling $\lim_{n_i\to\infty}E\rho^2_{M(b)}$ (Eq.~\ref{eq:task-ceiling}).}
\label{tab:coef_results}
\begin{tabular}{lccccccc}
\toprule
Benchmark & $E\rho^2_{M(i)}$ & $E\rho^2_{A(i)}$ & $\rho^{(b)}_{AA'}$ & $E\rho^2_{M(b)}$ & $E\rho^2_{A(b)}$ & $E\rho^2_{MA(b)}$ & $E\rho^2_{M(b)}(\infty)$ \\
\midrule
scicode                & 0.602 & 0.534 & 0.746 & 0.736 & 0.516 & 0.971 & 0.765 \\
gaia                   & 0.263 & 0.705 & 0.333 & 0.328 & 0.579 & \textbf{0.990} & 0.334 \\
taubench\_airline      & 0.549 & 0.539 & 0.545 & 0.533 & 0.759 & 0.938 & 0.619 \\
swebench\_verified\_mini & 0.599 & 0.857 & 0.504 & 0.499 & 0.639 & \textbf{0.992} & 0.506 \\
usaco                  & 0.238 & 0.253 & 0.392 & 0.429 & 0.478 & \textbf{0.993} & 0.432 \\
assistantbench         & 0.362 & 0.445 & 0.287 & 0.260 & 0.521 & 0.911 & 0.305 \\
corebench\_hard        & \textbf{0.636} & \textbf{0.941} & \textbf{0.828} & \textbf{0.817} & 0.805 & 0.972 & \textbf{0.866} \\
onlinemind2web         & 0.296 & 0.228 & 0.205 & 0.203 & 0.430 & 0.972 & 0.210 \\
scienceagentbench      & 0.558 & 0.753 & 0.567 & 0.561 & \textbf{0.881} & 0.982 & 0.610 \\
\bottomrule
\end{tabular}
\end{table}

\begin{figure}
    \centering
    \includegraphics[width=1\linewidth]{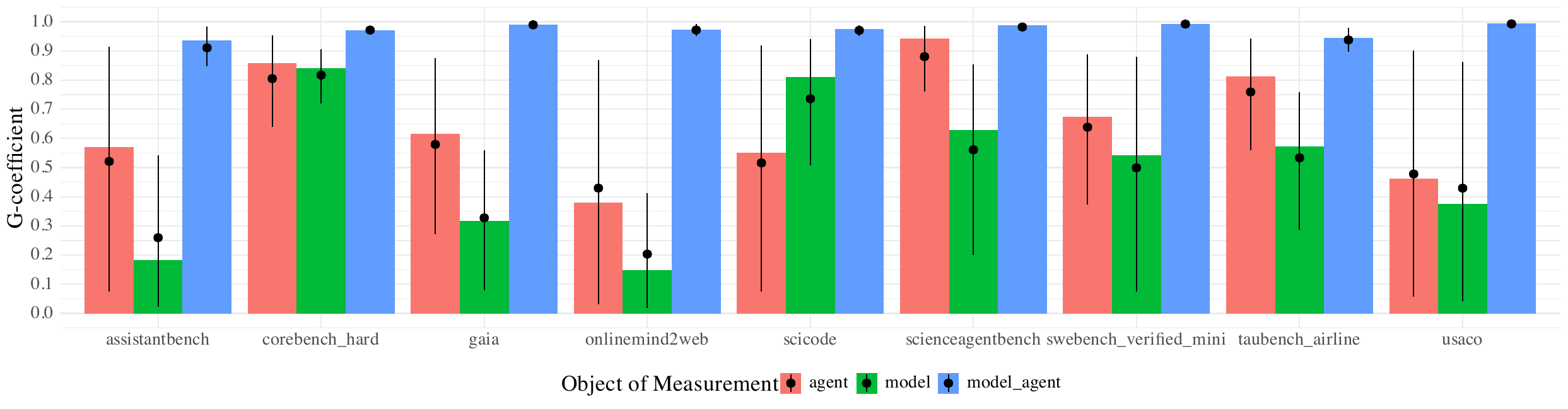}
    \caption{Reliability by Object of Measurement across all tasks in a benchmark. Estimated using variance components of Equation \ref{eq:perbenchmark}, and posteriors of Equations \ref{eq:perbench-model-reliability} and \ref{eq:perbench-system-reliability}.     Intervals are pseudo-standard error at 0.68 HDI, bar estimates are medians and points are means of posterior draw distributions. }
    \label{fig:gcoef_obj}
\end{figure}

\begin{table}[!h]
\centering
\caption{\textbf{Internal Score Transportability.} Kendall's tau correlation matrix of benchmarks and summary statistics against each benchmark's mean accuracy, the unweighted HAL mean score, and reliability-adjusted latent ($\widehat{\theta}$) model effect (median posterior)}
\label{tab:rankcors}
\resizebox{\ifdim\width>\linewidth\linewidth\else\width\fi}{!}{
\begin{tabular}{lrrrrrrrrrrr}
\toprule
 & \textsc{Assistant}
 & \textsc{CORE-Hard}
 & \textsc{GAIA}
 & \textsc{Mind2Web}
 & \textsc{SciCode}
 & \textsc{ScienceAgent}
 & \textsc{SWE-mini}
 & \(\tau\)\textsc{-Airline}
 & \textsc{USACO}
 & \textsc{Mean Score}
 & \(\widehat{\theta}\) \\
\midrule
assistantbench & 1.0 & 0.300 & \textbf{0.485} & -0.061 & 0.103 & 0.215 & -0.404 & 0.180 & 0.215 & 0.205 & 0.261\\
corebench\_hard & 0.300 & 1.0 & 0.487 & 0.400 & 0.492 & 0.458 & 0.148 & 0.581 & 0.167 & 0.687 & \textbf{0.835}\\
gaia & 0.485 & 0.487 & 1.0 & 0.028 & 0.260 & 0.479 & 0.215 & 0.363 & 0.182 & 0.453 & \textbf{0.606}\\
onlinemind2web & -0.061 & 0.400 & 0.028 & 1.0 & 0.171 & 0.056 & 0.424 & 0.315 & -0.141 & \textbf{0.581} & 0.348\\
scicode & 0.103 & \textbf{0.492} & 0.260 & 0.171 & 1.0 & 0.184 & 0.117 & 0.477 & 0.225 & 0.202 & 0.472\\
scienceagentbench & 0.215 & 0.458 & 0.479 & 0.056 & 0.184 & 1.0 & 0.129 & 0.587 & -0.085 & 0.392 & \textbf{0.499}\\
swebench\_verified\_mini & -0.404 & 0.148 & 0.215 & 0.424 & 0.117 & 0.129 & 1.0 & 0.023 & 0.032 & 0.509 & \textbf{0.554}\\
taubench\_airline & 0.180 & 0.581 & 0.363 & 0.315 & 0.477 & 0.587 & 0.023 & 1.0 & 0.073 & 0.404 & \textbf{0.588}\\
usaco & 0.215 & 0.167 & 0.182 & -0.141 & 0.225 & -0.085 & 0.032 & 0.073 & 1.0 & \textbf{0.390} & 0.338\\
% \addlinespace
% overall & 0.205 & 0.687 & 0.453 & 0.581 & 0.202 & 0.392 & 0.509 & 0.404 & 0.390 & NA & 0.501\\
% theta\_m & 0.261 & 0.835 & 0.606 & 0.348 & 0.472 & 0.499 & 0.554 & 0.588 & 0.338 & 0.501 & NA\\
\bottomrule
\end{tabular}}
\end{table}

\subsection{Repeated Task Subsampling}\label{app:subsampling}
We confirm the cost reduction findings by repeatedly sampling \(k\) tasks per benchmark without replacement, using the same sampled tasks for all models to ensure a paired comparison. Within each subsample, scores are first averaged at the benchmark--model level and then averaged across benchmarks, giving each benchmark equal weight. 
The resulting model ranking is compared with the ranking obtained from the complete dataset using Spearman's rank correlation coefficient (\(\rho\)) and Kendall's rank correlation coefficient (\(\tau\)). Repeating this procedure 500 times across many random subsamples yields a distribution of rank correlations for each \(k\), quantifying how stable the model rankings are with respect to the number and selection of evaluated tasks. Task subsampling corroborates the diminishing returns. Across \(500\) paired subsamples, \(15\) tasks per benchmark retain mean Spearman agreement of \(0.932\) with the complete-data ranking at an estimated \(82\%\) cost reduction; \(30\) tasks increase agreement to \(0.976\)
Agreement with the full ranking establishes information retention, but makes not claim to ranking validity. 
\begin{table}[!h]
\centering
\caption{\textbf{Task subsampling shows rank information preservation.} Bootstrapped correlations of subsampled tasks and leaderboard Ranks}
\label{tab:k_subsamples}
\resizebox{\ifdim\width>\linewidth\linewidth\else\width\fi}{!}{
\begin{tabular}{rrrrrrrrrrrr}
\toprule
k & mean \rho & median \rho & \rho min & \rho max & mean \tau & median \tau & \tau min & \tau max & top@1 agreement & mean abs rank change &cost savings\\
\midrule
5 & 0.790 & 0.801 & 0.584 & 0.917 & 0.634 & 0.640 & 0.458 & 0.766 & 0.640 & 7.221 &0.94\\
10 & 0.891 & 0.896 & 0.804 & 0.952 & 0.742 & 0.745 & 0.646 & 0.829 & 0.756 & 5.110 &0.88\\
% 14 & 0.923 & 0.928 & 0.847 & 0.966 & 0.786 & 0.789 & 0.695 & 0.852 & 0.864 & 4.285\\
15 & 0.932 & 0.936 & 0.870 & 0.966 & 0.799 & 0.802 & 0.720 & 0.858 & 0.834 & 4.052 &0.82\\
% 16 & 0.936 & 0.940 & 0.879 & 0.972 & 0.809 & 0.812 & 0.729 & 0.873 & 0.896 & 3.903\\
% \addlinespace
% 17 & 0.941 & 0.945 & 0.889 & 0.970 & 0.814 & 0.818 & 0.736 & 0.871 & 0.884 & 3.784\\
% 18 & 0.946 & 0.950 & 0.904 & 0.973 & 0.822 & 0.825 & 0.759 & 0.872 & 0.890 & 3.580\\
% 19 & 0.948 & 0.952 & 0.903 & 0.975 & 0.827 & 0.828 & 0.762 & 0.881 & 0.926 & 3.518\\
20 & 0.954 & 0.957 & 0.915 & 0.978 & 0.837 & 0.839 & 0.778 & 0.887 & 0.912 & 3.323 &0.76\\
25 & 0.967 & 0.969 & 0.940 & 0.982 & 0.863 & 0.865 & 0.811 & 0.901 & 0.960 & 2.822 &0.69\\
% \addlinespace
30 & 0.976 & 0.977 & 0.957 & 0.988 & 0.886 & 0.889 & 0.842 & 0.923 & 0.990 & 2.385 &0.63\\
% 33 & 0.980 & 0.981 & 0.964 & 0.990 & 0.896 & 0.898 & 0.854 & 0.932 & 0.988 & 2.192\\
\bottomrule
\end{tabular}}
% \caption*{}
\end{table}

\begin{figure}
    \centering
    \includegraphics[width=\linewidth]{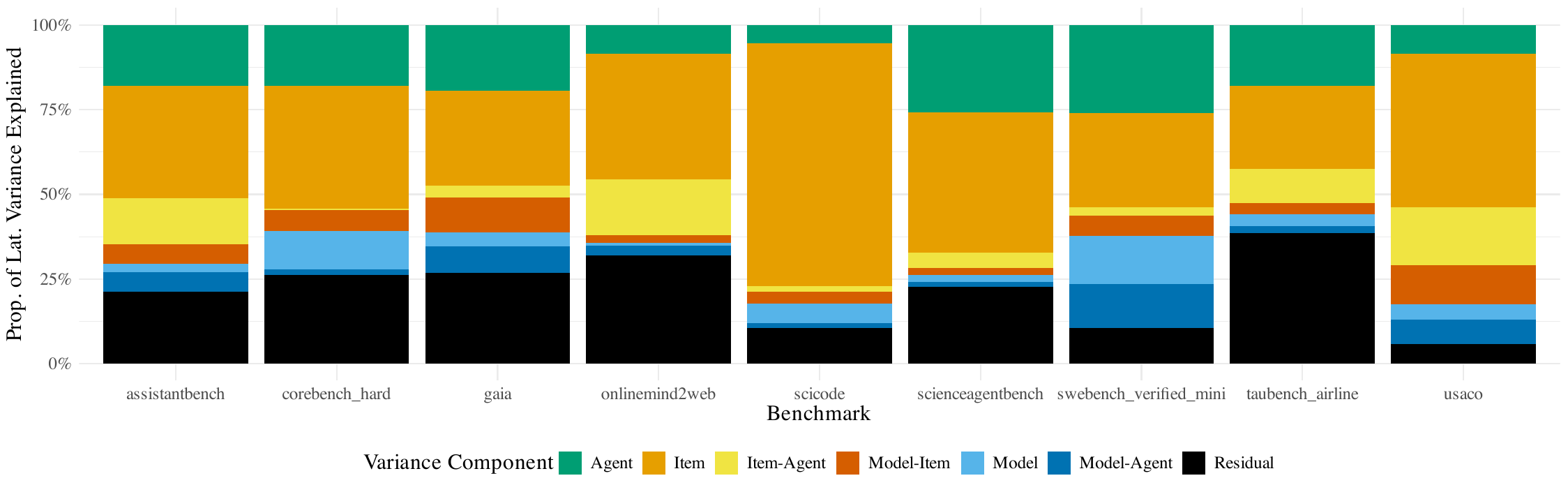}
    \caption{Proportion of Variance Explained in the posterior distributions from Eq. \ref{eq:perbenchmark}. ``Item'' from benchmarking literature represents ``task'' commonly found in agentic evaluation literature. ``Agent'' represents ``scaffold''.}
    \label{fig:prop_var_explained_per_bench}
\end{figure}

\begin{figure}[!ht]
    \centering
    \includegraphics[width=1\linewidth]{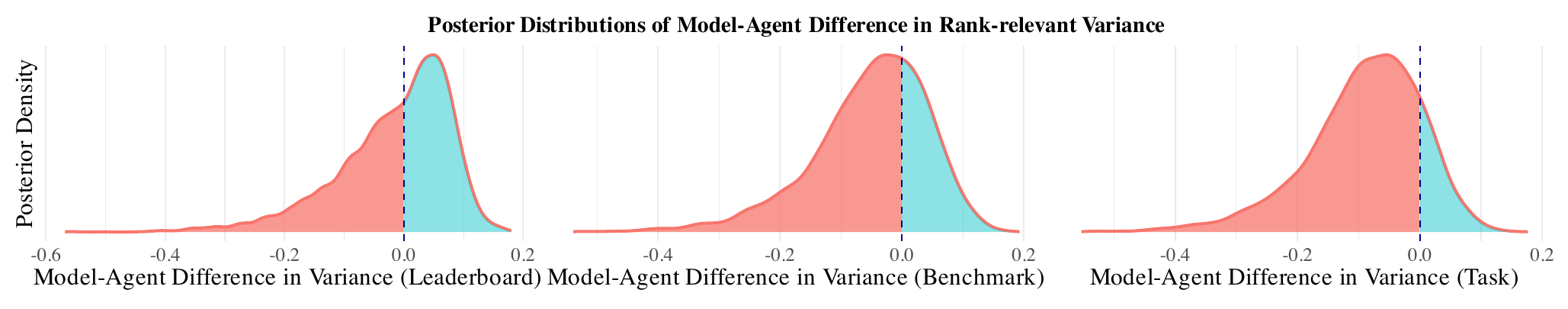}
    \caption{Posterior density of Model-Agent difference of rank-relevant variance across the leaderboard (left), across a sampled benchmark (middle), and across a sampled task (right)%, from Eq. \ref{eq:pooled}
    . Positive numbers suggest LLM variance contribution is greater than that of agent scaffold. 
    }
    \label{fig:vardiff_levels}
\end{figure}

\begin{figure}
    \centering
    \includegraphics[width=1\linewidth]{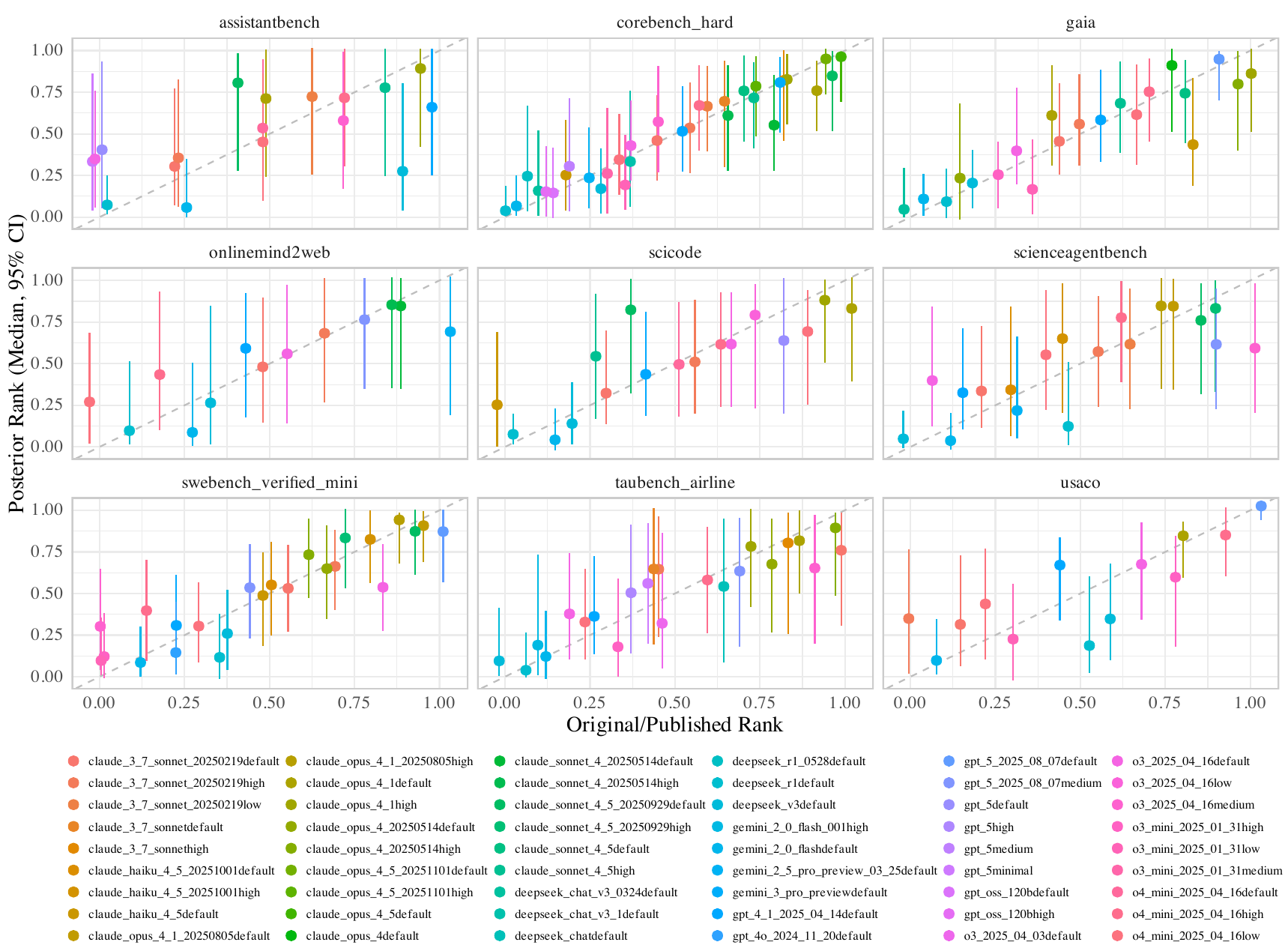}
    \caption{Published score versus posterior model percentile ranks using model-wise mean published score (x-axis) and draw-wise posterior median rankings with 95\% CIs. Vertical intervals are \(95\%\) posterior rank intervals; jitter separates horizontal ties. }
    \label{fig:old_new_rank_apx}
\end{figure}

\begin{figure}
    \centering
    \includegraphics[width=1\linewidth]{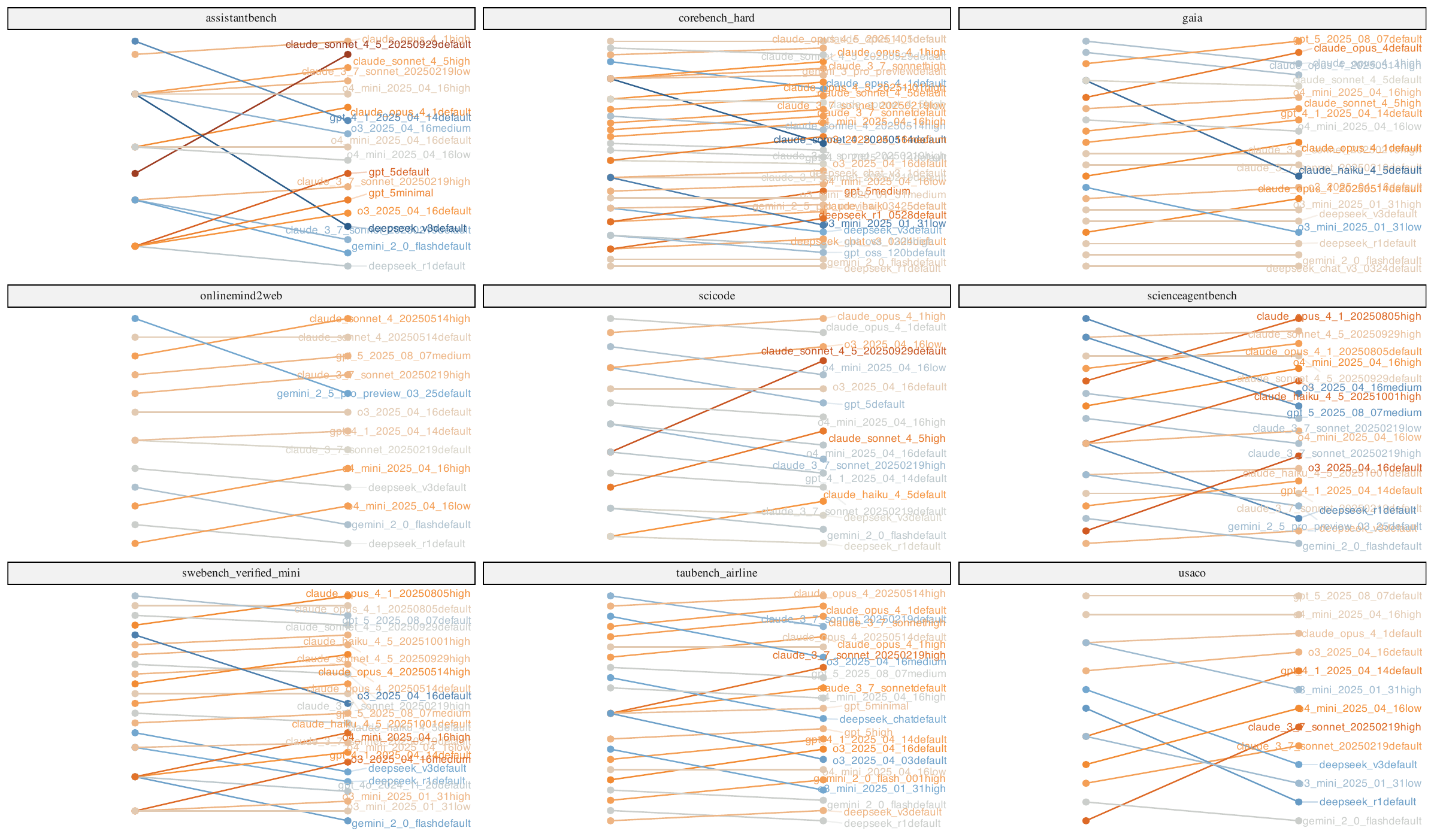}
    \caption{Rank order changes by benchmark based on comparison with median posterior rank from full variance decomposition model of Eq. \ref{eq:pooled}. %$\tau$-bench Airline and SWEBench (Verified, Mini) can be found in the main body Figure \ref{fig:rank_instability_swebench_tau}
    }
    \label{fig:rank_instability_appendix}
\end{figure}

\begin{figure}
    \centering
    \includegraphics[width=1\linewidth]{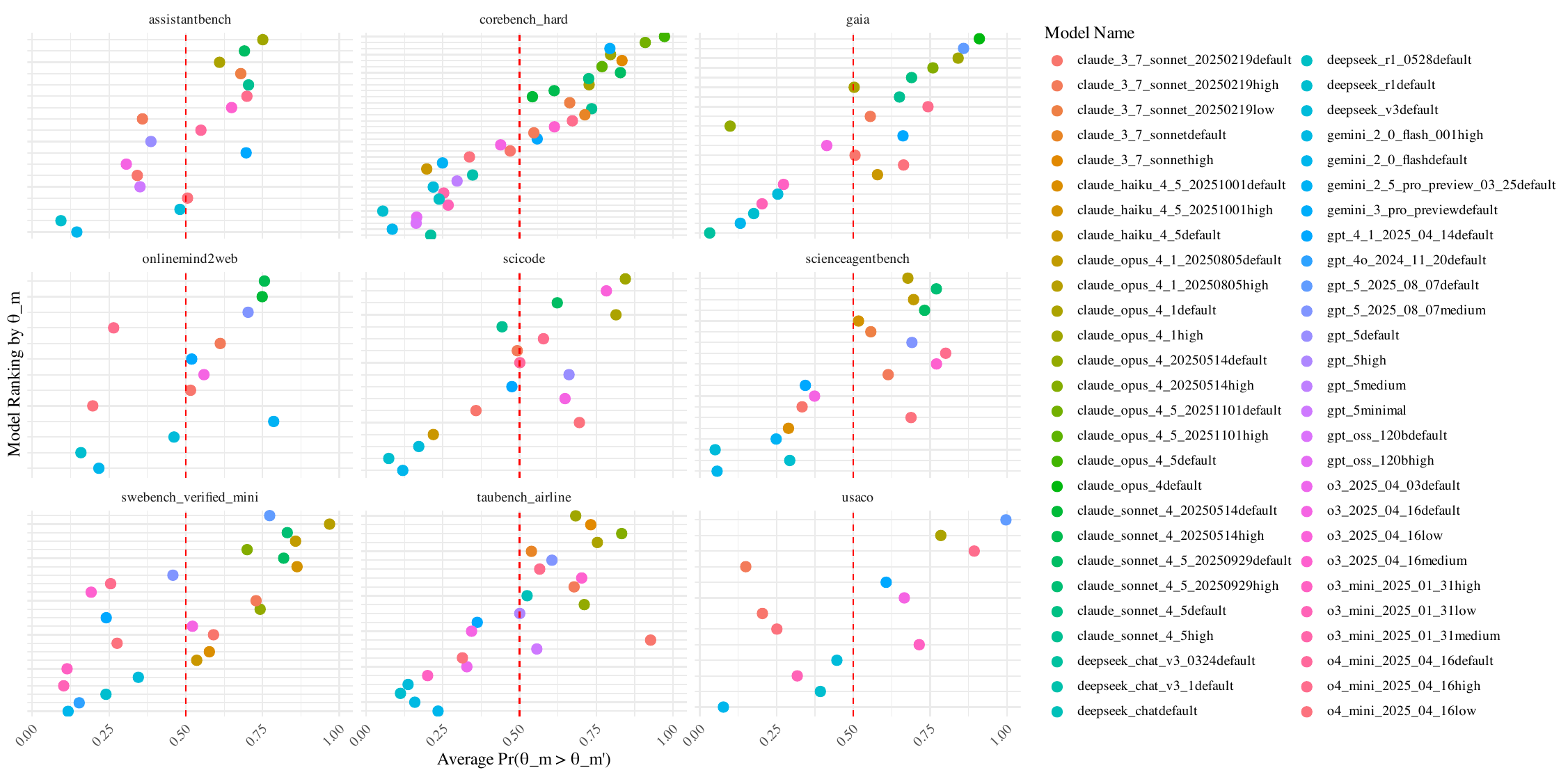}
    \caption{Cumulative posterior dominance rankings across individuals. The point estimate for each individual \(m\) represents their global dominance score, defined as the expected probability that their latent parameter \(\theta_{m} = u_{m}^{(M)}  + u_{mb}^(MB)\) is strictly greater than that of a randomly selected peer \(m^{\prime }\) from the population, given the observed data \(y\): \(\text{Global Dominance}(m)=\frac{1}{M-1}\sum _{m^{\prime }\ne m}\Pr (\theta _{m}>\theta _{m^{\prime }}\mid y)\)where \(M\) is the total number of individuals, and the pairwise probabilities are empirically derived via Bayesian draws as:\(\Pr (\theta _{m}>\theta _{m^{\prime }}\mid y)\approx \frac{1}{D}\sum _{d=1}^{D}\mathbb{I}\left(\theta _{m}^{(d)}>\theta _{m^{\prime }}^{(d)}\right)\)with \(\mathbb{I}(\cdot)\) denoting the indicator function and \(D\) representing the total number of posterior draws. Individuals are arranged along the vertical axis in ascending order of their global dominance scores. The vertical dashed line at \(\Pr = 0.5\) denotes the theoretical baseline of a perfectly average individual who exhibits structural parity relative to the rest of the cohort. Deviations toward \(1.0\) signify deterministic stochastic dominance over the population, whereas values approaching \(0.0\) indicate systematic underperformance relative to the group.}
    \label{fig:pr_theta_dominance}
\end{figure}

\begin{figure}
    \centering
    \includegraphics[width=1\linewidth]{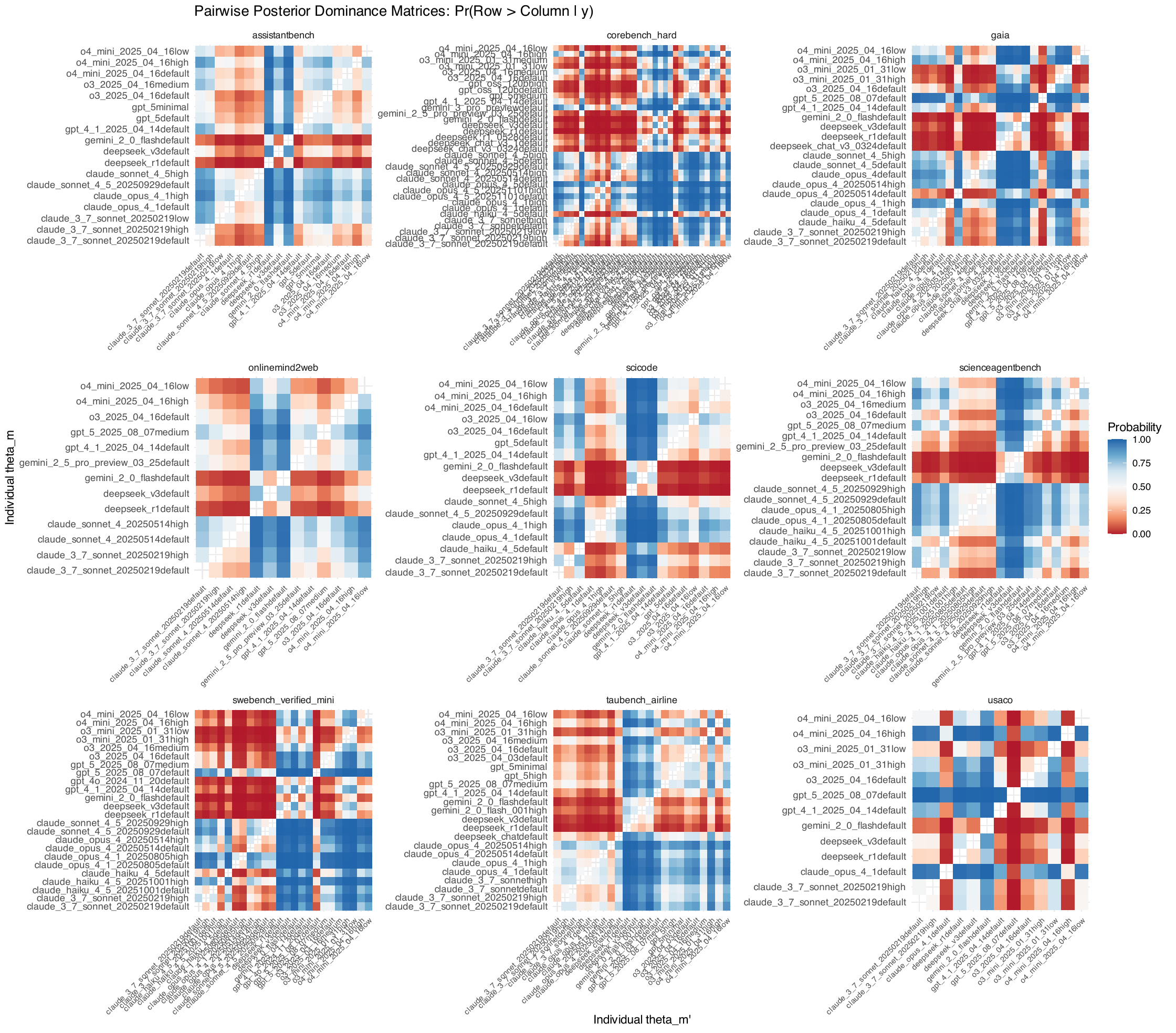}
    \caption{Pairwise Posterior Dominance, where color is Pr(Row > Column$\mid y$) across posterior draws. The point estimate for each individual \(m\) represents \(\Pr (\theta _{m}>\theta _{m^{\prime }}\mid y)\), defined by the expected probability of latent parameter \(\theta_{m} = u_{m}^{(M)}  + u_{mb}^(MB)\), for each benchmark $b$.}
    \label{fig:pairposteriordom}
\end{figure}

% \begin{figure}
%     \centering
%     \includegraphics[width=1\linewidth]{figs/invlogit_oldnew_score_comp_swetau.pdf}
%     \caption{Published scores and estimated scores with standard errors estimated from the Best Linear Unbiased Predictors from the posterior of Eq. \ref{eq:pooled}. Estimates for all benchmarks are in Figure \ref{fig:oldscore_newscore}. Variance components aggregated after inverse logit transform, thus reliabilities represent the observation space.}
%     \label{fig:swe_tau_SEs}
% \end{figure}

\newpage

\section{Method and Scale Comparison}\label{app:method_comps}

\subsection{Latent Space Estimations}\label{app:latent_estimations}
Variance components in our generalized linear mixed model (GLMM) decompositions live on the \emph{latent} (logit) scale, so $\rho^2$ should be interpreted as reliability of the linear predictor $\eta$ (i.e., of differences in \emph{log-odds}) rather than as reliability of raw percent-correct. This is standard in binary-response measurement: the latent scale is the scale on which additive random effects and G-theory D-study algebra apply, and it is the scale on which “true score + error” is well-defined without probability-dependent heteroskedasticity. Importantly, a given amount of latent variability implies different variability in percent-correct depending on where a model sits on the sigmoid: near $p\approx 0.5$, small changes in $\eta$ translate to large changes in $p$, while near $p\approx 0$ or $1$ they compress. For that reason, a latent-scale reliability such as $\rho^2\approx 0.45$ \emph{does} correspond to meaningful rank instability on the probability scale. 
% N makes clear that our conclusions are not an artifact of a latent metric: the 
The latent-scale $\rho^2$ is a mathematically coherent reliability target for binary data, and the associated posterior predictive mapping quantifies how it manifests as practically relevant leaderboard instability in percent-correct.

\begin{figure}
    \centering
    \includegraphics[width=1\linewidth]{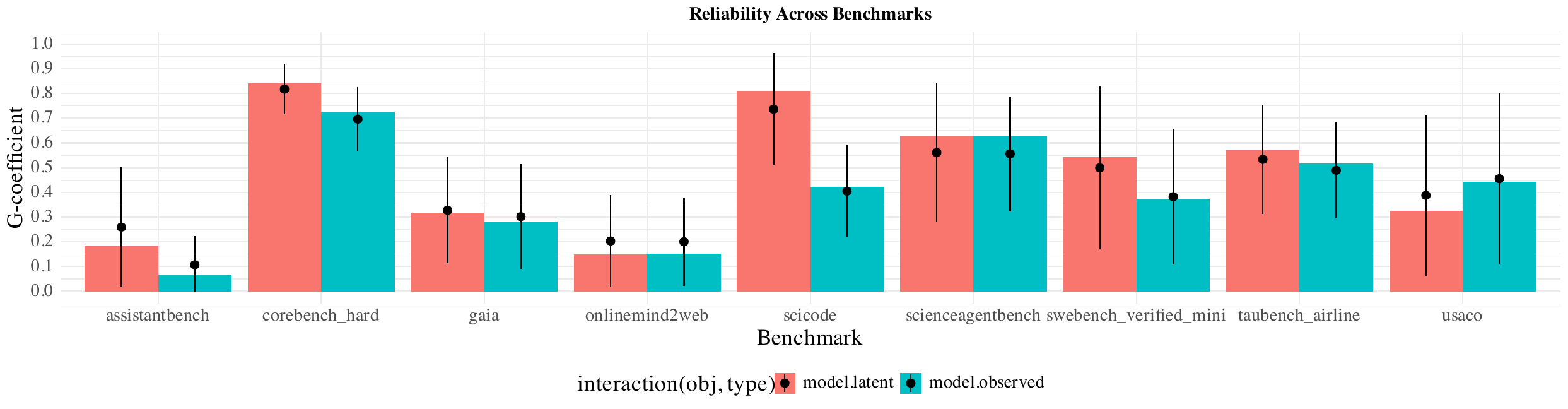}
    \caption{Rank order reliability, G-coefficient, per benchmark, for both latent and observed estimations with LLM as object of measurement. Bars are medians, points are means, and intervals are pseudo-standard error 0.68 HDI across the posterior using Eq. \ref{eq:perbench-model-reliability}}
    \label{fig:ep2lat_vs_obs}
\end{figure}

In addition to the interpretation benefits and alignment to measurement theoretic latent modeling, the logistic mixed effect approach also handles unbalanced data and values near zero, both common in leaderboard testing, better than the observation-level linear approach. This can be seen in Figure \ref{fig:ep2lat_vs_obs}, which shows the latent and observed modeling estimated reliabilities. Rank order reliabilities are higher for the latent estimations proportionate to the quantity of low scoring models (mean score per model $\bar{y}_{mb} \lessapprox0.1$). This is particularly true of SciCode where 100\% of models have an accuracy score less than 0.1.

\subsubsection{Variance shares and reliability on the latent scale}
For GLMM/Bayesian GLMM, we report variance shares as proportions of total latent variance:
\[
\pi_k \;=\; \frac{\sigma_k^2}{\sum_j \sigma_j^2},
\]
where the sum runs over included random effects and (optionally) a latent residual term. For Bayesian fits, $\pi_k$ is computed per posterior draw, yielding credible intervals. These can be found in Tables \ref{tab:propvardecomp3} and \ref{tab:propvardecomp4}.

\paragraph{D-study scaling.}
When an interaction term involves a sampled facet, its contribution to the variance of an averaged score shrinks with the number of sampled levels. For example, with tasks averaged ($n_i$ tasks), an $I\times M$ component contributes $\sigma^2_{IM}/n_i$.

\subsection{Distance components (DISCO) quasi-reliability}\label{sec:app-disco}
The DISCO decomposition \citep{rizzo_disco_2010} generalizes classical ANOVA to arbitrary metric spaces. For $K$ groups with $n_k$ observations each, define the within-group dispersion as
\begin{equation}
    \mathcal{S}_W = \sum_{k=1}^{K} \frac{1}{n_k} \sum_{i<j} \|X_{ki} - X_{kj}\|,
\end{equation}
and the total dispersion as
\begin{equation}
    \mathcal{S}_T = \frac{1}{N} \sum_{i<j} \|X_i - X_j\|,
\end{equation}
where $N = \sum_k n_k$.  The between-group component is $\mathcal{S}_B = \mathcal{S}_T - \mathcal{S}_W$, and the proportion attributable to the grouping factor is $\mathcal{S}_B / \mathcal{S}_T$.

For multi-facet designs, we compute DISCO sequentially for each facet, using the dispersion ratio as the analog of $\eta^2$.  Because DISCO does not produce a residual term, the facet proportions do not generally sum to one when computed independently; we normalize to aid comparison with the parametric estimates.

\begin{remark}\label{rem:disco-interactions}
DISCO attributes substantially more dispersion to interaction terms than parametric methods.  This occurs because pairwise distances capture nonlinear dependencies--such as a model performing anomalously well on a specific cluster of tasks--that additive random-effects models cannot represent. The DISCO interaction estimates should therefore be interpreted as an upper bound on the importance of non-additive effects, complementing the more conservative parametric decomposition.
\end{remark}

DISCO decomposes dispersion using pairwise distances rather than squared deviations, improving robustness under non-normality and heavy-tailed random effects. For an object grouping $g$ (e.g., model or model–scaffold pair), DISCO yields between-group and within-group dispersion terms $(T_g, T_{\text{within}})$, from which we define
\[
E\rho^2 \;=\; \frac{T_g}{T_g + T_{\text{within}}}.
\]
We use DISCO primarily as a methodological sensitivity analysis to validate that conclusions (dominant facets; low model-ranking reliability) are not artifacts of Gaussian random-effect assumptions.

\subsection{Why estimation method changes conclusions—and what to do about it}\label{app:estim_choices}

The five estimation methods are not interchangeable, and their disagreements are informative.

A notable outcome is that variance shares differ across linear mixed model (LMM), Bayesian LMM, generalized linera mixed model (GLMM), Bayesian GLMM, and DISCO, especially under sparse observations:
\begin{itemize}[nosep]
\item \textbf{LMM} tends to allocate a large portion to residual $\sigma^2$, which can understate structured interactions when the link is misspecified for Bernoulli data.
\item \textbf{GLMM} can collapse some components toward zero (boundary estimates) in unbalanced designs, yielding deceptively “clean” decompositions.
\item \textbf{Bayesian LMM} allows for estimates of reliability and distributional uncertainty on observed scale, but less equipped for floor or ceiling effects or highly imbalanced data.
\item \textbf{Bayesian GLMM} exposes skewness and large uncertainty, which is appropriate when the design under-identifies components.
\item \textbf{DISCO} can attribute more to interaction-like dispersion without parametric assumptions, often aligning with the intuition that agentic pipelines create nonlinear, heteroskedastic effects.
\end{itemize}

\textbf{Practical guidance.}
\begin{enumerate}[nosep]
\item Use \textbf{Bayesian} or \textbf{nonparametric} methods to communicate uncertainty when data are sparse; prefer point-estimate GLMM/LMM only when coverage is dense.
\item Triangulate: when all four methods agree on the \emph{ranking} of dominant facets (e.g., task vs interactions vs residual), recommendations are robust; when they disagree, the correct conclusion is that the benchmark is under-instrumented for that inference.
\end{enumerate}

\paragraph{Linear vs.\ generalized models.} The LME estimates generally attribute the largest share of variance to the residual (with a borderline exception of USACO), with correspondingly compressed named components. The GLME and Bayesian estimates, by contrast, allocate substantially more variance to specific tasks and models. This divergence is expected: the Gaussian identity-link model treats binary $\{0,1\}$ responses as continuous, and its residual absorbs the Bernoulli variance floor ($p(1-p)$) that the logit-link models separate out via the distributional assumption.  The practical implication is that \emph{linear G-theory estimates applied to pass/fail benchmarks systematically underestimate the proportion of variance attributable to named facets and overestimate residual noise}, leading to artificially compressed---but not necessarily more conservative---reliability estimates.

\paragraph{Bayesian vs.\ frequentist GLME.} The Bayesian and GLME estimates agree in broad strokes but diverge in two instructive ways.  First, the Bayesian posteriors for agent variance are markedly right-skewed, with posterior means 2--5$\times$ larger than posterior medians (e.g., SWE-bench agent mean $= 0.27$, median $= 0.19$).  The GLME point estimate, which approximates the posterior mode, misses this tail mass and underestimates the expected contribution of scaffold variance. Second, the credible intervals from the Bayesian analysis expose the \emph{decision-relevant uncertainty}: for several benchmarks, the 95\% HDI for model variance includes zero, meaning the data are consistent with no true model differentiation at all. This uncertainty is invisible in frequentist analyses.

\paragraph{Nonparametric DISCO.} The DISCO estimates depart most dramatically from the parametric methods on the interaction terms, difference likely due to heterogeneous cluster imbalances (and hence our preference for Bayesian estimates).  DISCO attributes 40--57\% of total dispersion to task$\times$model interactions across benchmarks, compared to 1--11\% from the Bayesian estimates.  This divergence likely reflects two mechanisms: (i)~DISCO's sensitivity to nonlinear dependencies that the additive random-effects models cannot capture, and (ii)~the absence of a separate residual term in DISCO, which forces unexplained variation into the named interactions. The practical upshot is that DISCO serves as a useful \emph{upper bound} on interaction effects and a reminder that the additive decomposition assumed by mixed-effects models may understate the complexity of the task$\times$model relationship.

% A negative error proportion indicates structural variance masking. The high dense consistency of larger groups is masking the true local variance of smaller group. It reveals that the groups are too structurally disparate in sample size and scale to be compared using standard global pooling, thus our reliance on Bayesian methods. The spatial separation between the extreme outliers of a minority group and the centroid of a majority group is so vast that it mathematically eclipses the global baseline dispersion of the pooled data.
%  In energy statistics, a negative within-group dispersion indicates that the average distance between data points inside the same group is mathematically calculated to be larger than the total pooled pairwise distances.The Interpretation: Geometrically, this implies the groups are clustered so tightly and pushed so artificially far apart from each other that the overall baseline variance pool (\(T_{\alpha }\)) is severely underestimated. The proportion assigned to the "Between" facet is no longer a realistic indicator of group differences, but rather a structural artifact of an invalid distance metric. Likely encounter a massive Type I error (false positive), concluding that a specific facet has a highly significant, dominant effect on dispersion when it may actually just be a consequence of poor data scaling or a non-Euclidean distance metric

\subsection{Variance decomposition explains the reliability ceiling and generalizability}
\label{sec:results-variance}

\begin{figure}[!ht]
    \centering
    \includegraphics[width=1\linewidth]{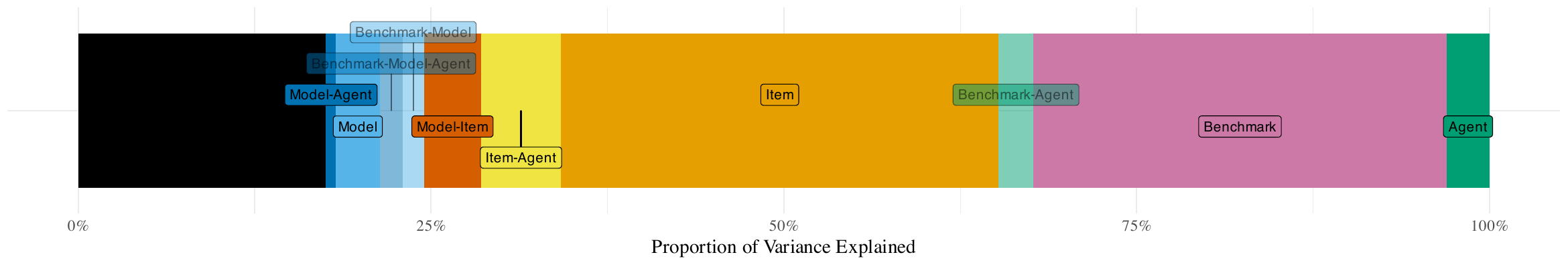}
    \caption{Full variance decomposition of Eq. \ref{eq:pooled}
    % ; proportions estimated via Eq. \ref{eq:var-share}
    as percentages of total variance. Variance decompositions for individual benchmarks are in Appendix~\ref{app:add_plots}. Black represents residual variance.}
    \label{fig:variance-decomposition}
\end{figure}

Figure~\ref{fig:variance-decomposition} reports posterior variance shares from the leaderboard model. Persistent model variation is smaller than the combined variation associated with scaffolds, benchmark-conditioned model performance, and item-conditioned interactions. Most observed item outcomes therefore contain more condition-specific variation than globally transferable model signal.

This does not imply that model capability is unimportant. Reliability is population-dependent: contemporary frontier models are often similar, whereas tasks and scaffolds are heterogeneous. A benchmark could have ranked a historically broader model population reliably and yet fail to distinguish the current frontier. Likewise, a future benchmark may become uninformative as models saturate it. Reliability must consequently be re-estimated as the evaluated population and agent ecosystem change.

For strong generalization to new tasks, we would want to see larger main effects for models and scaffolds, making decisions clearer. However, the reality is that the second-order task-level interactions, model--task and scaffold--task, represent larger variation components than their respective main effects. 
% Importantly, the decomposition clarifies how well model and scaffold selections generalize. By comparing the relative proportions of variance, we see that the task-level interactions with both model and agent are larger than the respective main effects. 
This quantifies that, instead of providing clear architecture choices, some models or scaffolds are better at certain tasks than others, implying that a developer should, for a given task, choose a specific model and scaffold for each new task, rather than for the group of tasks represented by a benchmark or leaderboard (because of a smaller benchmark--model component). This is not ideal for developers, as it means they would need to know about the demands of each task and the appropriateness of architecture selections for it. This partly explains why there is such low measurement reliability for fewer numbers of items in Figures \ref{fig:mod_v_agent_cis} and \ref{fig:ep2_and_snr}: a different model scaffold might be optimal for each task item. 

% Further, a small model--scaffold component means that relative model performance does not depend on the harness as much as the task depends 

% The decomposition also distinguishes two failure modes that aggregate robustness statistics combine. A large model--scaffold component means that relative model performance depends on the harness. A large benchmark--model component means that capability is benchmark-specific. These require different remedies: the former calls for multiple common scaffolds or explicit system-level reporting; the latter calls for a broader and better-connected benchmark battery.

\subsection{Absolute Reliability on Linear Scale}\label{app:abs_reliab}
Generalizability theory also permits estimation of reliability of numeric scores. Absolute reliability, as it is known, is always less than or equal to the corresponding estimate of relative reliability. But because scores are on the observed scale, all of the analyses in this section are based on the Bayesian linear mixed model (\S~\ref{app:bayes_lme}). They are illustrative and serve as a framework for practitioners looking to measure reliability of assigned numeric scores.

\begin{figure}
    \centering
    \includegraphics[width=1\linewidth]{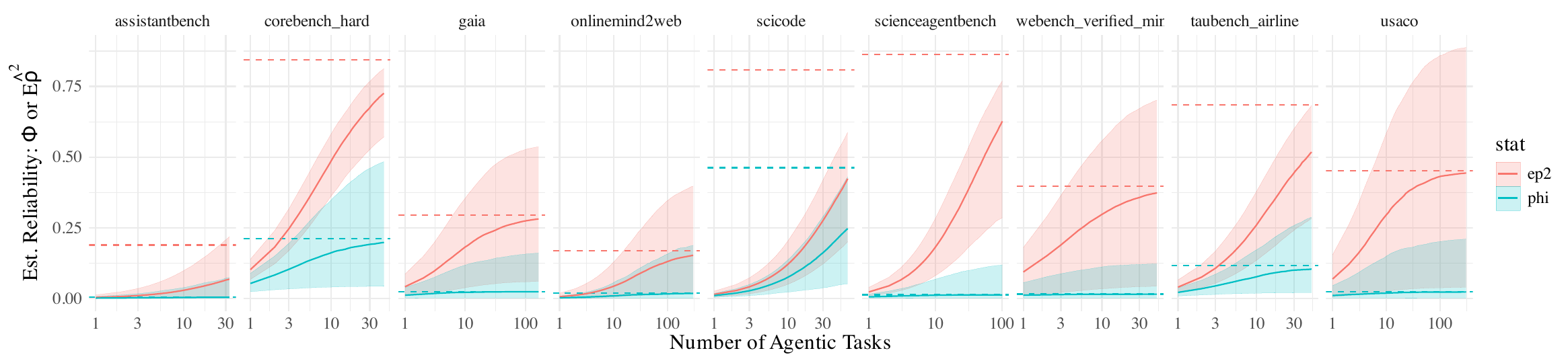}
    \caption{Absolute vs Relative Reliability on Observed Scale}
    \label{fig:observed_scale_rel_abs}
\end{figure}

\paragraph{Reliable \textit{rankings} do not imply reliable \textit{scores}.} Relative reliability (\(E\rho^2\)) asks whether the ranking of models stays similar across evaluation conditions, while absolute reliability (\(\Phi\)) asks whether their reported scores stay similar. Figure~\ref{fig:observed_scale_rel_abs} shows that this distinction matters substantially in practice: absolute reliability is consistently lower across the nine benchmarks, and for several benchmarks remains near zero even as ranking reliability increases with additional tasks. An evaluation can therefore support a stable ranking without supporting equally strong claims about a model's absolute score or whether it exceeds a fixed threshold.

While not recommended for the HAL dataset and benchmarks, there may be some agentic benchmarks where absolute scores are needed for decision-making. Absolute reliability is most interpretably performed on the observed scale, rather than the latent scale used throughout the main body of this study. For the HAL dataset, there is simply not enough signal relative to noise in the benchmarks for any reliable absolute scoring. Nevertheless, the concept of absolute reliability may be important for some agentic measures.

\subsubsection{Absolute reliability: numeric values of scores carry meaning}

Rank-based reliability measures whether an evaluation preserves the ordering of models or systems, but it does not establish whether their score levels are reproducible. Absolute reliability is important when scores are interpreted directly---for example, when assessing whether a model exceeds a deployment threshold, meets a minimum capability requirement, or achieves a specified performance target. In these settings, a task set or scaffold that shifts every model's score equally can change the decision even without changing the ranking. The absolute reliability, or dependability coefficient, \(\Phi\), therefore includes these common shifts in its error variance.

Treating tasks and scaffolds as random facets, the absolute reliability of a model score under an equal-allocation design is
\begin{equation}
\Phi_{M(b)}(n_i,n_a)
=
\frac{\sigma_M^2}
{\sigma_M^2
+(\sigma_I^2+\sigma_{IM}^2)/n_i
+(\sigma_A^2+\sigma_{MA}^2)/n_a
+(\sigma_{IA}^2+\sigma_{IMA,e}^2)/(n_i n_a)}.
\label{eq:perbench-model-absolute-reliability}
\end{equation}
Relative to model-ranking reliability, the denominator additionally includes task main effects, scaffold main effects, and task--scaffold interactions. These components affect the absolute score of a model averaged over sampled tasks and scaffolds, even when they do not affect its position relative to other models. Increasing the number of tasks reduces task-related error, whereas increasing the number of scaffolds reduces scaffold-related error.

When the object of measurement is the complete \textit{model--scaffold system}, scaffold differences are part of the system's universe score rather than measurement error. Absolute reliability is then
\begin{equation}
\Phi_{MA(b)}(n_i)
=
\frac{\sigma_M^2+\sigma_A^2+\sigma_{MA}^2}
{\sigma_M^2+\sigma_A^2+\sigma_{MA}^2
+(\sigma_I^2+\sigma_{IM}^2+\sigma_{IA}^2+\sigma_{IMA,e}^2)/n_i}.
\label{eq:perbench-system-absolute-reliability}
\end{equation}
Here, task main effects enter the denominator because sampling an easier or harder task set can shift a system's score across an absolute decision threshold. In contrast, scaffold main effects and model--scaffold interactions remain signal because they distinguish the systems being evaluated.

For either object of measurement, absolute reliability is no greater than the corresponding ranking reliability:
\begin{equation}
\Phi_{M(b)}(n_i,n_a)
\leq E\rho_{M(b)}^2(n_i,n_a),
\qquad
\Phi_{MA(b)}(n_i)
\leq E\rho_{MA(b)}^2(n_i).
\end{equation}
A large gap indicates that rankings may be reproducible even though reported score levels are sensitive to the sampled evaluation conditions. Likewise, high inter-scaffold ranking reliability does not guarantee absolute agreement: two scaffolds can preserve model ordering while producing systematically different scores.

\begin{figure}
    \centering
    \includegraphics[width=1\linewidth]{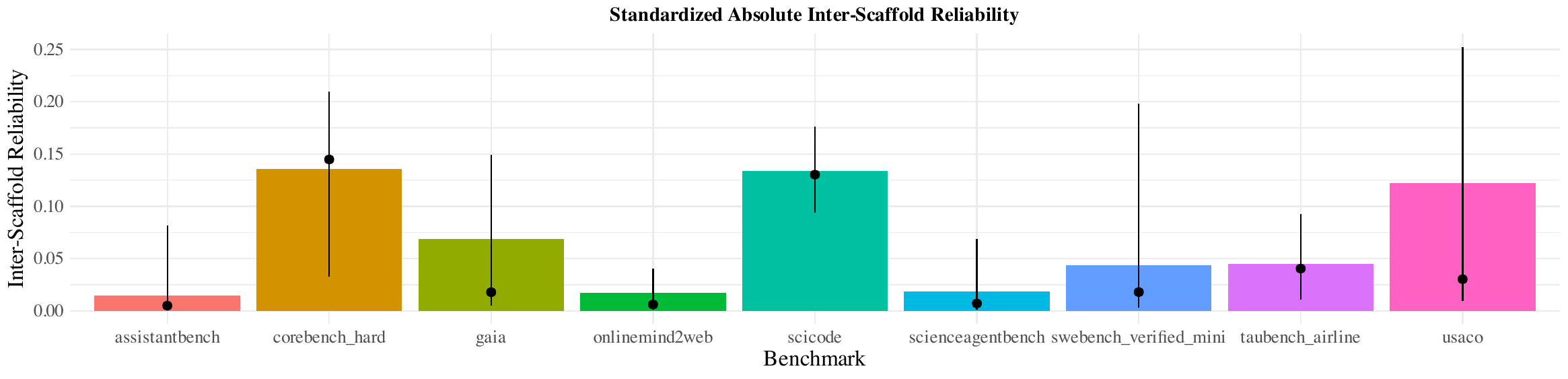}
    \caption{Standardized Absolute Inter-scaffold Reliability. Standardized inter-scaffold reliability looks at the reliability for a single fixed task, rather than the composite reliability across all tasks in a benchmark.}
    \label{fig:abs_interscaffold}
\end{figure}

\subsection{Full Estimation Tables for Proportion of Latent Variation Explained}\label{sec:app-tables}

Full variance-proportion tables for all models and all five estimation methods are provided %in the supplementary materials. Tables in 
% the main text report Bayesian posterior summaries; 
in the following tables including Bayesian posterior summaries and also LME, GLME, and DISCO estimates to enable cross-method comparison.

\begin{table}[!h]
\centering
\scriptsize
\caption{Proportion of variation explained per benchmark. For comparability across methods, Bayesian estimates represent posterior means (rather than medians found in most of the study) to preserve proportional relationships.}
\label{tab:propvardecomp3}
\resizebox{\ifdim\width>\linewidth\linewidth\else\width\fi}{!}{
\begin{tabular}{llrrrrr}
\toprule
\textbf{Benchmark} & \textbf{Facet} & \textbf{LMM} & \textbf{Bayes LMM} & \textbf{GLMM} & \textbf{Bayes GLMM} & \textbf{DISCO}\\
\midrule
scicode & item & 0.235 & 0.232 & 0.766 & 0.718 & 0.209\\
scicode & agent & 0.001 & 0.041 & 0.004 & 0.054 & 0.001\\
scicode & model & 0.010 & 0.011 & 0.053 & 0.057 & 0.012\\
scicode & item\_agent & 0.037 & 0.036 & 0.012 & 0.016 & 0.252\\
scicode & item\_model & 0.130 & 0.123 & 0.012 & 0.035 & 0.509\\

scicode & model\_agent & 0.002 & 0.004 & 0.003 & 0.015 & 0.017\\
scicode & sigma & 0.584 & 0.553 & 0.150 & 0.105 & NA\\

\addlinespace
gaia & item & 0.210 & 0.113 & 0.342 & 0.280 & 0.169\\
gaia & agent & 0.041 & 0.471 & 0.029 & 0.194 & 0.017\\
gaia & model & 0.030 & 0.020 & 0.040 & 0.042 & 0.046\\
gaia & item\_agent & 0.036 & 0.020 & 0.035 & 0.036 & 0.209\\
gaia & item\_model & 0.097 & 0.051 & 0.041 & 0.102 & 0.481\\
gaia & model\_agent & 0.059 & 0.039 & 0.083 & 0.079 & 0.077\\
gaia & sigma & 0.527 & 0.286 & 0.431 & 0.267 & NA\\

\addlinespace
taubench\_airline & item & 0.166 & 0.140 & 0.277 & 0.244 & 0.161\\
taubench\_airline & agent & 0.043 & 0.237 & 0.044 & 0.180 & 0.021\\
taubench\_airline & model & 0.027 & 0.024 & 0.039 & 0.035 & 0.025\\
taubench\_airline & item\_agent & 0.086 & 0.070 & 0.107 & 0.101 & 0.250\\
taubench\_airline & item\_model & 0.035 & 0.024 & 0.000 & 0.033 & 0.484\\
taubench\_airline & model\_agent & 0.012 & 0.012 & 0.016 & 0.020 & 0.059\\
taubench\_airline & sigma & 0.631 & 0.493 & 0.518 & 0.386 & NA\\

\addlinespace
swebench\_verified\_mini & item & 0.153 & 0.070 & 0.349 & 0.277 & 0.092\\
swebench\_verified\_mini & agent & 0.246 & 0.645 & 0.103 & 0.260 & 0.073\\
swebench\_verified\_mini & model & 0.056 & 0.027 & 0.272 & 0.142 & 0.119\\
swebench\_verified\_mini & item\_agent & 0.068 & 0.034 & 0.024 & 0.026 & 0.190\\
swebench\_verified\_mini & item\_model & 0.089 & 0.039 & 0.009 & 0.059 & 0.399\\
swebench\_verified\_mini & model\_agent & 0.057 & 0.036 & 0.042 & 0.130 & 0.126\\
swebench\_verified\_mini & sigma & 0.331 & 0.149 & 0.202 & 0.106 & NA\\

\addlinespace
usaco & item & 0.257 & 0.141 & 0.551 & 0.450 & 0.239\\
usaco & agent & 0.072 & 0.469 & 0.000 & 0.090 & 0.003\\
usaco & model & 0.062 & 0.022 & 0.000 & 0.049 & 0.027\\
usaco & item\_agent & 0.200 & 0.114 & 0.130 & 0.172 & 0.262\\
usaco & item\_model & 0.184 & 0.103 & 0.000 & 0.116 & 0.440\\
usaco & model\_agent & 0.000 & 0.027 & 0.103 & 0.066 & 0.030\\
usaco & sigma & 0.225 & 0.124 & 0.216 & 0.057 & NA\\

\addlinespace
assistantbench & item & 0.116 & 0.074 & 0.437 & 0.331 & 0.156\\
assistantbench & agent & 0.000 & 0.377 & 0.000 & 0.181 & 0.002\\
assistantbench & model & 0.000 & 0.003 & 0.000 & 0.024 & 0.017\\
assistantbench & item\_agent & 0.083 & 0.057 & 0.107 & 0.137 & 0.219\\
assistantbench & item\_model & 0.047 & 0.022 & 0.000 & 0.057 & 0.574\\
assistantbench & model\_agent & 0.013 & 0.009 & 0.069 & 0.058 & 0.032\\
assistantbench & sigma & 0.741 & 0.458 & 0.386 & 0.213 & NA\\

\addlinespace
corebench\_hard & item & 0.228 & 0.178 & 0.415 & 0.362 & 0.153\\
corebench\_hard & agent & 0.063 & 0.286 & 0.047 & 0.180 & 0.027\\
corebench\_hard & model & 0.071 & 0.056 & 0.134 & 0.113 & 0.074\\
corebench\_hard & item\_agent & 0.010 & 0.009 & 0.000 & 0.005 & 0.200\\
corebench\_hard & item\_model & 0.099 & 0.074 & 0.029 & 0.063 & 0.462\\
corebench\_hard & model\_agent & 0.012 & 0.012 & 0.011 & 0.016 & 0.083\\
corebench\_hard & sigma & 0.517 & 0.385 & 0.365 & 0.262 & NA\\

\addlinespace
onlinemind2web & item & 0.283 & 0.205 & 0.414 & 0.372 & 0.239\\
onlinemind2web & agent & 0.000 & 0.272 & 0.000 & 0.085 & 0.000\\
onlinemind2web & model & 0.002 & 0.004 & 0.003 & 0.009 & 0.007\\
onlinemind2web & item\_agent & 0.133 & 0.097 & 0.164 & 0.163 & 0.296\\
onlinemind2web & item\_model & 0.022 & 0.014 & 0.000 & 0.023 & 0.446\\
onlinemind2web & model\_agent & 0.016 & 0.013 & 0.027 & 0.029 & 0.012\\
onlinemind2web & sigma & 0.545 & 0.395 & 0.393 & 0.319 & NA\\

\addlinespace
scienceagentbench & item & 0.246 & 0.129 & 0.532 & 0.413 & 0.208\\
scienceagentbench & agent & 0.042 & 0.505 & 0.048 & 0.258 & 0.011\\
scienceagentbench & model & 0.016 & 0.008 & 0.026 & 0.020 & 0.016\\
scienceagentbench & item\_agent & 0.086 & 0.045 & 0.049 & 0.046 & 0.249\\
scienceagentbench & item\_model & 0.034 & 0.017 & 0.000 & 0.021 & 0.493\\
scienceagentbench & model\_agent & 0.000 & 0.003 & 0.008 & 0.014 & 0.023\\
scienceagentbench & sigma & 0.577 & 0.294 & 0.338 & 0.227 & NA\\
\bottomrule
\end{tabular}}
\end{table}

\begin{table}[!h]
\centering
\caption{Proportion of variation explained full leaderboard. For comparability across methods, Bayesian estimates represent posterior means (rather than medians found in most of the study) to preserve proportional relationships.}
\label{tab:propvardecomp4}
\resizebox{\ifdim\width>\linewidth\linewidth\else\width\fi}{!}{
\begin{tabular}{lrrrrr}
\toprule
\textbf{Facet} & \textbf{LMM} & \textbf{Bayes LMM} & \textbf{GLMM} & \textbf{Bayes GLMM} & \textbf{DISCO}\\
\midrule
agent & 0.016 & 0.039 & 0.020 & 0.040 & 0.014\\
benchmark & 0.075 & 0.108 & 0.259 & 0.299 & 0.023\\
benchmark\_agent & 0.028 & 0.036 & 0.026 & 0.031 & 0.030\\
benchmark\_item & 0.252 & 0.234 & 0.328 & 0.297 & 0.117\\
benchmark\_item\_agent & 0.070 & 0.065 & 0.052 & 0.055 & 0.140\\
\addlinespace
benchmark\_item\_model & 0.047 & 0.043 & 0.004 & 0.040 & 0.240\\
benchmark\_model & 0.010 & 0.008 & 0.020 & 0.015 & 0.040\\
benchmark\_model\_agent & 0.013 & 0.014 & 0.014 & 0.016 & 0.046\\
model & 0.026 & 0.025 & 0.033 & 0.032 & 0.015\\
model\_agent & 0.007 & 0.006 & 0.009 & 0.008 & 0.035\\
\addlinespace
benchmark\_task\_model\_agent + sigma & 0.456 & 0.422 & 0.236 & 0.168 & 0.302\\
\bottomrule
\end{tabular}}
\end{table}

\end{document}